\documentclass[letterpaper]{article}
\usepackage[letterpaper]{geometry}
\usepackage{graphicx}
\usepackage{booktabs}
\usepackage{amsmath,amssymb}
\usepackage{mathtools}
\usepackage{adjustbox,placeins}
\usepackage[table*]{xcolor}
\usepackage{comment}
\usepackage{colortbl}
\usepackage{enumitem}
\usepackage{array}
\definecolor{Gray}{cmyk}{0.1,0.1,0.1,0.1}
\definecolor{lightgreen}{RGB}{144,238,144}
\newcolumntype{H}{>{\setbox0=\hbox\bgroup}c<{\egroup}@{}}
\usepackage{algorithm}
\usepackage{algorithmic}
\usepackage{newfloat}
\usepackage{hyperref}
\usepackage{listings}
\usepackage{titlesec}
\usepackage{sectsty}
\usepackage{etoolbox}

\title{\textbf{Diverse Diffusion: Enhancing Image Diversity in Text-to-Image Generation}}
\author{
    \textbf{Mariia Zameshina}\\
    \small{Univ Gustave Eiffel, CNRS, LIGM} \\ \small{F-77454 Marne-la-Vallee, France}\\
    \small{mariia.zameshina@esiee.fr}
    \and
    \textbf{Olivier Teytaud}\\
    \small{TAO, CNRS - INRIA - LRI}\\
    \and
    \textbf{Laurent Najman}\\
    \small{Univ Gustave Eiffel, CNRS, LIGM} \\
    \small{F-77454 Marne-la-Vallee, France}
}

\date{}

\begin{document}

\maketitle

\begin{abstract}

Latent diffusion models excel at producing high-quality images from text. Yet, concerns appear about the lack of diversity in the generated imagery. To tackle this, we introduce Diverse Diffusion, a method for boosting image diversity beyond gender and ethnicity, spanning into richer realms, including color diversity.

Diverse Diffusion is a general unsupervised technique that can be applied to existing text-to-image models. Our approach focuses on finding vectors in the Stable Diffusion latent space that are distant from each other. We generate multiple vectors in the latent space until we find a set of vectors that meets the desired distance requirements and the required batch size.

To evaluate the effectiveness of our diversity methods, we conduct experiments examining various characteristics, including color diversity, LPIPS metric, and ethnicity/gender representation in images featuring humans.

The results of our experiments emphasize the significance of diversity in generating realistic and varied images, offering valuable insights for improving text-to-image models. Through the enhancement of image diversity, our approach contributes to the creation of more inclusive and representative AI-generated art.

\end{abstract}

\section{Introduction}

Latent diffusion models have gained significant attention for text-to-image generation, with DALL-E \ref{pmlr-v139-ramesh21a} and Stable Diffusion \ref{rombach2022high} being some of the most prominent examples. While they have shown impressive results in generating high-quality images from textual descriptions, there have been concerns regarding potential causes of their usage for various sensitive applications.

To prevent the potential misuse of these models, a few recent efforts have been made. For instance, \ref{chambon2022adapting} investigated the efficacy of Stable Diffusion for the medical imaging. Additionally, \ref{somepalli2022diffusion} explored image-retrieval frameworks to detect content replication in generated images. \ref{sha2022fake} developed a classifier to trace back generated images to their source models, ensuring accountability for creators. Furthermore, \ref{karthik2023if} created faithful diffusion that selects images corresponding to a prompt.

\begin{figure}[h!] \begin{center}
\par
{\includegraphics[width=0.97\columnwidth]{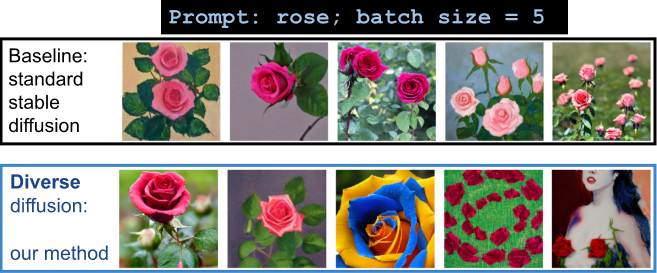}} %
\hfill
 \caption{\label{SD_schema} Images generated with standard Stable Diffusion and our  method for the prompt ``rose'', batch size = 5. Here, our method is shown to select images not only diverse in colors, but also in ideas, compared to Stable Diffusion. }
\end{center}
\end{figure}

Bias can arise from various sources, such as the training data or the algorithmic biases inherent in the model architecture. \ref{struppek2022biased} shows that text-to-image models pick up cultural biases linked to various Unicode scripts. While increasing training data quality remains a significant concern and solutions such as \ref{smith2023balancing} have been proposed to tackle dataset level bias, it is also important to ensure that the diffusion-based methods do not amplify any biases present in the training data.

The lack of diversity problem was addressed in \ref{cheuk}, \ref{bianchi2023easily} and \ref{fraserdiversity}. There, the authors notice that images generated by Stable Diffusion lack diverse cultural representation and are prone to gender stereotypes. \ref{berns2022increasing} highlights the need for algorithmic adjustments in generative models to increase the diversity of their output for multi-solution tasks. It also proposes a framework that integrates automated machine learning with computational creativity to automate key tasks in artistic pipelines and increase the creative autonomy of computational agents. Further, \ref{theron2023evidence} argues that Stable Diffusion v1-4 violates demographic parity in generating images of a doctor given a gender- and skin-tone-neutral prompt. The author  observed that the model is biased towards generating images of perceived male figures with lighter skin, with a significant bias against figures with darker skin, as well as a notable bias against perceived female figures. According to \ref{gizmodo}, AI image generators often display gender and cultural biases. Stable Diffusion as other models has inherent biases from the training datasets. It was found, for instance, that Stable Diffusion depicts all engineers as male despite women making up around 20\% of people in engineering professions. According to \ref{chefer2023hidden} Stable Diffusion bears the nontrivial biases due to learned relations between not necessarily related concepts like ``millennials'' and ``drinking''.

There are recent methods focusing on diversity for Stable Diffusion generated images as well. Most of them focus on specific domains to broaden the image variability. For instance, \ref{shipard2023diversity} and \ref{bansal2022well} report that adding  supporting context to text-to-image models prompts increases diversity in both general and human-specific fashion. In \ref{samuel2023norm} the authors report increase in generation of rare-concept images following the seed manipulation. \ref{kim2023stereotyping} trains prompt embeddings that would guide to generate images fairly according to the set of sensitive attributes such as gender and ethnicity.

Promoting novelty and diversity is a challenge that exists both for image and text generation. In \ref{xu2018diversity} the authors emphasize the importance of producing novel and diverse textual outputs. By leveraging a language-model-based discriminator, their model DP-GAN assigns high rewards to text that exhibits both novelty and fluency. Our approach for image generation also goes beyond addressing sensitive fairness concerns and opens new avenues for creative expression, as we focus not only on representation of people with different ethnicities and genders, but also, as our approach is unsupervised, on diversity of color and other dissimilarities of images in a batch.

It is common to generate multiple image versions using latent diffusion models. Even in non-sensitive scenarios, having diverse outputs is essential. A collection of similar images holds little advantage over a single image. Therefore, we propose a method that encompasses various domains, including faces, cars, animals and more to enhance diversity in text-to-image generation. By increasing diversity within batches, we can reduce the number of required generation iterations. For instance, by increasing the probability of obtaining satisfactory images from $1\%$ with a vanilla Stable Diffusion to $5\%$ using our method, the expected number of generated batches before satisfaction decreases by a factor~5.

Overall, in this article, we present Diverse Diffusion, {\em i.e.}, modifications to the Stable Diffusion algorithm  that facilitate the generation of diverse images and thereby help to create {more inclusive} art within {a limited number of Stable Diffusion generations} (and therefore {less computational power}) in {unsupervised} fashion. Images generated with and without our approach are illustrated in Figure \ref{SD_schema}.

We highlight the following contributions: (i) A general unsupervised technique that can be applied to existing text-to-image models to increase image diversity, which is essential for generating realistic and varied images. (ii) Experiments that demonstrate the diversity advantages of our proposed approach.

\section{Diversity algorithms}
There are different approaches to generating diverse point sets in an unsupervised manner. For example, Latin Hypercube Sampling \ref{LHS}, low discrepancy methods  \ref{atanassov2004discrepancy}\ref{hammersley}\ref{nie}   and low dispersion. Here, we focus on low dispersion, optimized by a very limited random search for staying in the domain of validity of our latent variables. We aim to find vectors in the latent space of Stable Diffusion that are distant from one another. To accomplish this, we generate multiple vectors in the latent space until we obtain a set that contains the required number of vectors (determined by the batch size) and satisfies a specific distance requirement.

In the ``\textbf{baseline}'' setting, we generate images using the standard, unmodified version of Stable Diffusion without imposing any distance requirements on the latent space.

In the ``\textbf{cap}'' setting, we enforce a minimum requirement of $d_{\text{min}}$ on the vectors corresponding to all pairs of images within a batch. We illustrate the procedure of choosing a set of latent vectors $V$ for batch size $B$ and minimum distance requirement $d_{min}$ in algorithm \ref{algocap}.

\begin{algorithm}[tb]
\caption{Generating diverse vectors in the ``cap'' setting \label{algocap}}
\label{algo:cap}
\begin{algorithmic}[1]
\REQUIRE batch size $B$, minimum distance $d_{\min}$
\ENSURE Set of diverse vectors $V$
\STATE $V \leftarrow \emptyset$ \COMMENT{Initialize an empty set of vectors}
\WHILE{$|V| < B$}
    \STATE $v_{\text{new}} \leftarrow \text{GenerateNewVector()}$ \COMMENT{Generate a new vector in the latent space of stable diffusion}
    \IF{$\text{MinDistance}(v_{\text{new}}, V) \geq d_{\min}$}
        \STATE $V \leftarrow V \cup \{v_{\text{new}}\}$ \COMMENT{Add the new vector to the set}
    \ENDIF
\ENDWHILE
\RETURN $V$ \COMMENT{Return the set of diverse vectors}
\end{algorithmic}
\end{algorithm}

In the \textbf{``max''} setting, we impose {a} maximum number of iterations on searching for a new vector that would have {a} maximal minimal distance to all the already selected vectors in the batch. We illustrate the procedure of choosing a set of latent vectors $V$ for batch size $B$ and {a} maximum number of iterations requirement $N_{\text{max}}$ in algorithm \ref{algomax}.

\begin{algorithm}[tb]
\caption{Generating diverse vectors in the ``max'' setting\label{algomax}}
\begin{algorithmic}[1]
\REQUIRE batch size $B$, maximum iterations number $N_{\text{max}}$
\ENSURE Set of diverse vectors $V$
\STATE $V \leftarrow \emptyset$ \COMMENT{Initialize an empty set of vectors}
\FOR{$b = 1$ \TO $B$}
 \STATE $v_{\text{farthest}} \leftarrow \text{GenerateNewVector()}$
 \FOR{$N = 1$ \TO $N_{\text{max}}-1$}

    \STATE $v_{\text{new}} \leftarrow \text{GenerateNewVector()}$ \COMMENT{Generate a new vector in the latent space of stable diffusion}

    \IF{$\text{MinDistance}(v_{\text{new}}, V) > \text{MinDistance}(v_{\text{farthest}}, V)$}
        \STATE $v_{\text{farthest}} \leftarrow v_{\text{new}}$
    \ENDIF
    \COMMENT{Find the vector with the maximum minimum distance to the existing set}
    \STATE $V \leftarrow V \cup \{v_{\text{farthest}}\}$ \COMMENT{Add the farthest vector}
 \ENDFOR
\ENDFOR
\RETURN $V$ \COMMENT{Return the set of diverse vectors}
\end{algorithmic}
\end{algorithm}

In setting \textbf{``pooling\_cap''} and  \textbf{``pooling\_max''} we apply the same exact methods as in ``cap'' and ``max'' but the distance is calculated differently. Specifically, the distance is computed between the vectors that were initially processed by average pooling $8 \times 8$ which down-samples the vector size to $4 \times 8 \times 8$.

We use  two different settings for generating diverse image batches: \textbf{standard experiment} and \textbf{ long experiment}.

The parameters of a standard experiment are the following: in setting  ``cap'' the minimal distance between latent vectors should be at least $182$, while in the setting  ``pooling\_cap'' it should be at least $3.1$, in setting  ``max'' and ``pooling\_max''  the number of iterations after which the farthest vector is found is $100$.

The parameters of a long experiment are the following:  (i) in setting  ``cap'' the minimal distance between latent vectors should be at least $183$, while in setting  ``pooling\_cap'' it should be at least $3.1$. In the setting  ``max'' and ``pooling\_max''  the number of iterations after which the farthest vector is found is $10000$.

The choice of settings provides variable diversity levels and variable computational complexity. In the current paper for small batch sizes ($3, 5, 10$), we create both standard and long experiments, and for the big batch sizes ($50$) we create only standard experiments. In the cap and pooling cap setting, due to the distance limitations and no limit on a number of iterations, the experiment for big batches becomes increasingly slower. That is why, for purposes of limiting computational cost, we recommend the ``pooling\_max'' method for batch sizes more than $50$.

\maketitle
\section{Evaluation Methods}
In order to evaluate diversity of generated images, we use both quantitative and human evaluation methods. For quantitative methods, we focus primarily on color diversity image similarity metric LPIPS \ref{zhang2018unreasonable}. Increasing image color diversity is an important aspect of ensuring general diversity of the images: we would like to get representation for people, animals, and objects of all colors.

\subsection{Color diversity }\label{color_diversity}
To assess the color diversity in an image batch $b$, we employ a method that involves extracting color information from each image in the batch using the RGB color model, which represents colors as a combination of red, green, and blue channels. Specifically, we compute the mean value for each channel ($R_i, B_i, G_i$) in the given image $i$, and identify whether one of these colors is predominantly present in the image $i$.

To determine the dominant color in image $i$ with respect to a coefficient $K$, we define the image as having a dominant color of Red if $R_i >K\times max(G_i, B_i)$, as Green if $G_i >K\times max(R_i, B_i)$, as Blue if $B_i >K\times max(G_i, R_i)$, and as None if none of these inequalities are true. We denote the dominant color of image $i$ with respect to the coefficient $K$ as $D_K(i)$.

To evaluate color diversity of an image batch $b$, we compute a number of dominant colors in that batch with respect to a coefficient $K$, $N_K(b)$.
\begin{align}
N_K(b) = 3 &- \prod_{i \in b} (1 - I(D_K(i) == Red)) \nonumber\\
&- \prod_{i \in b} (1 - I(D_K(i) == Green)) \nonumber\\
&- \prod_{i \in b} (1 - I(D_K(i) == Blue)),
\end{align}
where $I$ is an indicator function and $D_K(i)$ is a dominant color of an image $i$ with respect to the coefficient $K$.

The first color diversity metric across the batches set $B$ is an average number of dominant colors present in the batches of that set with respect to a coefficient $K$.
\begin{align}
Avg_K(B) = \dfrac{\sum_{b \in B} N_K(b)}{||B||},
\end{align}
where $||B||$ is the total number of batches in the set $B$.

The second color diversity metric aims to compute  proportion of batches that contain at least one image predominantly exhibiting red, blue or green color, considering RGB image encoding. This allows for the quantification of color variability within a batch.

Specifically, for various values of a coefficient $K$ and batch set B we compute the proportion
\begin{align}
C3_K(B) = \dfrac{\sum_{b \in B} I(N_K(b)) == 3}{||B||},
\end{align}
where $||B||$ is the total number of batches in the set $B$  and $I$ is an indicator function.

Another color diversity metric is the proportion of batches containing images with different dominant colors (not all the images in a batch are predominantly red, blue or green).

Specifically, for various values of a coefficient $K$ and batch b we compute  the proportion
\begin{align}
C2_K(B) = \dfrac{\sum_{b \in B} I(N_K(b)) >= 2}{||B||},
\end{align}
where $||B||$ is the total number of batches in the set $B$  and $I$ is an indicator function.

\subsection{LPIPS metric}
Similarly to \ref{ham2023modulating} and \ref{huang2022draw}, we use LPIPS (Learned Perceptual Image Patch Similarity) \ref{zhang2018unreasonable} to evaluate the diversity of the images generated by Stable Diffusion. We measure the pairwise similarity between images in a batch and compare the obtained values for different modifications of Stable Diffusion generation algorithm.

%\subsection{Entropy across clusters}

%Finally, following \ref{samuel2023norm} and \ref{samuel2023all} we compute mean entropy across each image batch.

\subsection{Ethnicity and gender classification for images portraying humans}
As mentioned in \ref{theron2023evidence}, Stable Diffusion may lack diversity in ethnicity representation. That is why, for the prompts that we use for the human face generation, we compare ethnic diversity in between our methods and basic version of Stable Diffusion. In order to identify the ethnicity of a person present on an image, we use DeepFace ethnicity recognition \ref{taigman2014deepface}. In particular, we identify the following groups of ethnicities: (i) Black, (ii) Asian, (iii) Hispanic, (iv) White or Middle Eastern. For gender classification (male/female), we also use DeepFace.
We compute the percentage of batches where all pairs of genders and ethnicities are present or at least 3 out of 4 ethnicities are present (similarly to colors).

\subsection{Multiplicative improvement}
Our method aims to increase representation of underrepresented groups. That is why for all the metrics mentioned above we compute percentage versus multiplicative improvement score. Here percentage stands for the percentage of batches that follow some characteristic $C$ (for example, contain images of Asian men) and multiplicative improvement stands for multiplicative increase in this percentage for our preferred method (\textbf{``pooling\_cap''}) compared to the baseline method (\textbf{standard Stable Diffusion}). By using this metric, we can evaluate our efforts in promoting the inclusivity of underrepresented classes.

\section{Experiment setting}
As a baseline, we use Stable Diffusion v-5 with PNDM Scheduler.
We measure diversity using LPIPS or artificial classes (image hue) or human-centered criteria (gender/ethnicity) and check various batch sizes.
We use machines with $8$ Tesla V100-SXM2-32GB GPUs and $80$ x86\_64 CPUs. {The full code is (anonymously) provided in https://anonymous.4open.science/r/DiverseDiffusion-1012.}
{See supplementary material for reproducibility details.}

\subsection{Experiments on small batches \label{small_prompt_list}}

For our small batch experiment setup, we choose the following list of prompts: ``face'', ``rose'', ``butterfly'', ``cat'', ``horse'', ``car'', ``ornament'', ``bird'', ``color'', ``a professional photograph of an adult person face'', ``photo of an animal in the grass'' and ``octane, hyperrealistic, backlit''.

In each experiment, we consider batches of $3$, $5$ and $10$ images in order to compute diversity metrics in each of these batches, and to compare our modifications with original Stable Diffusion. For each batch size and each modification, we create at least $2500$ batches.

\subsection{Experiments on big batches \label{big_prompt_list}}

For our big batch experiment setup, we choose the following list of prompts: ``a professional photograph of a man face'', ``a photograph of a person with different colored eyes'', ``a passport-style photograph of a person's face'', ``a professional photograph of an adult person face'', ``a close-up photograph of an elderly person's face'' and ``a beauty shot of a model's face''. These prompts are all centered on human photos and thus allow us to evaluate not only LPIPS and color diversity but also gender and ethnicity variation, which are crucial to ensure in any human-centered applications of diffusion models. In each experiment, we consider batches of $50$ images in order to compute diversity metrics in each of these batches and compare our modifications with the baseline: original Stable Diffusion. For each modification, we create at least $900$ batches.

\section{Experiments and results}
\begin{figure}[h!] \begin{center}
\par
{\includegraphics[width=0.97\columnwidth]{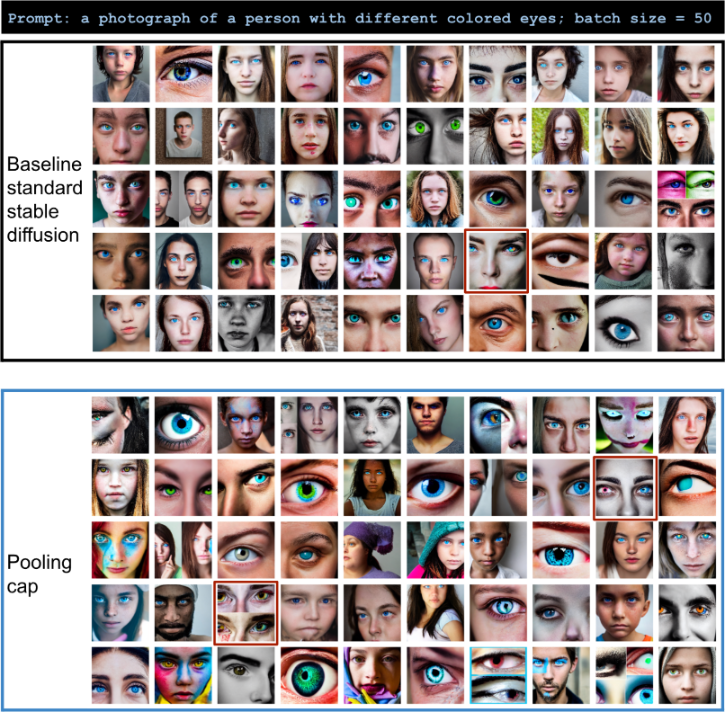} }%
\hfill
\caption{\label{images_bs50} Examples of images generated with the prompt ``a photograph of a person with different colored eyes'':
 pooling\_cap against baseline Stable Diffusion, batch size=50. Images corresponding to the %prompt
 {``expected output'' of the prompt} are highlighted in red. }
\end{center}
\end{figure}

An example of image batches generated with our method ``pooling\_cap'' and original Stable Diffusion is presented in Figure \ref{images_bs50}. In two randomly chosen examples among all batches generated by the baseline and ``pooling\_cap'' for the prompt ``a photograph of a person with different colored eyes'' and batch size 50, we can see that while getting images corresponding to the prompt remains difficult for our chosen method, we still get improvement both in number of images corresponding to the prompt and ethnic diversity.

Further, we present the experimental evaluation of our proposed methods for promoting diversity in generated image batches. We assess the color diversity, gender and ethnicity representation, and image diversity using the LPIPS metric. The experiments are conducted on various prompts and compared against the baseline Stable Diffusion method.

\subsection{Color evaluation}
In this subsection, we evaluate the color diversity of the generated image batches. We present multiplicative improvement results, summarizing the color diversity improvement for various values of the coefficient $K$ across different prompts specified in Section ``Experiments on small batches''. Other experiment results are provided in the supplementary material.
\begin{figure}[h!] \begin{center}
\par
{\includegraphics[width=0.97\columnwidth]{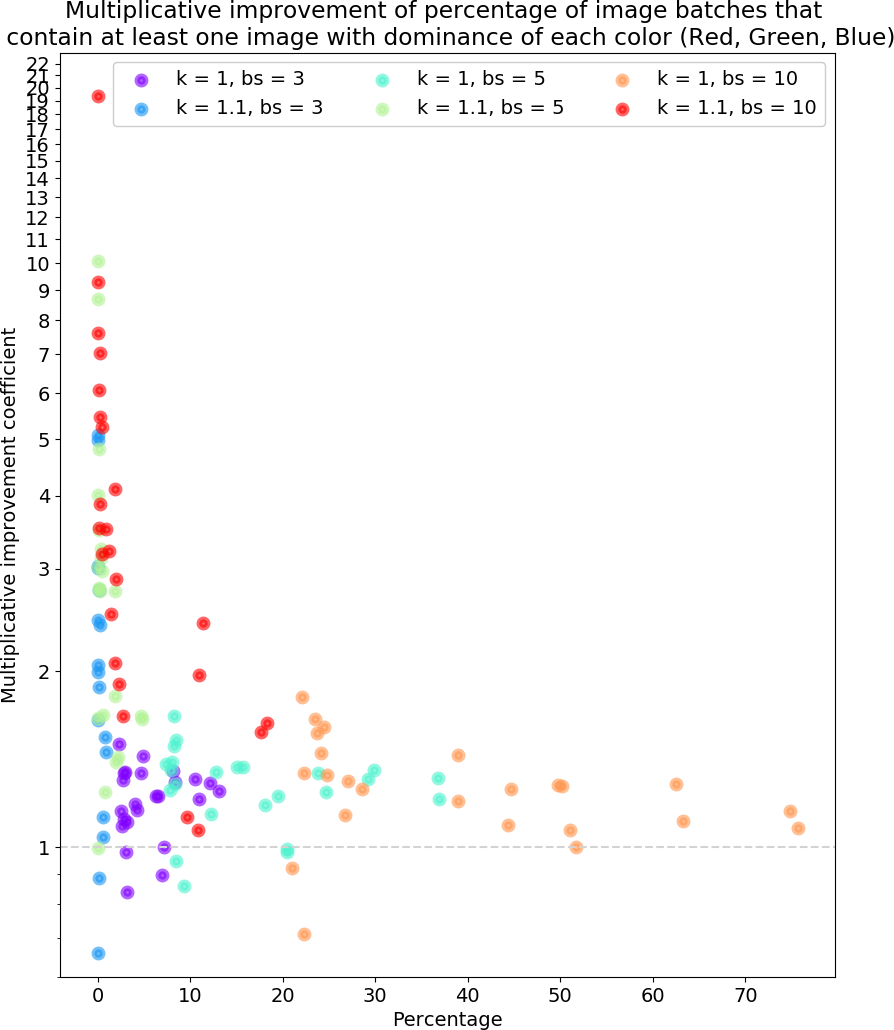} }%
\hfill
\caption{\label{all} Multiplicative improvement of percentage of batches containing images with all 3 dominant colors: pooling\_cap against baseline Stable Diffusion, depending on $K$ as specified in the text and batch {size $bs$. Improvements} are greater on the left, {\em i.e.} more difficult cases.}
\end{center}
\end{figure}

In Figure \ref{all}, we can see that in majority of the cases using pooling\_cap method does not decrease the representation of batches with all $3$ colors dominance. We also can see that we get very high ($> 2.5$) improvement coefficient numbers in cases where color dominance is defined using coefficient k = 1.1, which significantly increases the number of batches featuring this rare characteristic.

\begin{figure}[h!] \begin{center}
\par
{\includegraphics[width=0.97\columnwidth]{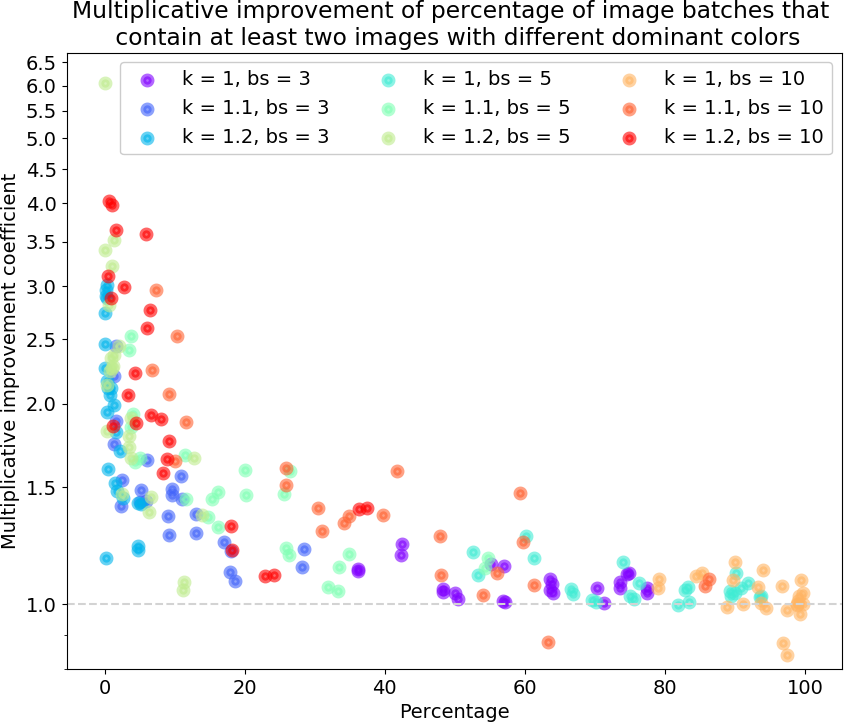} }%
\hfill
\caption{\label{all_2_3} Multiplicative improvement for the percentage of batches containing images of at least 2 of the 3 categories: the improvement is greater for more difficult cases, which are on the left (pooling\_cap against baseline Stable Diffusion method). }
\end{center}
\end{figure}
In Figure \ref{all_2_3}, we can see that in majority of the cases using pooling\_cap method does not decrease the representation of batches featuring at least $2$ dominant colors. Similar to figure \ref{all}, we observe very high ($> 2.5$) improvement coefficient numbers only in cases where color dominance is defined using coefficient $K > 1$, which significantly increases the number of batches featuring this rare characteristic. We also can see that for modes where having at least 2 dominant colors per batch is a rare characteristic (less than $50 \%$ of batches feature it), we always have improvement coefficient $> 1$, thus increasing the color diversity.

\subsection{Gender and ethnicity evaluation}
Here, we evaluate the impact of our proposed method on the diversity of gender and ethnicity representation in the generated image batches.
\begin{figure}[h!] \begin{center}
\par
{\includegraphics[width=0.97\columnwidth]{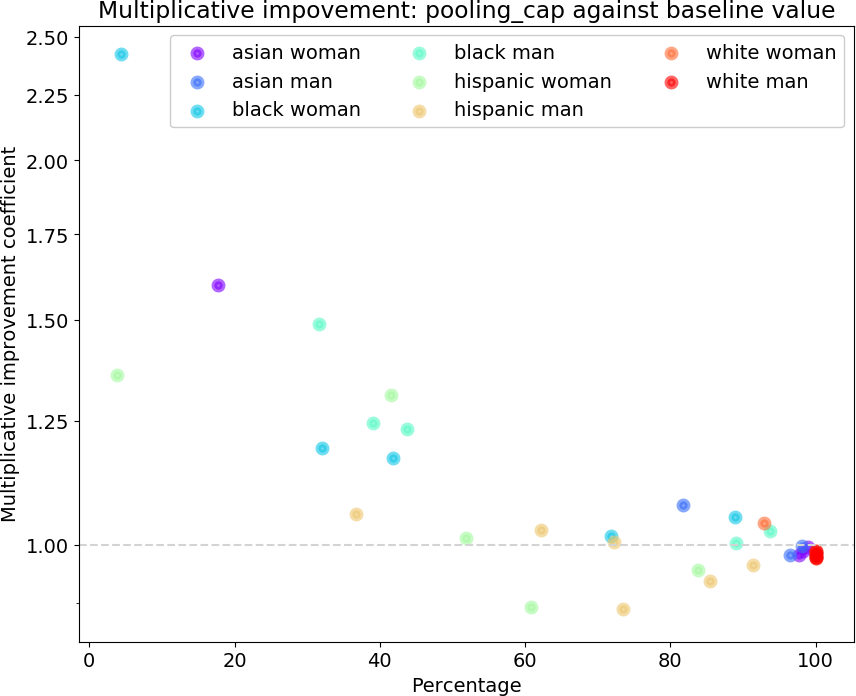} }%
\hfill
\caption{\label{eth_gen} Multiplicative improvement (increase) of percentage of batches containing various ethnicity + gender pairs: we get a greater improvement for more difficult cases (pooling\_cap against baseline Stable Diffusion method).}
\end{center}
\end{figure}

Our results, as illustrated in Figure \ref{eth_gen}, demonstrate a remarkable enhancement in the representation of underrepresented categories. This improvement is consistently observed across all ethnicity/gender pairs that initially appeared in less than 60\% of the batches. Notably, our cap pooling technique has led to a substantial increase in their presence, with certain categories being up to 2.4 times more prevalent than in the standard Stable Diffusion version.

While we see some detrimental results for Hispanics representation, it is important to note that ethnicity identification based solely on a person's image is not always accurate for artificial intelligence methods. In this research, we primarily focus on the diversity between white and black individuals, as they can be visually distinguished with minimal ambiguity \ref{maluleke2022studying}. In Figure  \ref{eth_gen}, we observe a positive improvement in the representation of black individuals, both male and female, compared to the baseline results for Stable Diffusion.

Overall, these findings highlight the effectiveness of our approach in promoting diversity and addressing underrepresentation in gender and ethnicity across various image batches.

\subsection{LPIPS evaluation}
In this subsection, we evaluate the generated image diversity using the LPIPS metric. LPIPS measures the perceptual similarity between images based on deep neural network representations. A lower pairwise LPIPS score indicates greater diversity among the generated images.

Here we compare the performance of different methods, including the baseline and our proposed ``pooling\_cap'' method, across different batch sizes.
\begin{figure}[h!] \begin{center}
\par
{\includegraphics[width=0.95\columnwidth]{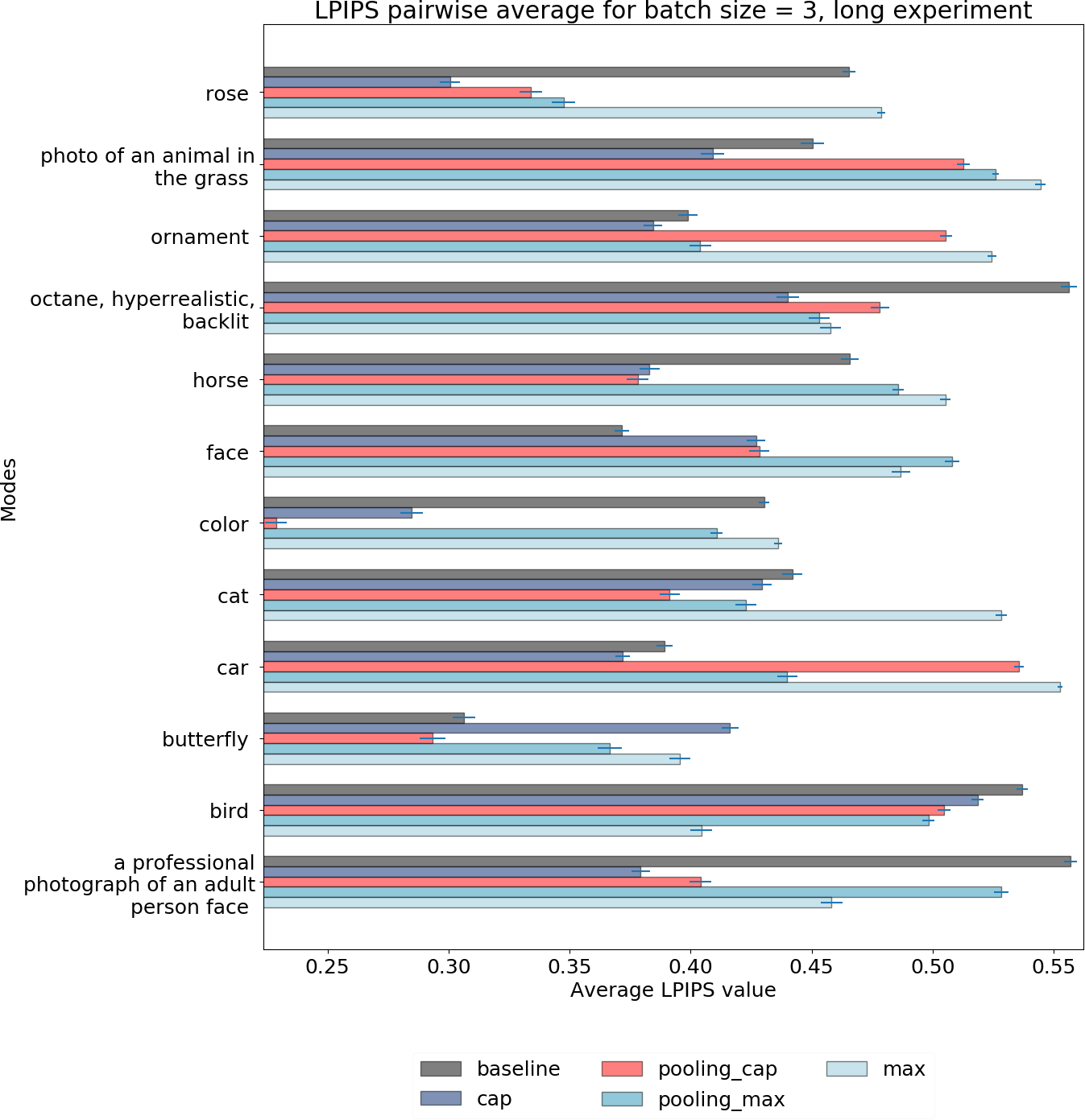} }%
\hfill
\caption{Average batch pairwise LPIPS, batch size=3\label{3_lpips}{: no clear conclusion overall,} {due to the small batch size.}}
\end{center}
\end{figure}

Figure \ref{3_lpips} presents the average batch pairwise LPIPS distance for a batch size of 3. We can see that for most of the experiments ``pooling\_cap'' method proves to be more diverse than the baseline, while for certain cases such as ``face'', baseline outperforms ``pooling\_cap'' . We also notice that ``pooling\_cap'' is not the single method that performs well. Others, such as ``cap'', prove to be the best for various prompts as well. For example: ``rose'', ``ornament'' and ``car''.
\begin{figure}[h!] \begin{center}
\par
{\includegraphics[width=0.97\columnwidth]{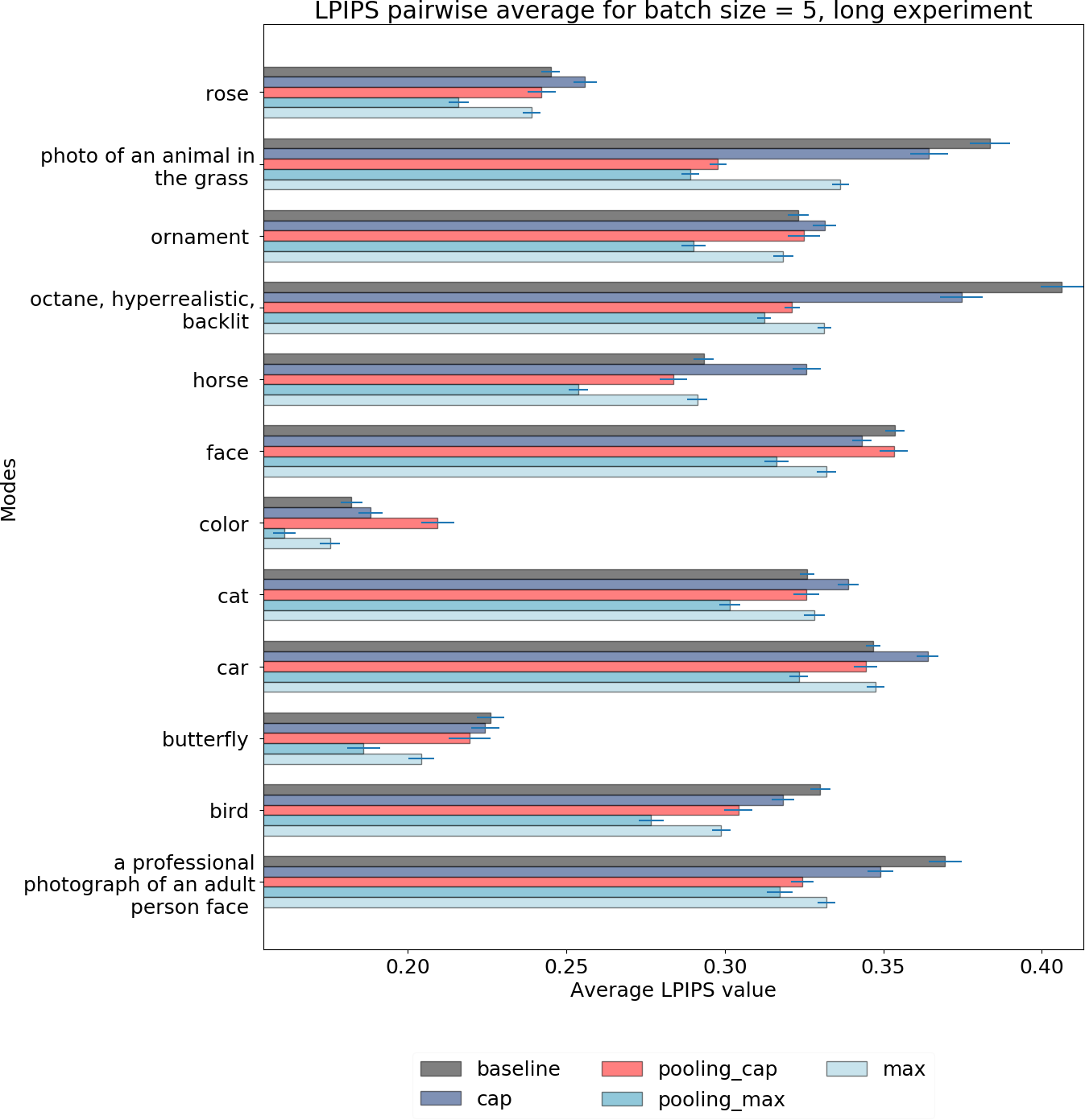} }%
\hfill
\caption{Average batch pairwise LPIPS, batch size=5: \label{5_lpips}{the two Pooling methods are  often better than the baseline, though results are clearer for batch size 50}.}
\end{center}
\end{figure}

Figure \ref{5_lpips} illustrates the average batch pairwise LPIPS distance for a batch size of 5. We can see that for most of the experiments ``pooling\_max'' method proves to be most diverse among all the methods outperforming ``pooling\_cap'', while ``pooling\_cap'' is still better than the baseline for most of the cases. Similarly to Figure \ref{3_lpips}, we can also notice that for the cases where baseline had a small average LPIPS distance (for instance ``octane'' and ``photo of an animal''), all the proposed methods significantly outperform the baseline. This fact shows that even for small batches, our methods provide diversity benefits in cases when it was really lacking.
\begin{figure}[h!] \begin{center}
\par
{\includegraphics[width=0.97\columnwidth]{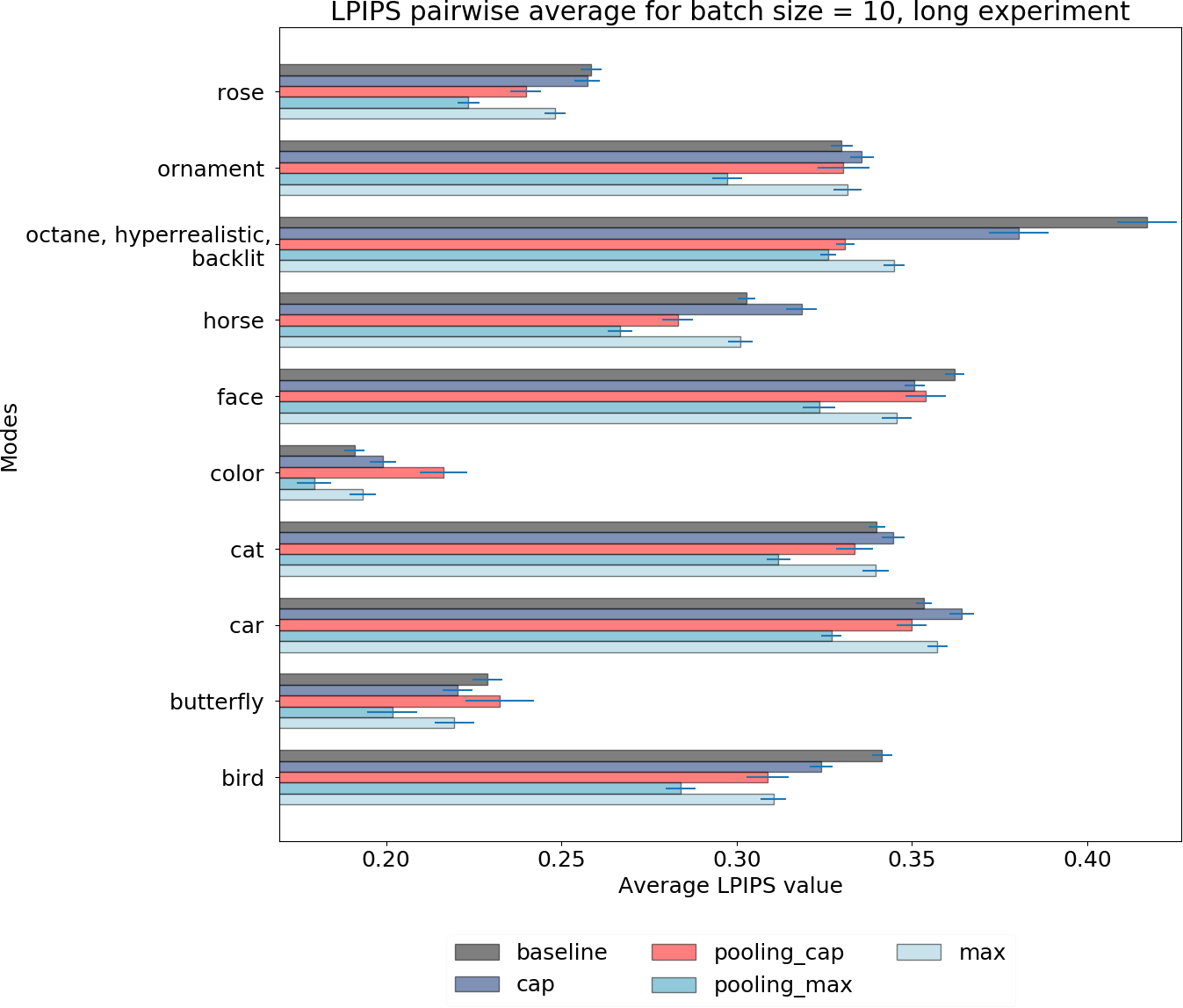} }%
\hfill
\caption{Average batch pairwise LPIPS, batch size=10\label{10_lpips}. {The two Pooling methods are  often better than the baseline, though results are clearer for batch size 50}.}
\end{center}
\end{figure}

Figure \ref{10_lpips} showcases the average batch pairwise LPIPS distance for a batch size of 10. Here, we can see that for most of the experiments ``pooling\_max''  is the most diverse among all the methods, while ``pooling\_cap'' is still better than the baseline for most of the cases. We again notice that in cases where the baseline had a small average LPIPS distance (for instance ``octane''), all of the proposed methods significantly outperform the baseline. This fact shows that even for small batches, our methods provide diversity benefits in cases when it was really lacking. %We also see that, unlike in Figure \ref{3_lpips}, the baseline for the Figure \ref{10_lpips} never outperforms pooling\_max, which shows that for bigger batches our methods provide better diversity guaranties.
{In contrast to the results presented in Figure \ref{3_lpips}, in Figure \ref{10_lpips} ``pooling\_max'' method shows to always outperform the baseline. This observation indicates that our methods provide better diversity guaranties for the bigger batch sizes. }

\begin{figure}[h!] \begin{center}
\par
{\includegraphics[width=0.97\columnwidth]{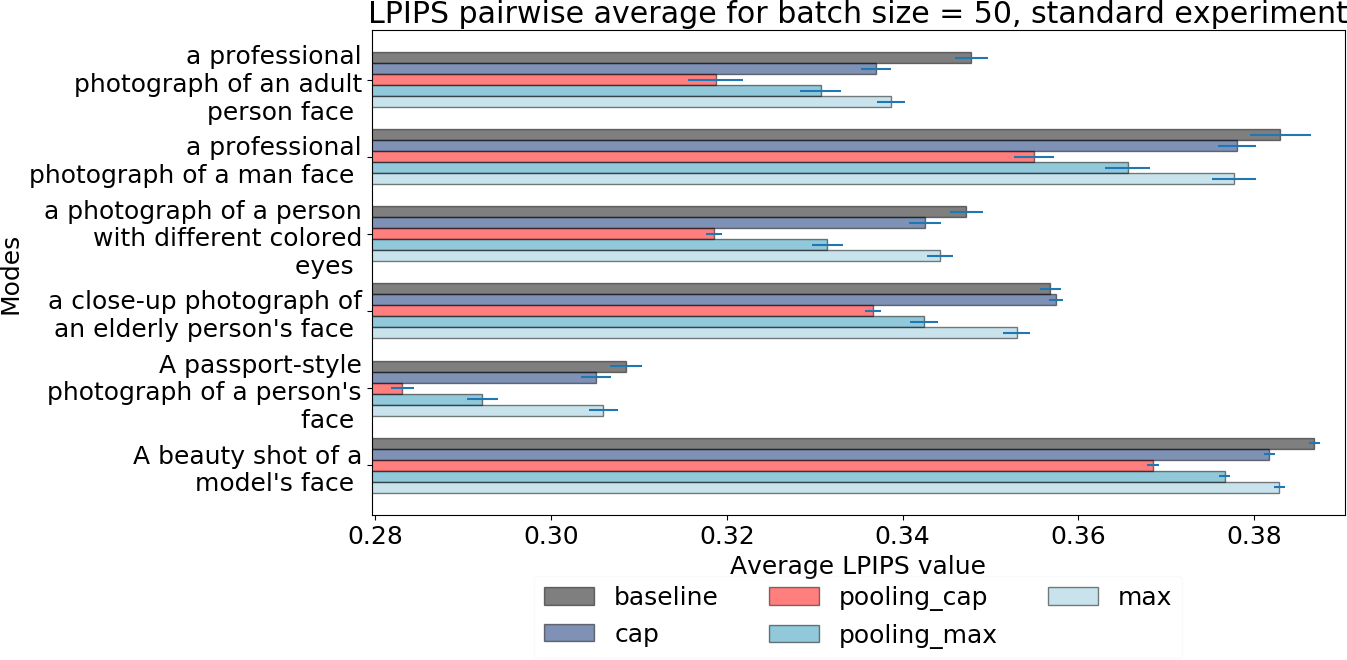} }%
\hfill
\caption{Average batch pairwise LPIPS, batch size=50. For this batch size, {in all cases} pooling methods (both max and cap) are beneficial to diversity as measured by LPIPS. \label{50_lpips}}
\end{center}
\end{figure}

 {In contrast to the results shown in Figures \ref{3_lpips}, \ref{5_lpips} and \ref{10_lpips}, in Figure \ref{50_lpips} we can see that the ``pooling\_cap'' method is consistently highlighted as the most diverse across all experiments.  This emphasizes that larger batch sizes lead to better diversity due to the increased availability of reference points for each new image. This experiment strengthens our earlier observations on color diversity and solidifies the superiority of the ``pooling\_cap'' variant over the standard Stable Diffusion approach in terms of diversity.}

\begin{table}[h!]
    \centering
    \caption{Average pairwise LPIPS value across different batches with the same prompt. \label{lpips_avg}}
    \begin{tabular}{|c|c|}
        \hline
        Method & average LPIPS \\
        \hline
        Baseline & 0.354 $\pm$ 0.004\\
        ENTIGEN & 0.325 $\pm$ 0.004\\
        \textbf{Pooling$\_$cap} & \textbf{0.294} $\pm$ \textbf{0.005}\\
        \hline
    \end{tabular}
\end{table}
For the experiment presented in Table \ref{lpips_avg}, we compute LPIPS distance between several image pairs for the same prompt (not necessarily in the same batch) and then average it between different prompts. We do this experiment across standard Stable Diffusion, our preferred method ``pooling\_cap'' and another diversity method and ENTIGEN \ref{bansal2022well} diversity method applied to the same prompts as in subsection ``Experiments on big batches''. More precisely, to each prompt, we add either ``irrespective of their gender'' or ``irrespective of their color''. We can see that even though our generation is focused mainly on batch setting, our method achieves better diversity both than Stable Diffusion and ENTIGEN, that was specifically designed to eliminate human-centered biases such as gender and ethnicity.

\FloatBarrier

\section{Conclusions}

In conclusion, our contributions include (i) a general  technique that can be applied to existing text-to-image models to increase image diversity, which works in an unsupervised manner with a negligible overhead
and (ii) experiments that showcase the diversity advantages of our proposed approach applied to Stable Diffusion, through classification (both image hue and gender/ethnicity) and LPIPS measurements.

Experimental results highlight the impact of our proposed method on color diversity. By analyzing the multiplicative {improvement in the} batches containing images with dominant colors, we observe that the ``pooling\_cap'' method consistently maintains or increases the representation of batches with all three dominant colors. Additionally, we notice significant {enhancement} when using the color dominance coefficient $K > 1$ ({\em i.e.}, cases in which success is rarer, hence the problem is more difficult). These findings validate the effectiveness of our approach in promoting diverse color compositions in generated images.

Our experimental results consistently demonstrate the superiority of our ``pooling\_max'' and ``pooling\_cap'' methods over the baseline (standard Stable Diffusion) in terms of average pairwise LPIPS distance {within} batches. This evaluation metric serves as a reliable measure of image diversity, and our methods consistently outperform the baseline, showcasing their ability to generate more diverse images. These findings highlight the effectiveness of our approach in expanding the range of image variations and improving overall diversity.

Furthermore, the results show a notable improvement in the representation of underrepresented categories across different ethnicity/gender pairs. The ``pooling\_cap'' method led to a substantial increase in the presence of these categories, with some categories being up to $2.4$ times more {frequent}, compared to the baseline. It is important to acknowledge the inherent limitations of ethnicity identification based solely on visual cues, but our focus on improving the representation of underrepresented categories, particularly between white and black individuals, aligns with the goal of promoting diversity and inclusion in image generation.

%Overall, our contributions highlight the effectiveness of our proposed method in promoting diversity, as evidenced by improved representation of underrepresented categories and enhanced color diversity. The detailed experimental results provide evaluation and validation of our approach's effectiveness.

In the future, we intend to assess the generalizability of our approach by applying it to different text-to-image models. This will allow us to determine the extent to which our technique can be adapted and utilized in various contexts. We also intend to {integrate} our method {into} the Stable Diffusion image generation.
Our ultimate goal is to advance the field of text-to-image generation by pushing the boundaries of diversity and realism, enabling the creation of visually varied and inclusive images for a wide range of applications.

\section*{Acknowledgments}
We are very grateful to Dagstuhl Seminar 23351 for discussing a preliminary version of this paper. We developed a previous version of this work at Meta on artificial data, which was presented at Dagstuhl 23351.

\section*{References}

\begin{enumerate}[label={[\arabic*]}]

\item \label{atanassov2004discrepancy} Emanouil I Atanassov. On the discrepancy of the Halton sequences. \textit{Math. Balkanica (NS)}, 18(1-2):15–32, 2004.
\item \label{bansal2022well} Hritik Bansal, Da Yin, Masoud Monajatipoor, and Kai-Wei Chang. How well can text-to-image generative models understand ethical natural language interventions? \textit{arXiv preprint arXiv:2210.15230}, 2022.
\item \label{gizmodo} Kyle Barr. Ai image generators routinely display gender and cultural bias. In gizmodo.com, 2022.
\item \label{berns2022increasing} Sebastian Berns. Increasing the diversity of deep generative models. In Proceedings of the AAAI Conference on Artificial Intelligence, volume 36, pages 12870–12871, 2022.
\item \label{bianchi2023easily} Federico Bianchi, Pratyusha Kalluri, Esin Durmus, Faisal Ladhak, Myra Cheng, Debora Nozza, Tatsunori Hashimoto, Dan Jurafsky, James Zou, and Aylin Caliskan. Easily accessible text-to-image generation amplifies demographic stereotypes at large scale. In Proceedings of the 2023 ACM Conference on Fairness, Accountability, and Transparency, pages 1493–1504, 2023.
\item \label{chambon2022adapting} Pierre Chambon, Christian Bluethgen, Curtis P Langlotz, and Akshay Chaudhari. Adapting pretrained vision-language foundational models to medical imaging domains. \textit{arXiv preprint arXiv:2210.04133}, 2022.
\item \label{chefer2023hidden} Hila Chefer, Oran Lang, Mor Geva, Volodymyr Polosukhin, Assaf Shocher, Michal Irani, Inbar Mosseri, and Lior Wolf. The hidden language of diffusion models. \textit{arXiv preprint arXiv:2306.00966}, 2023.
\item \label{fraserdiversity} Kathleen C Fraser, Svetlana Kiritchenko, and Isar Nejadgholi. Diversity is not a one-way street: Pilot study on ethical interventions for racial bias in text-to-image systems. ICCV, accepted, 2023.
\item \label{ham2023modulating} Cusuh Ham, James Hays, Jingwan Lu, Krishna Kumar Singh, Zhifei Zhang, and Tobias Hinz. Modulating pretrained diffusion models for multimodal image synthesis. \textit{arXiv preprint arXiv:2302.12764}, 2023.
\item \label{hammersley} J. M. Hammersley. Monte-carlo methods for solving multivariate problems. \textit{Annals of the New York Academy of Sciences}, 86(3):844–874, 1960.
\item \label{cheuk} Cheuk Ting Ho. Stable diffusion: Why are diverse results so hard to come by? In anaconda.com, 2023.
\item \label{huang2022draw} Nisha Huang, Fan Tang, Weiming Dong, and Changsheng Xu. Draw your art dream: Diverse digital art synthesis with multimodal guided diffusion. In Proceedings of the 30th ACM International Conference on Multimedia, pages 1085–1094, 2022.
\item \label{karthik2023if} Shyamgopal Karthik, Karsten Roth, Massimiliano Mancini, and Zeynep Akata. If at first you don’t succeed, try, try again: Faithful diffusion-based text-to-image generation by selection. \textit{arXiv preprint arXiv:2305.13308}, 2023.
\item \label{kim2023stereotyping} Eunji Kim, Siwon Kim, Chaehun Shin, and Sungroh Yoon. De-stereotyping text-to-image models through prompt tuning. ICML 2023 Workshop on Deployable GenerativeAI, accepted, 2023.
\item \label{maluleke2022studying} Vongani H Maluleke, Neerja Thakkar, Tim Brooks, Ethan Weber, Trevor Darrell, Alexei A Efros, Angjoo Kanazawa, and Devin Guillory. Studying bias in GANs through the lens of race. In \textit{Computer Vision–ECCV 2022: 17th European Conference, Tel Aviv, Israel, October 23–27, 2022, Proceedings, Part XIII}, pages 344–360. Springer, 2022.
\item \label{LHS} Michael D. McKay, Richard J. Beckman, and William J. Conover. A Comparison of Three Methods for Selecting Values of Input Variables in the Analysis of Output from a Computer Code. \textit{Technometrics}, 21:239–245, 1979.
\item \label{nie} Harald Niederreiter. Random Number Generation and quasi-Monte Carlo Methods. Society for Industrial and Applied Mathematics, Philadelphia, PA, USA, 1992.
\item \label{pmlr-v139-ramesh21a} Aditya Ramesh, Mikhail Pavlov, Gabriel Goh, Scott Gray, Chelsea Voss, Alec Radford, Mark Chen, and Ilya Sutskever. Zero-shot text-to-image generation. In Marina Meila and Tong Zhang, editors, Proceedings of the 38th International Conference on Machine Learning, volume 139 of Proceedings of Machine Learning Research, pages 8821–8831. PMLR, 18–24 Jul 2021.
\item \label{rombach2022high} Robin Rombach, Andreas Blattmann, Dominik Lorenz, Patrick Esser, and Björn Ommer. High-resolution image synthesis with latent diffusion models. In Proceedings of the IEEE/CVF Conference on Computer Vision and Pattern Recognition, pages 10684–10695, 2022.
\item \label{samuel2023norm} Dvir Samuel, Rami Ben-Ari, Nir Darshan, Haggai Maron, and Gal Chechik. Norm-guided latent space exploration for text-to-image generation. \textit{arXiv preprint arXiv:2306.08687}, 2023.
\item \label{sha2022fake} Zeyang Sha, Zheng Li, Ning Yu, and Yang Zhang. De-fake: Detection and attribution of fake images generated by text-to-image diffusion models. \textit{arXiv preprint arXiv:2210.06998}, 2022.
\item \label{shipard2023diversity} Jordan Shipard, Arnold Wiliem, Kien Nguyen Thanh, Wei Xiang, and Clinton Fookes. Diversity is definitely needed: Improving model-agnostic zero-shot classification via stable diffusion. In Proceedings of the IEEE/CVF Conference on Computer Vision and Pattern Recognition, pages 769–778, 2023.
\item \label{smith2023balancing} Brandon Smith, Miguel Farinha, Siobhan Mackenzie Hall, Hannah Rose Kirk, Aleksandar Shtedritski, and Max Bain. Balancing the picture: Debiasing vision-language datasets with synthetic contrast sets. \textit{arXiv preprint arXiv:2305.15407}, 2023.
\item \label{somepalli2022diffusion} Gowthami Somepalli, Vasu Singla, Micah Goldblum, Jonas Geiping, and Tom Goldstein. Diffusion art or digital forgery? investigating data replication in diffusion models. \textit{arXiv preprint arXiv:2212.03860}, 2022.
\item \label{struppek2022biased} Lukas Struppek, Dominik Hintersdorf, and Kristian Kersting. The biased artist: Exploiting cultural biases via homoglyphs in text-guided image generation models. \textit{arXiv preprint arXiv:2209.08891}, 2022.
\item \label{taigman2014deepface} Yaniv Taigman, Ming Yang, Marc’Aurelio Ranzato, and Lior Wolf. Deepface: Closing the gap to human-level performance in face verification. In Proceedings of the IEEE conference on computer vision and pattern recognition, pages 1701–1708, 2014.
\item \label{theron2023evidence} Danie Theron. Evidence of unfair bias across gender, skin tones \& intersectional groups in generated images from stable diffusion. In towardsdatascience.com, 2023.
\item \label{xu2018diversity} Jingjing Xu, Xuancheng Ren, Junyang Lin, and Xu Sun. Diversity-promoting GAN: A cross-entropy based generative adversarial network for diversified text generation. In Proceedings of the 2018 conference on empirical methods in natural language processing, pages 3940–3949, 2018.
\item \label{zhang2018unreasonable} Richard Zhang, Phillip Isola, and Alexei A Efros. The unreasonable effectiveness of deep features as a perceptual metric. In Proceedings of the IEEE Conference on Computer Vision and Pattern Recognition, pages 586–595, 2018.
\end{enumerate}

\newpage

\section{Appendix -- Diverse Diffusion: Enhancing Image Diversity in Text-to-Image Generation}

\section{Additional results for color diversity}

Here, we provide additional experimental results to further illustrate the color diversity improvement achieved by our proposed approach. The results are organized based on different coefficient values ($K$) and batch sizes. The figures presented below showcase the multiplicative improvement of batches featuring certain color dominance characteristics using the ``pooling\_cap''  and ``pooling\_max'' methods compared to the baseline Stable Diffusion.

\subsection{K=1}

\textbf{Batch size = 3}

\begin{figure*}[h!]
\centering
\includegraphics[width=0.79\textwidth]{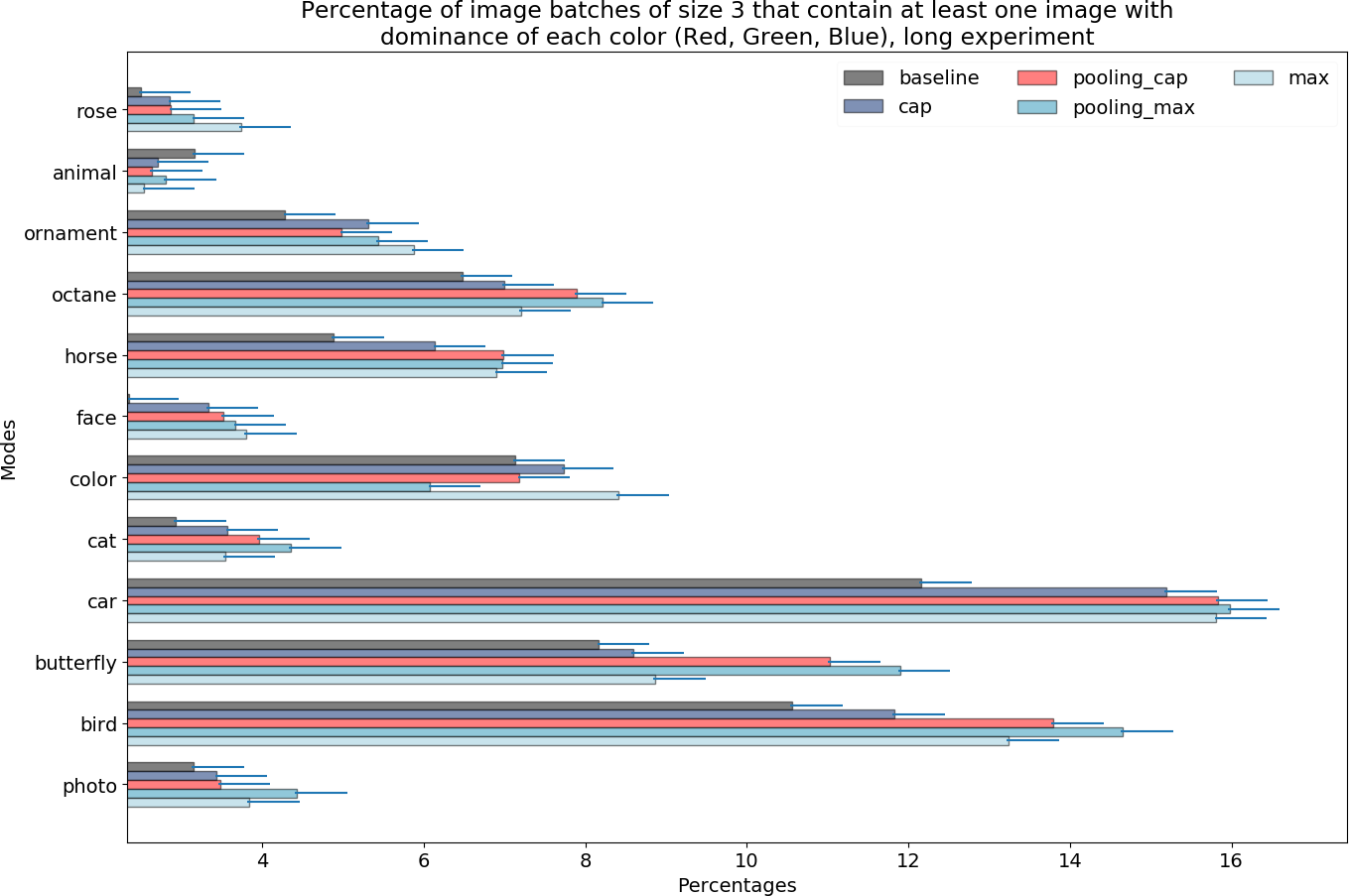}
\includegraphics[width=0.79\textwidth]{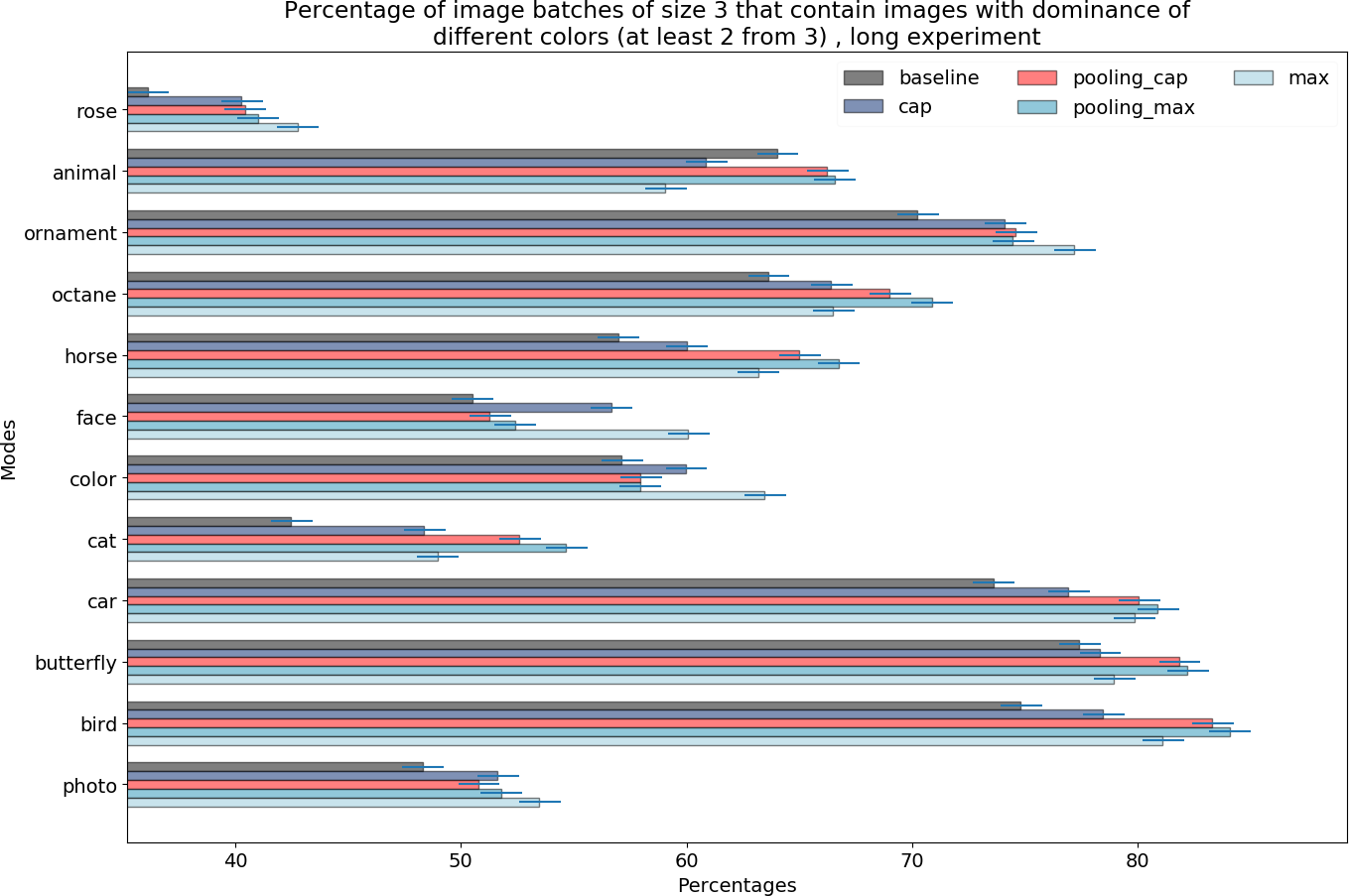}
\caption{Comparison of different modes for various prompts  in regard to the percentage of batches containing images with different dominant colors for the following parameters:   $K=1$, batch size =  3, long experiment}
\label{k_1_3}
\end{figure*}

\begin{figure*}[h!]
\centering
\includegraphics[width=0.79\textwidth]{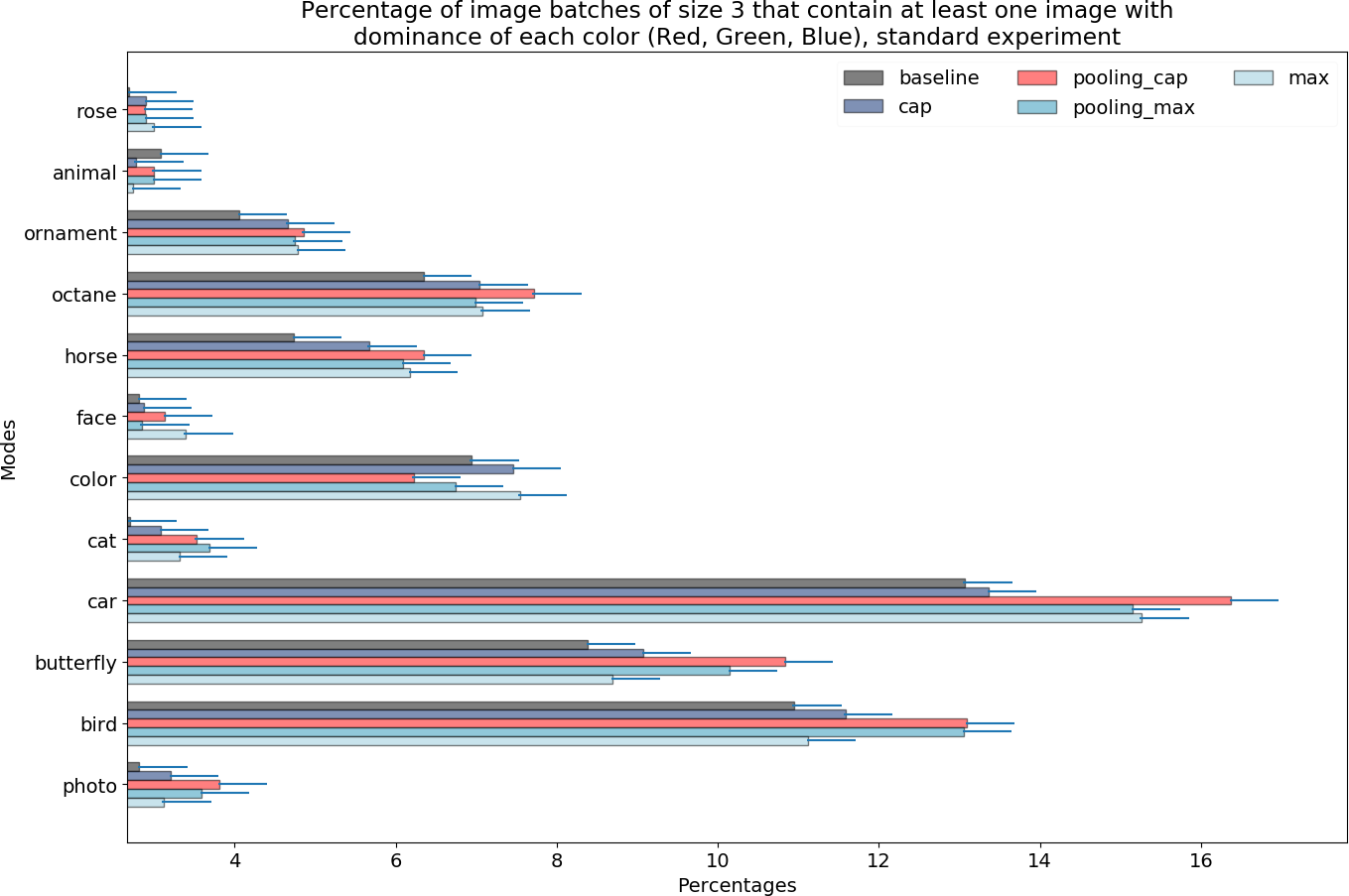}
\includegraphics[width=0.79\textwidth]{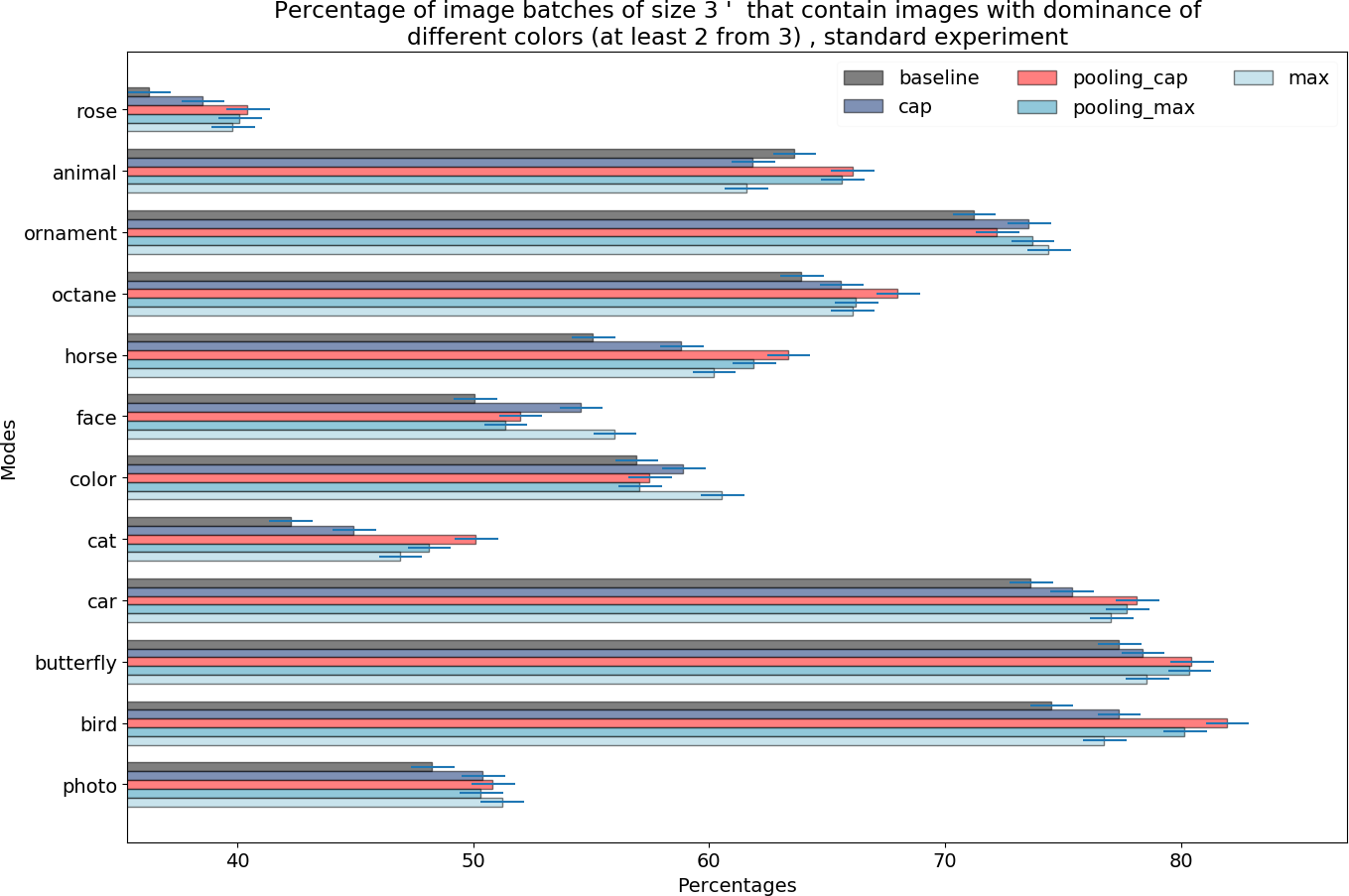}
\caption{Comparison of different modes for various prompts  in regard to the percentage of batches containing images with different dominant colors for the following parameters:   $K=1$, batch size =  3, standard experiment}
\label{k_1_3_standard}
\end{figure*}

Figures \ref{k_1_3} and \ref{k_1_3_standard} display the results for the $K=1$ coefficient and a batch size of 3 in both long and standard experiments.

\textbf{Batch size = 5}

\begin{figure*}[h!]
\centering
\includegraphics[width=0.79\textwidth]{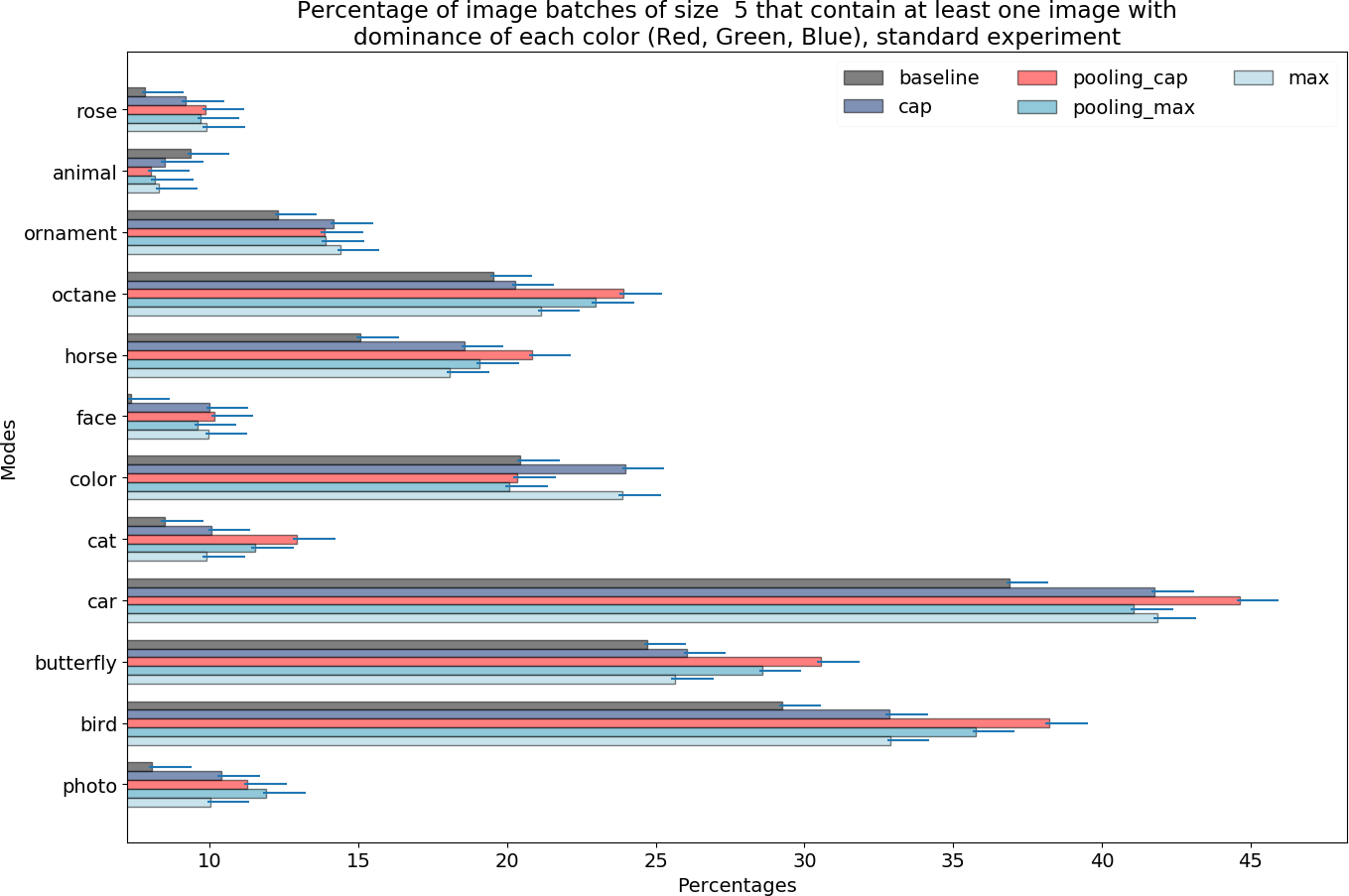}
\includegraphics[width=0.79\textwidth]{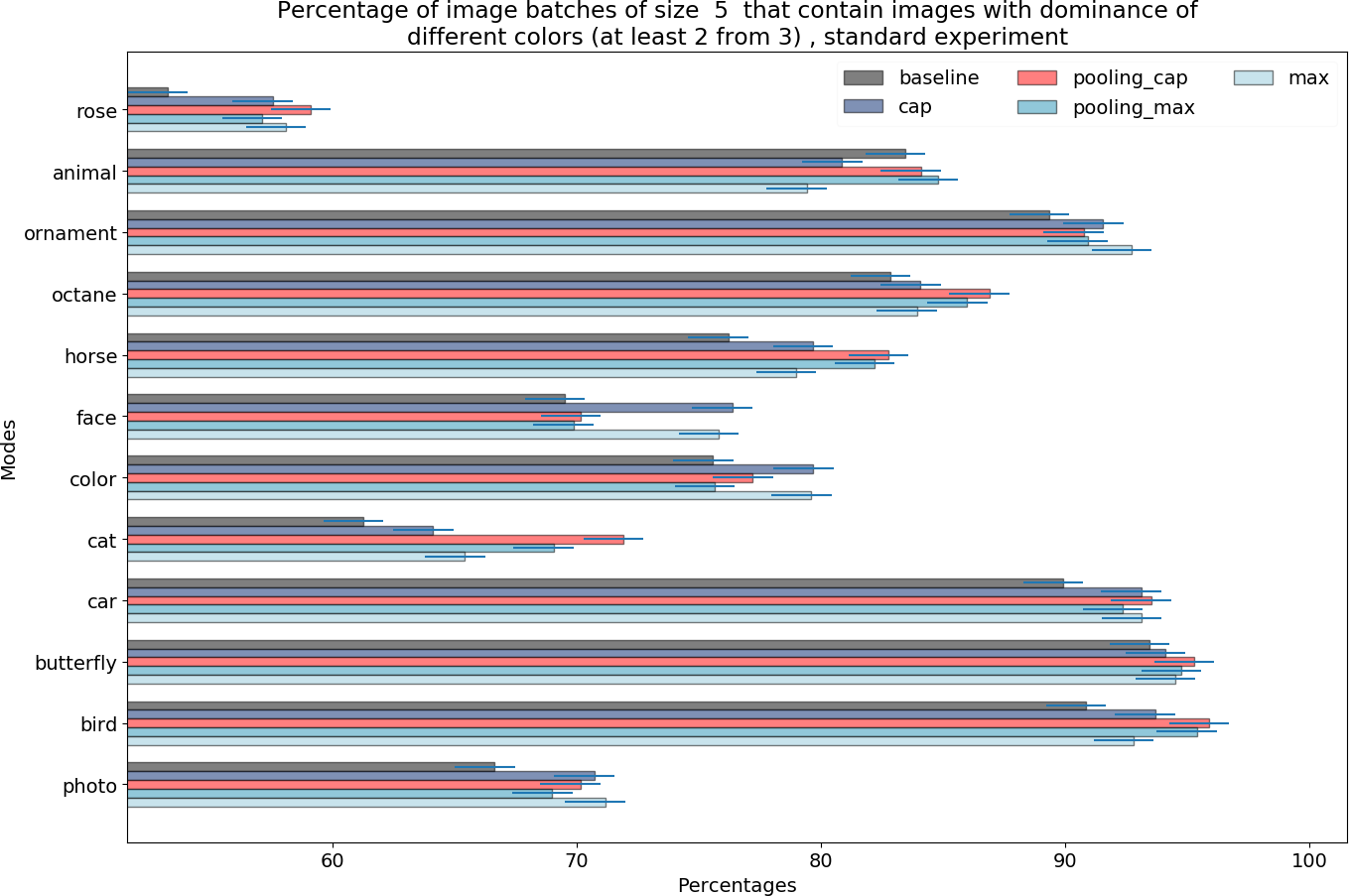}
\caption{Multiplicative improvement of batches containing all 3 dominant colors with $K=1$, batch size =  5, standard experiment}
\label{k_1_5_standard}
\end{figure*}

Figures \ref{k_1_3} and \ref{k_1_5_standard} present the results for the $K=1$ coefficient and a batch size of 5 in both long and standard experiments.

\textbf{Batch size = 10}

\begin{figure*}[h!]
\centering
\includegraphics[width=0.79\textwidth]{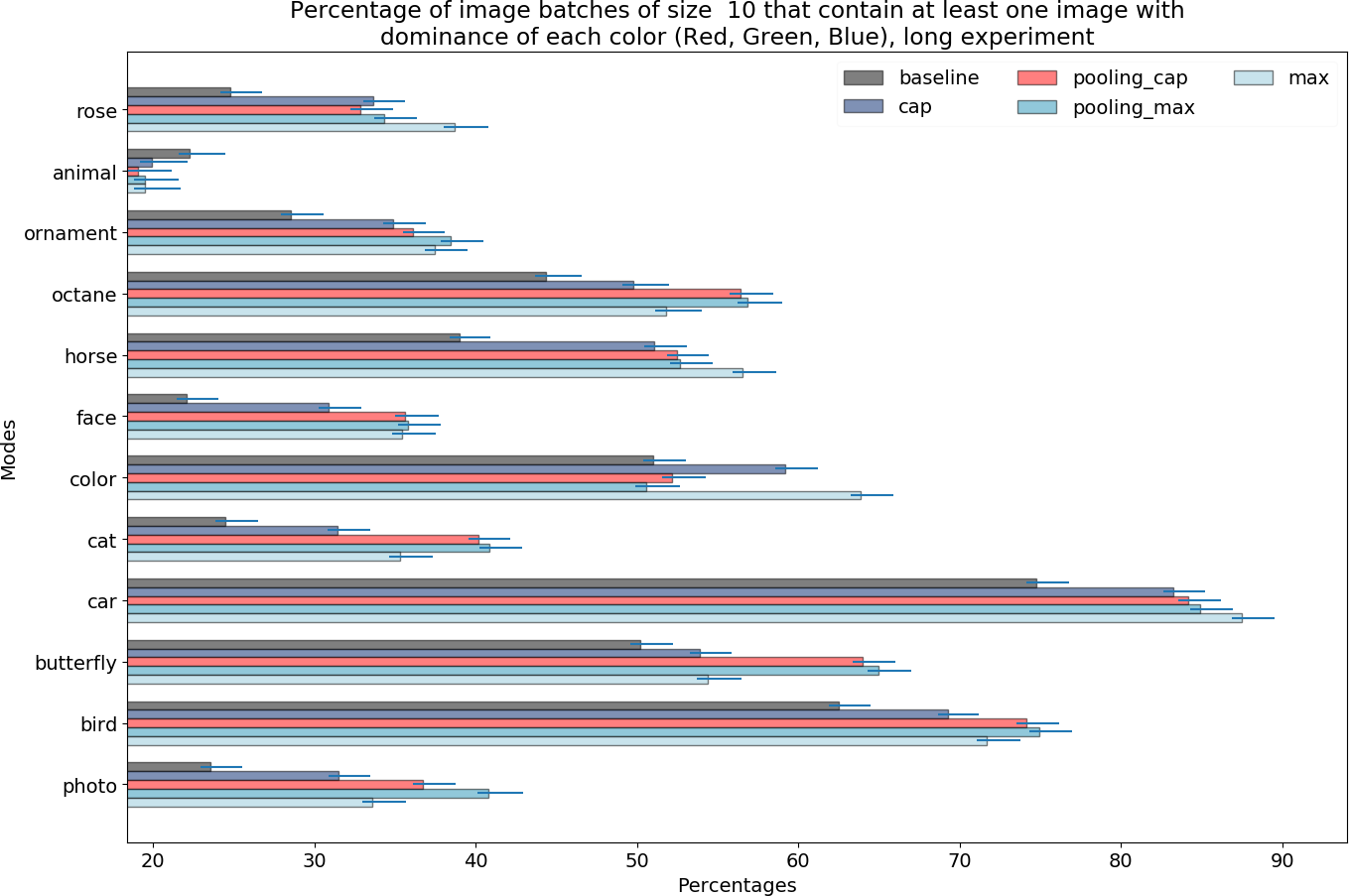}
\includegraphics[width=0.79\textwidth]{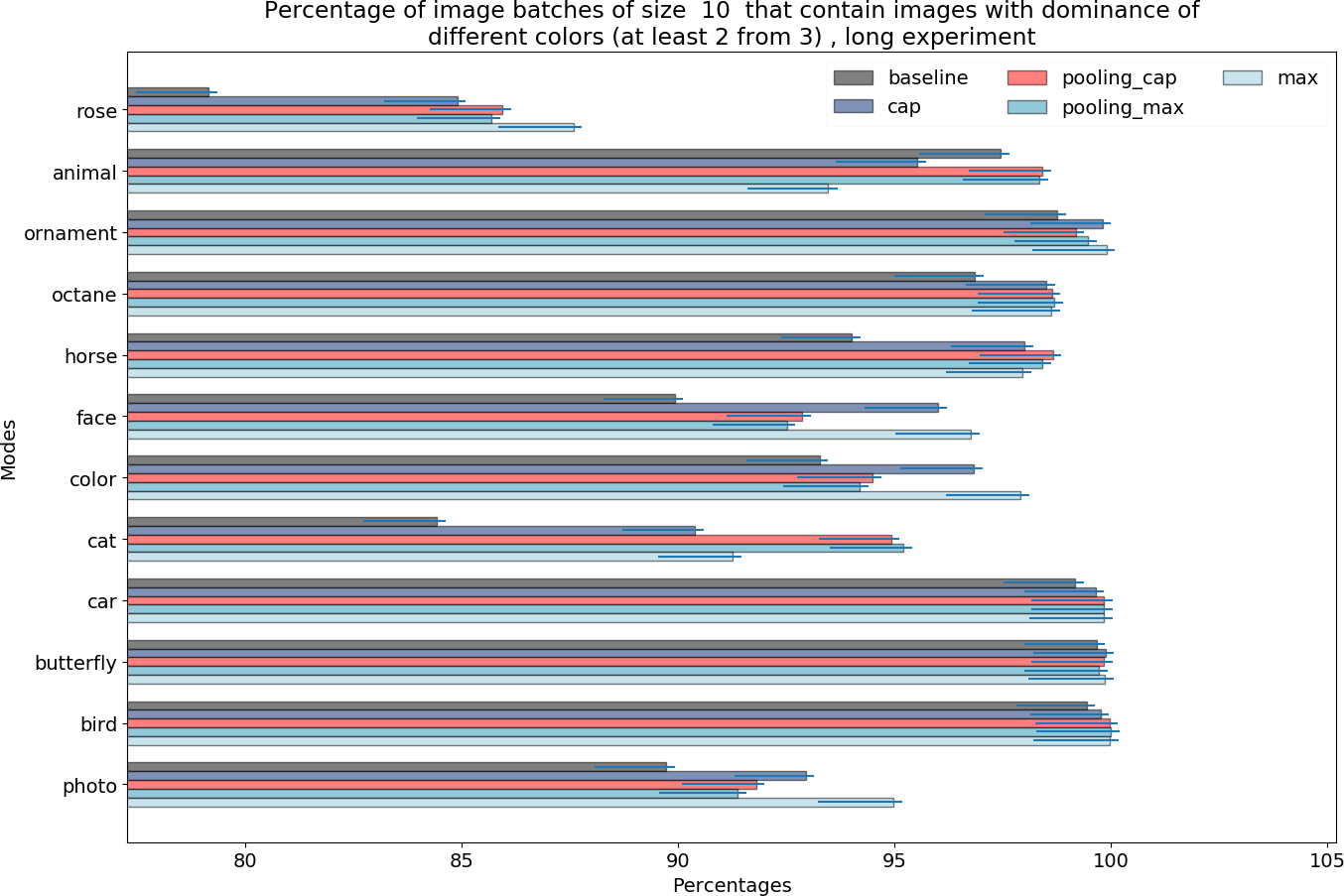}
\caption{Comparison of different modes for various prompts  in regard to the percentage of batches containing images with different dominant colors for the following parameters:   $K=1$, batch size =  10, long experiment}
\label{k_1_10}
\end{figure*}

\begin{figure*}[h!]
\centering
\includegraphics[width=0.79\textwidth]{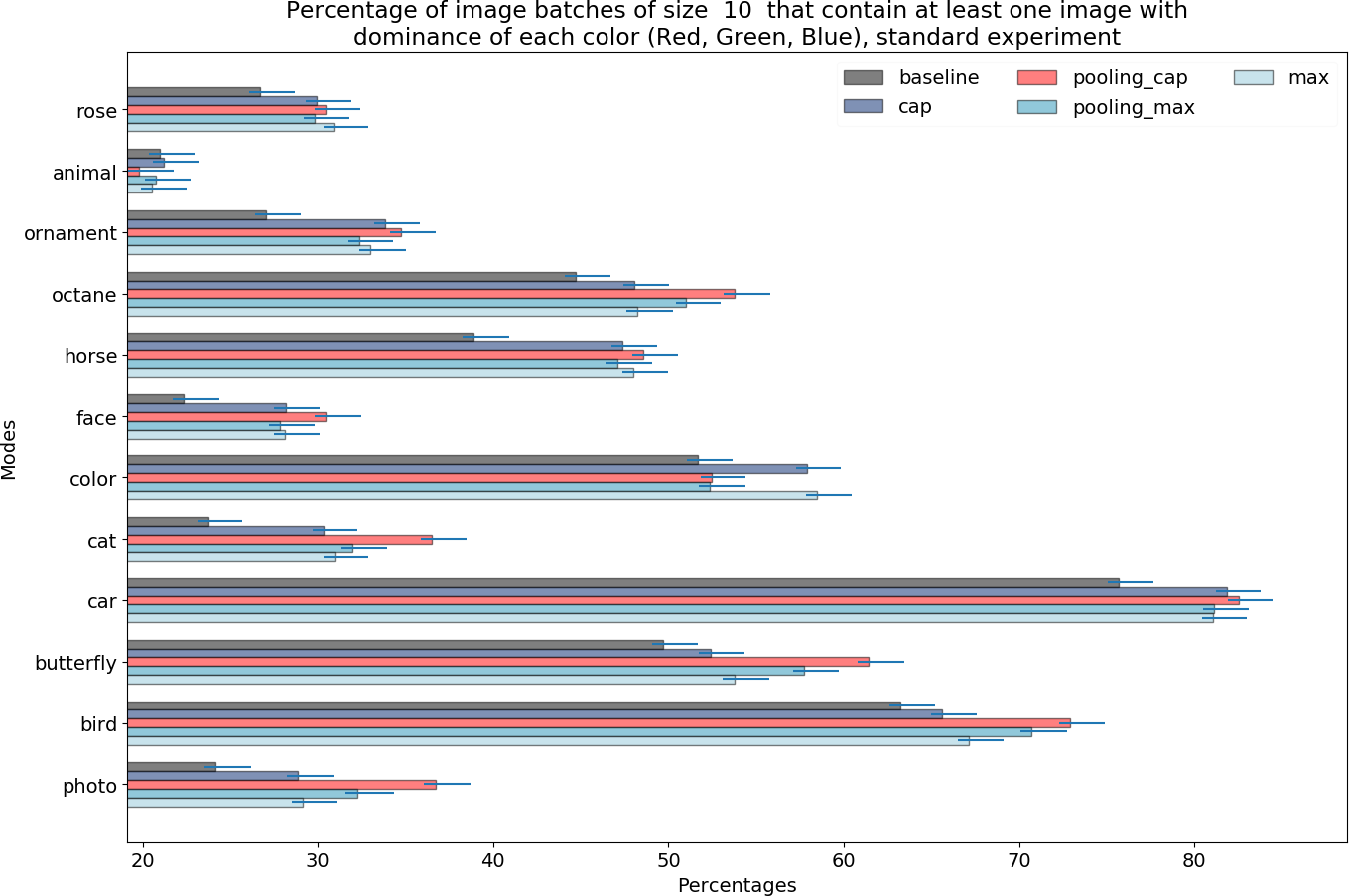}
\includegraphics[width=0.79\textwidth]{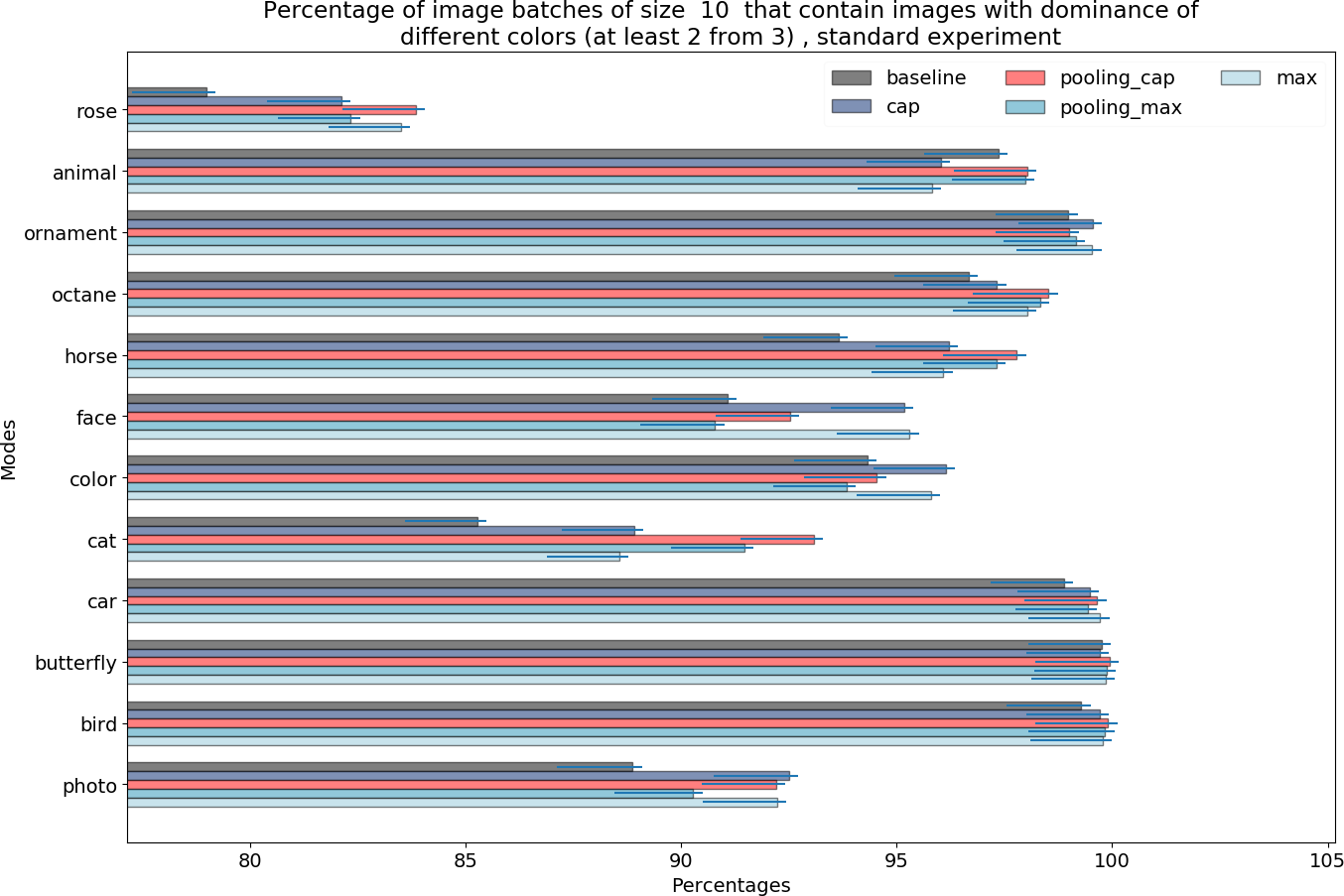}
\caption{Comparison of different modes for various prompts  in regard to the percentage of batches containing images with different dominant colors for the following parameters:   $K=1$, batch size =  10, standard experiment}
\label{k_1_10_standard}
\end{figure*}

Figures \ref{k_1_10} and \ref{k_1_10_standard} illustrate the results for the $K=1$ coefficient and a batch size of 10 in both long and standard experiments.

\textbf{Batch size = 50}

\begin{figure*}[h!]
\centering
\includegraphics[width=0.79\textwidth]{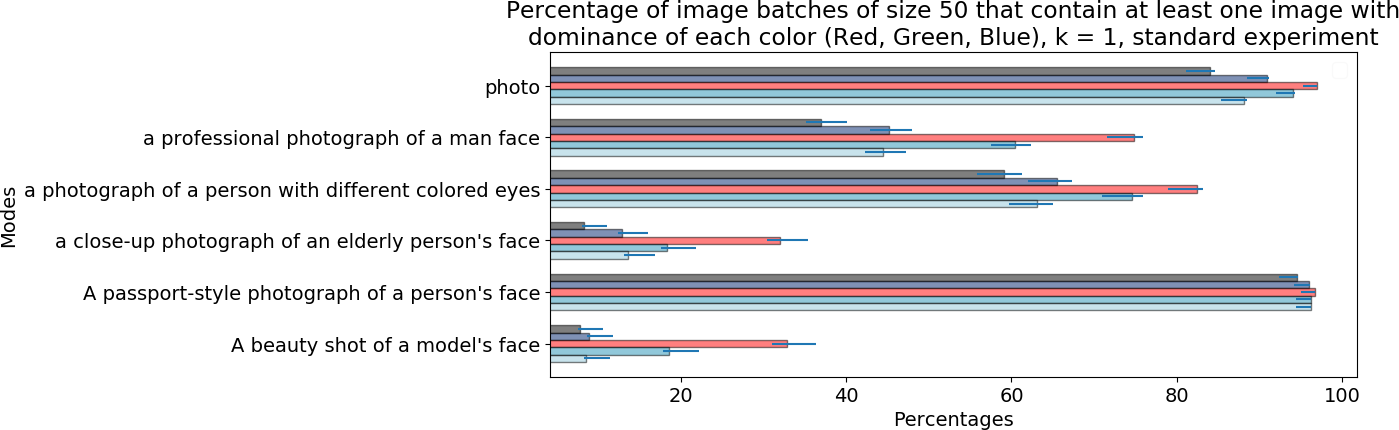}
\includegraphics[width=0.79\textwidth]{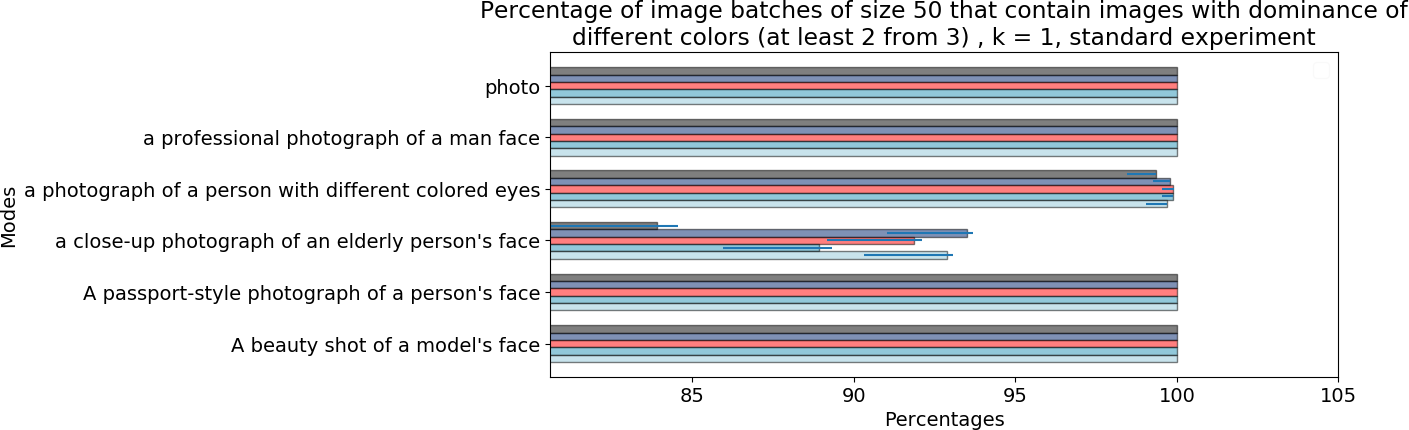}
\caption{Comparison of different modes for various prompts  in regard to the percentage of batches containing images with different dominant colors for the following parameters:   $K=1$, batch size = 50, standard experiment}
\label{k_1_50}
\end{figure*}

Figure \ref{k_1_50} illustrates the results for the $K=1$ coefficient and a batch size of 50 in the standard experiment.

\subsection{K=1.1}

\textbf{Batch size = 3}

\begin{figure*}[h!]
\centering
\includegraphics[width=0.79\textwidth]{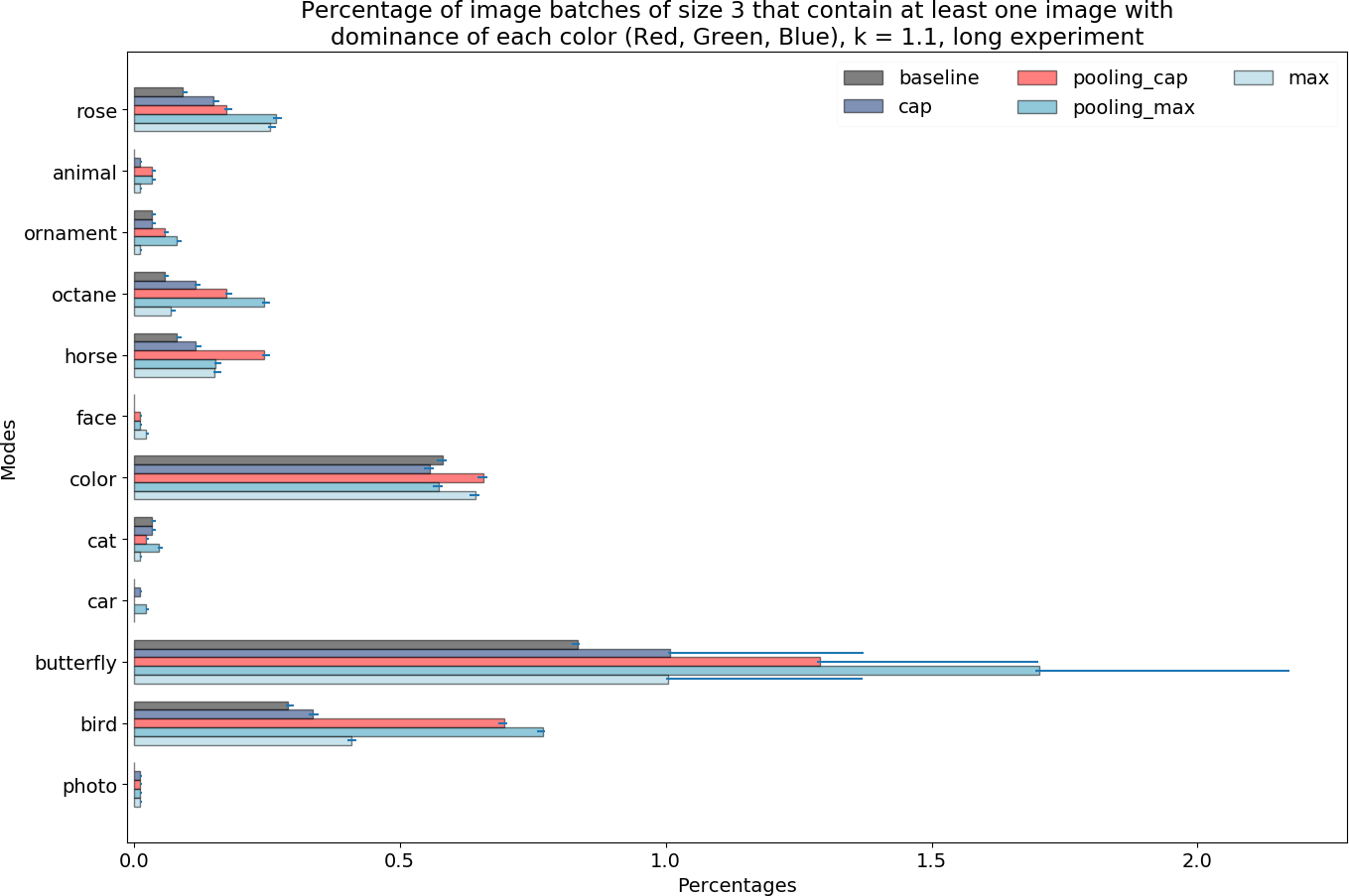}
\includegraphics[width=0.79\textwidth]{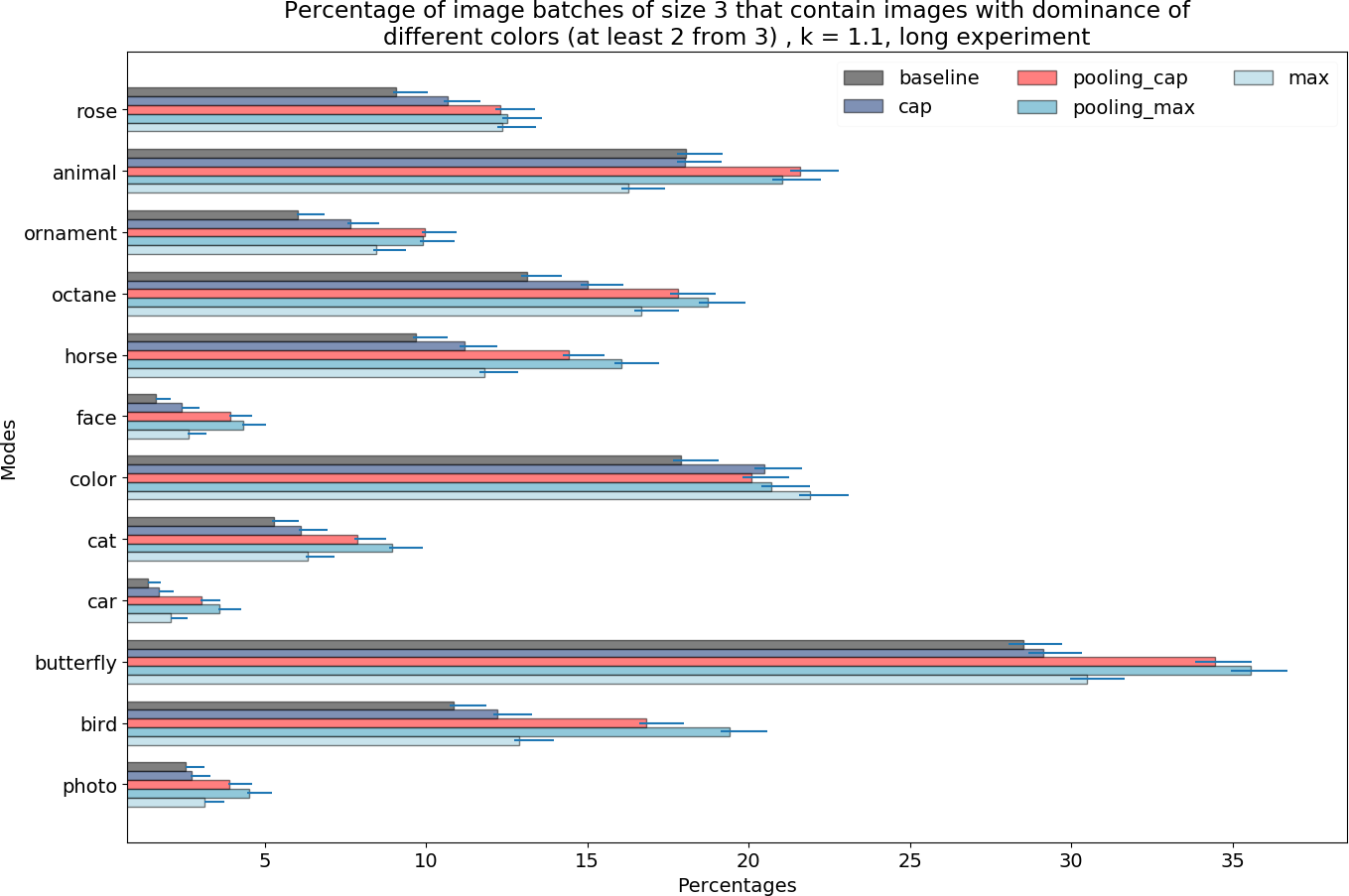}
\caption{Comparison of different modes for various prompts  in regard to the percentage of batches containing images with different dominant colors for the following parameters:   $K=1.1$, batch size =  3, long experiment}
\label{k_11_3}
\end{figure*}

\begin{figure*}[h!]
\centering
\includegraphics[width=0.79\textwidth]{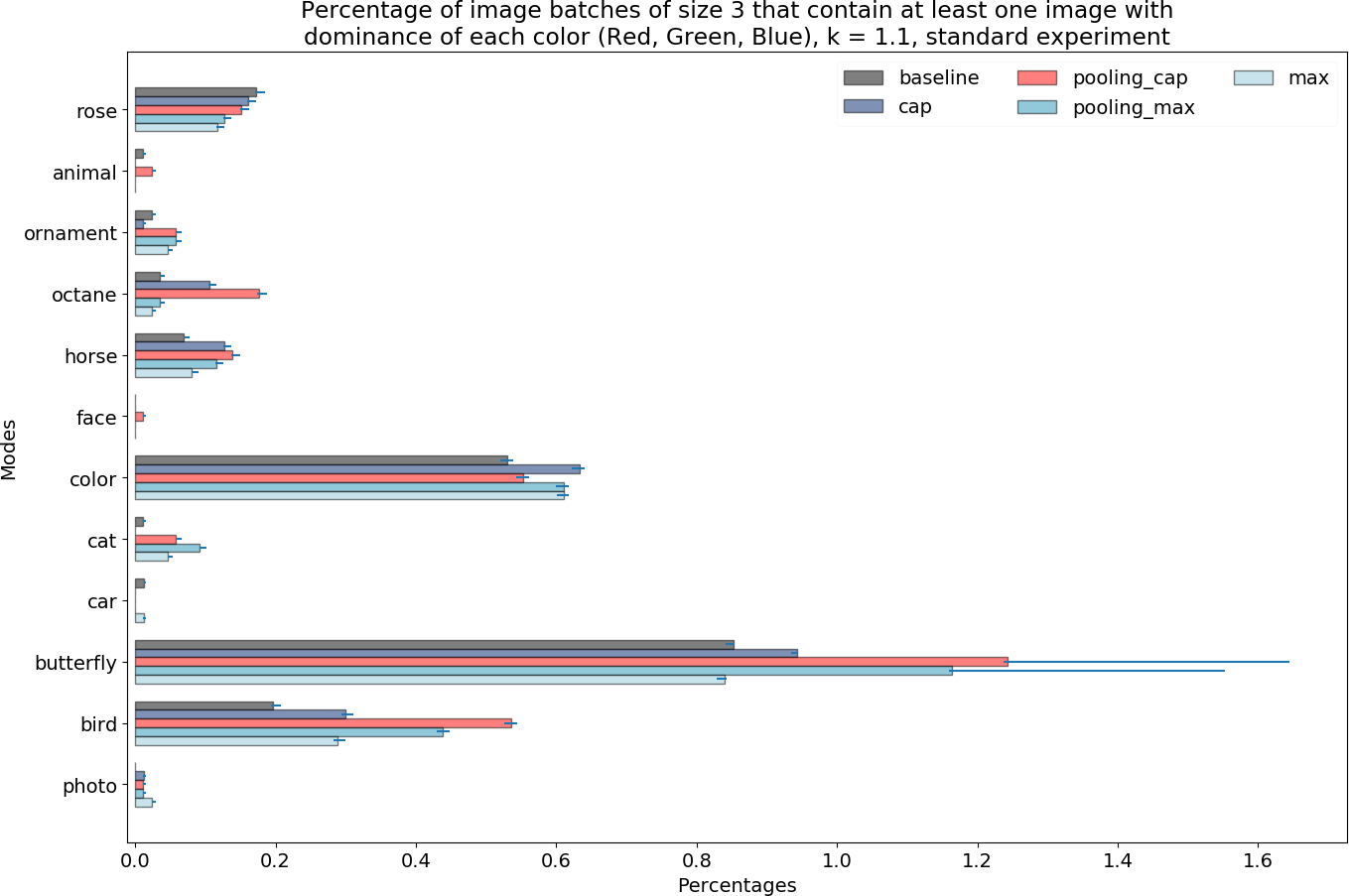}
\includegraphics[width=0.79\textwidth]{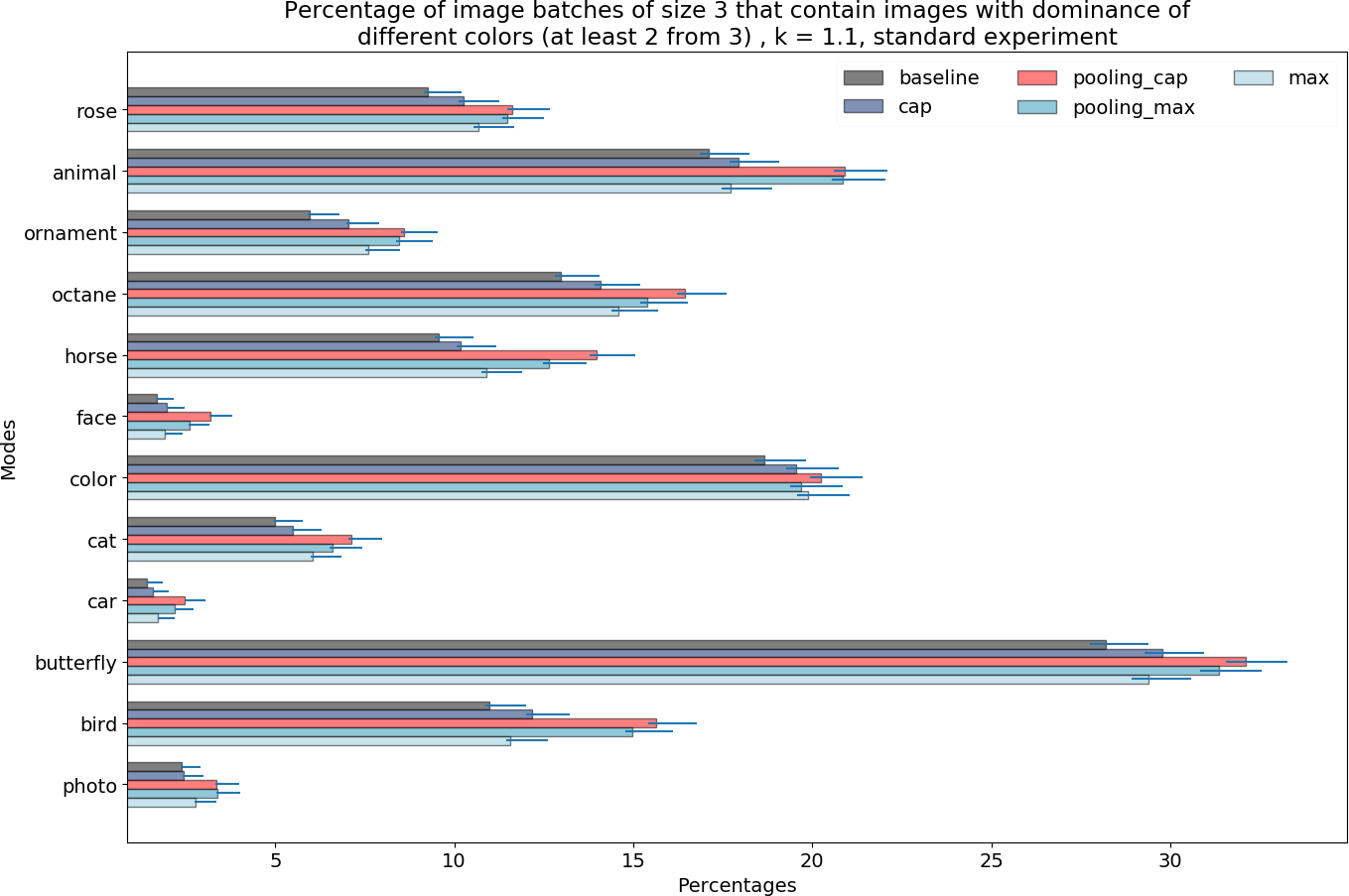}
\caption{Comparison of different modes for various prompts  in regard to the percentage of batches containing images with different dominant colors for the following parameters:   $K=1.1$, batch size =  3, standard experiment}
\label{k_11_3_standard}
\end{figure*}

Figures \ref{k_11_3} and \ref{k_11_3_standard} display the results for the $K=1.1$ coefficient and a batch size of 3 in both long and standard experiments.

\textbf{Batch size = 5}

\begin{figure*}[h!]
\centering
\includegraphics[width=0.79\textwidth]{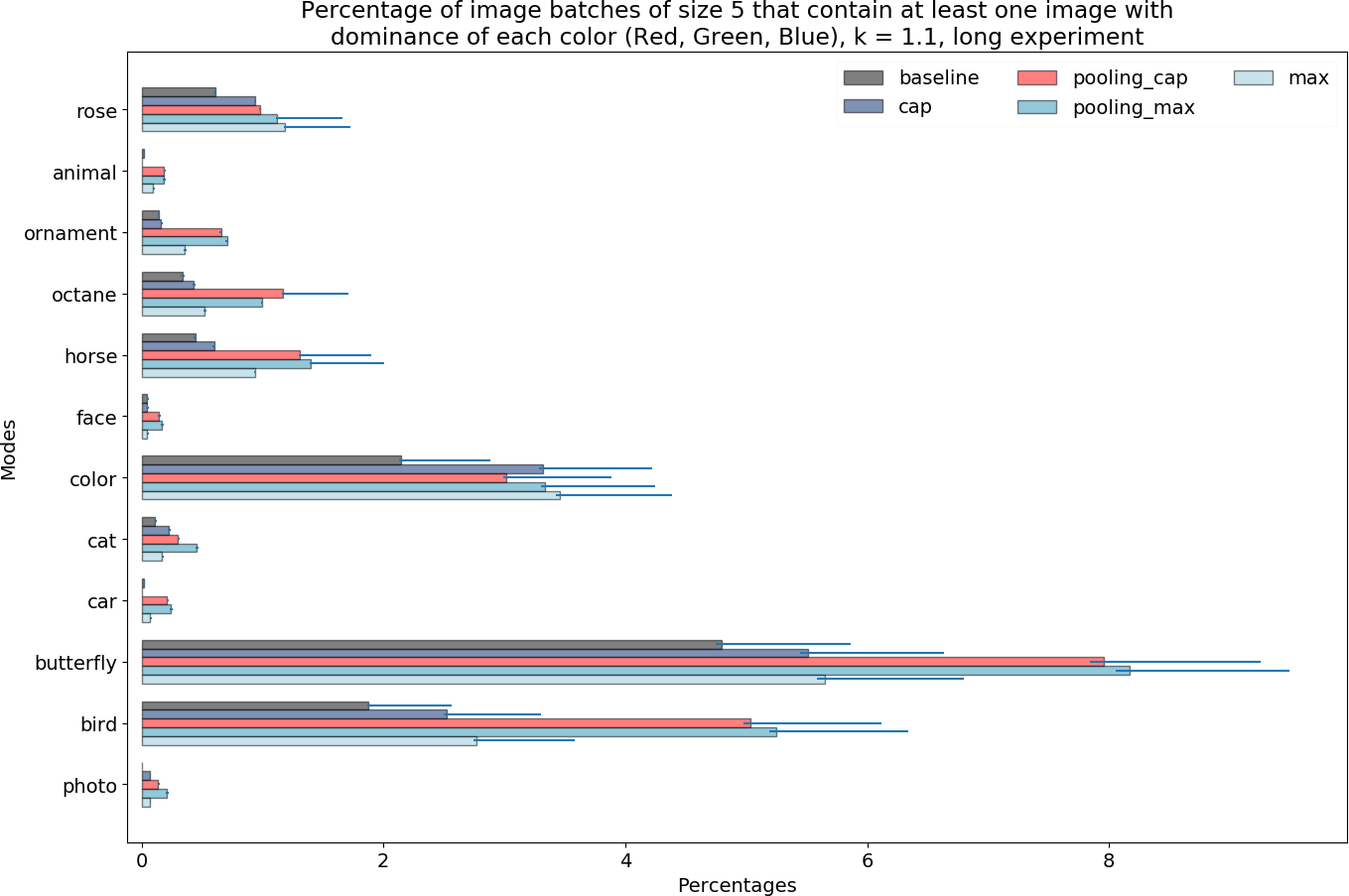}
\includegraphics[width=0.79\textwidth]{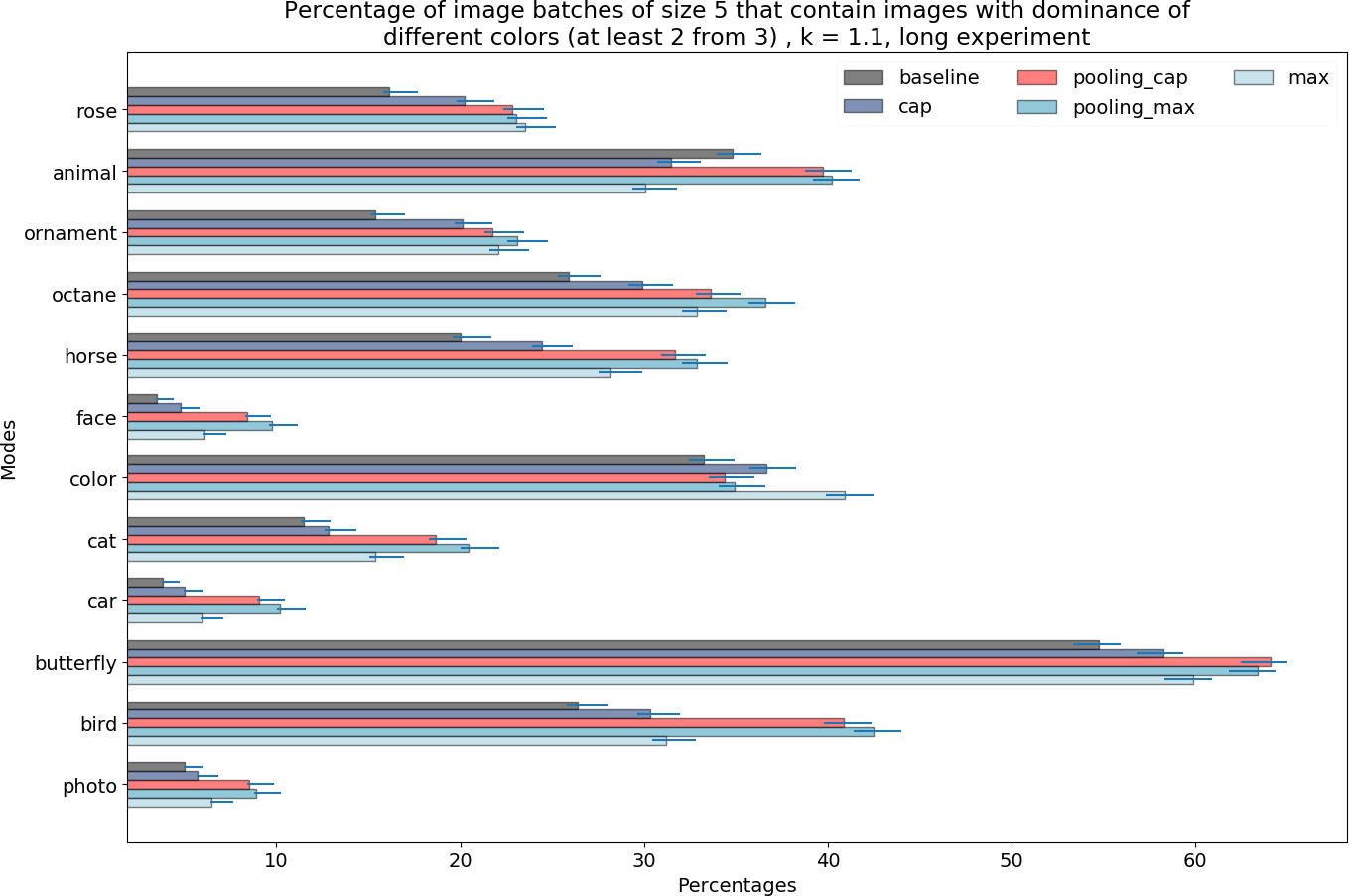}
\caption{Comparison of different modes for various prompts  in regard to the percentage of batches containing images with different dominant colors for the following parameters:   $K=1.1$, batch size =  5, long experiment}
\label{k_11_5}
\end{figure*}

\begin{figure*}[h!]
\centering
\includegraphics[width=0.79\textwidth]{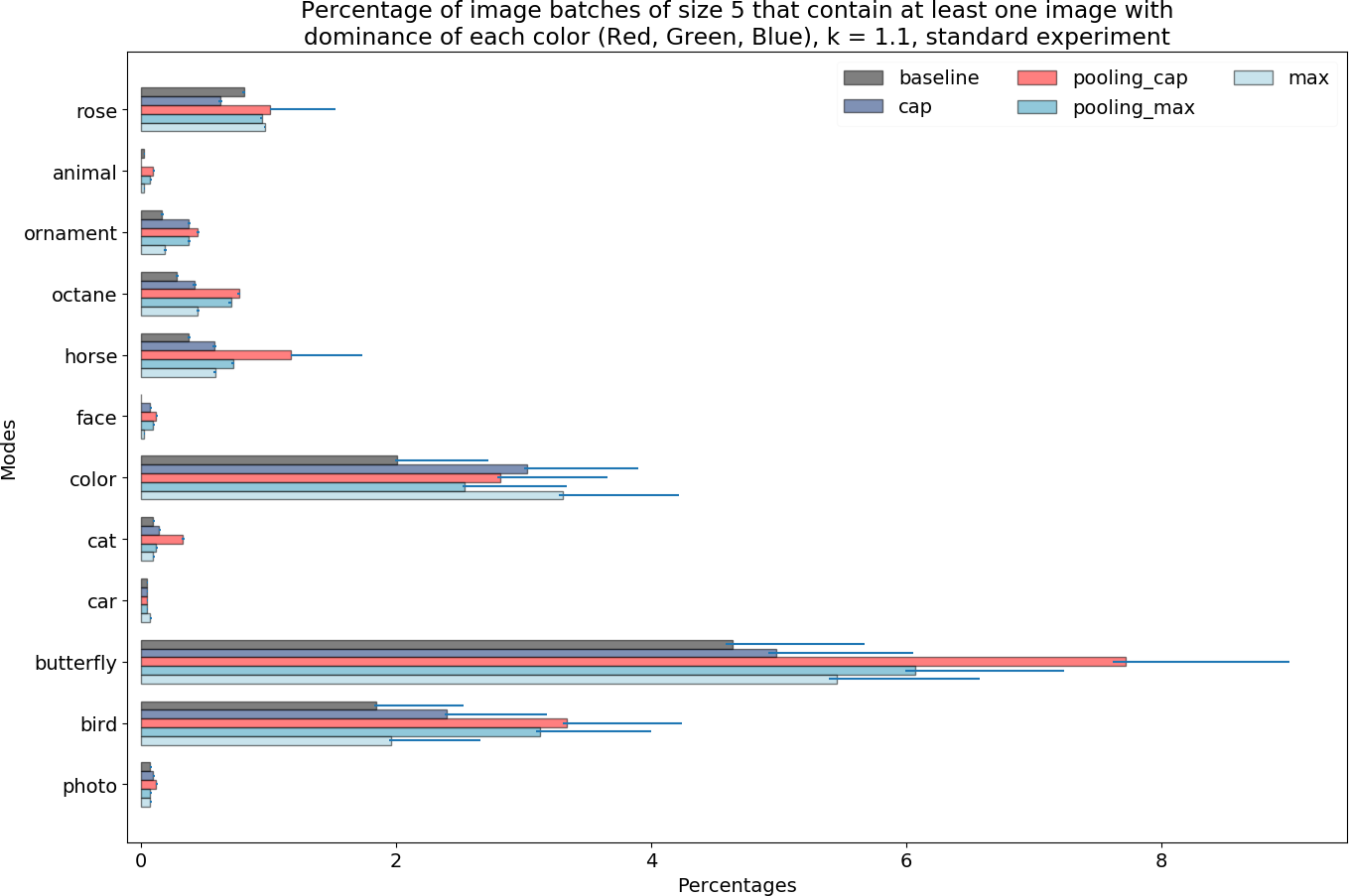}
\includegraphics[width=0.79\textwidth]{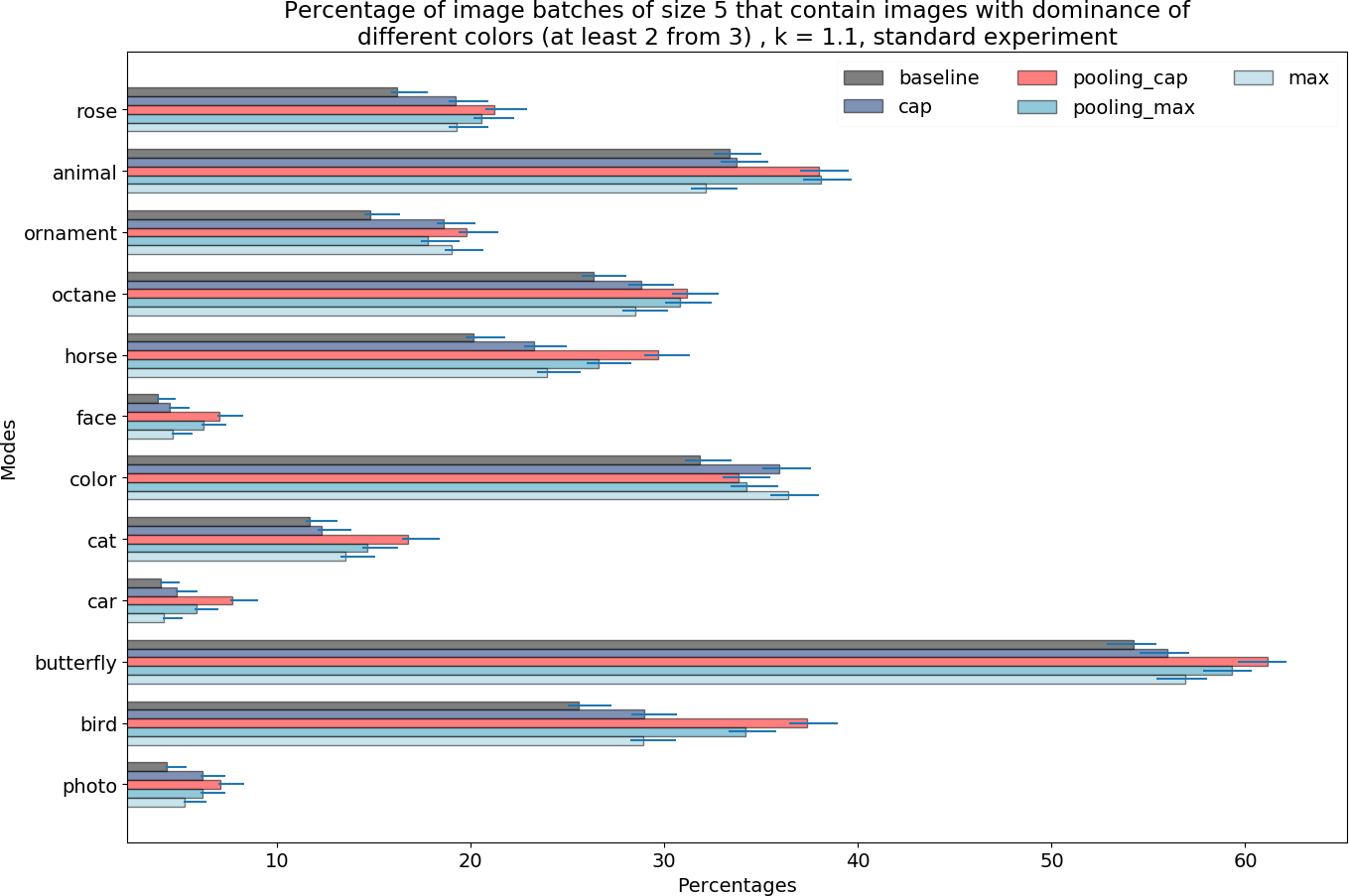}
\caption{Comparison of different modes for various prompts  in regard to the percentage of batches containing images with different dominant colors for the following parameters:   $K=1.1$, batch size =  5, standard experiment}
\label{k_11_5_standard}
\end{figure*}

Figures \ref{k_11_5} and \ref{k_11_5_standard} present the results for the $K=1.1$ coefficient and a batch size of 5 in both long and standard experiments.

\textbf{Batch size = 10}

\begin{figure*}[h!]
\centering
\includegraphics[width=0.79\textwidth]{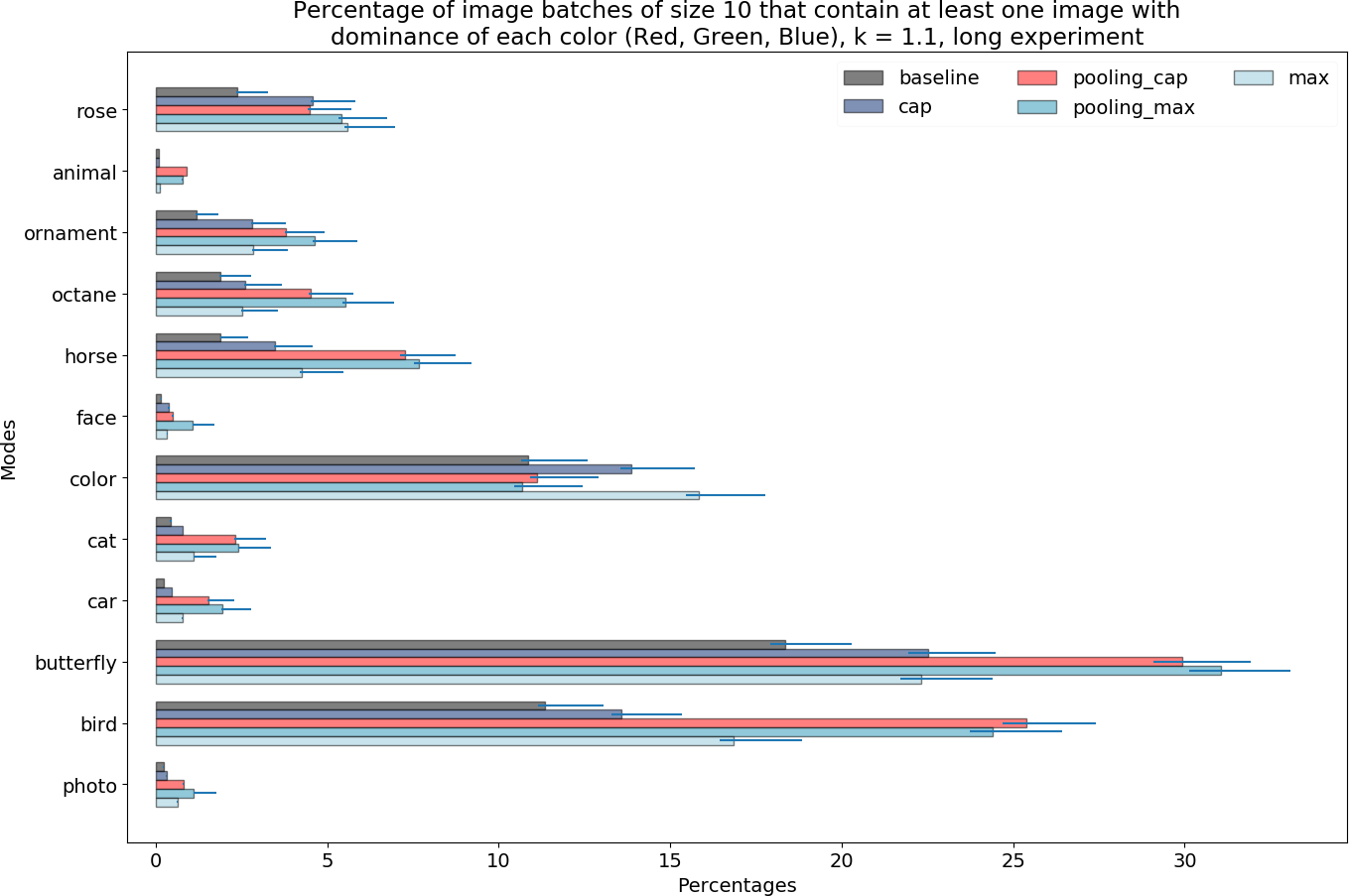}
\includegraphics[width=0.79\textwidth]{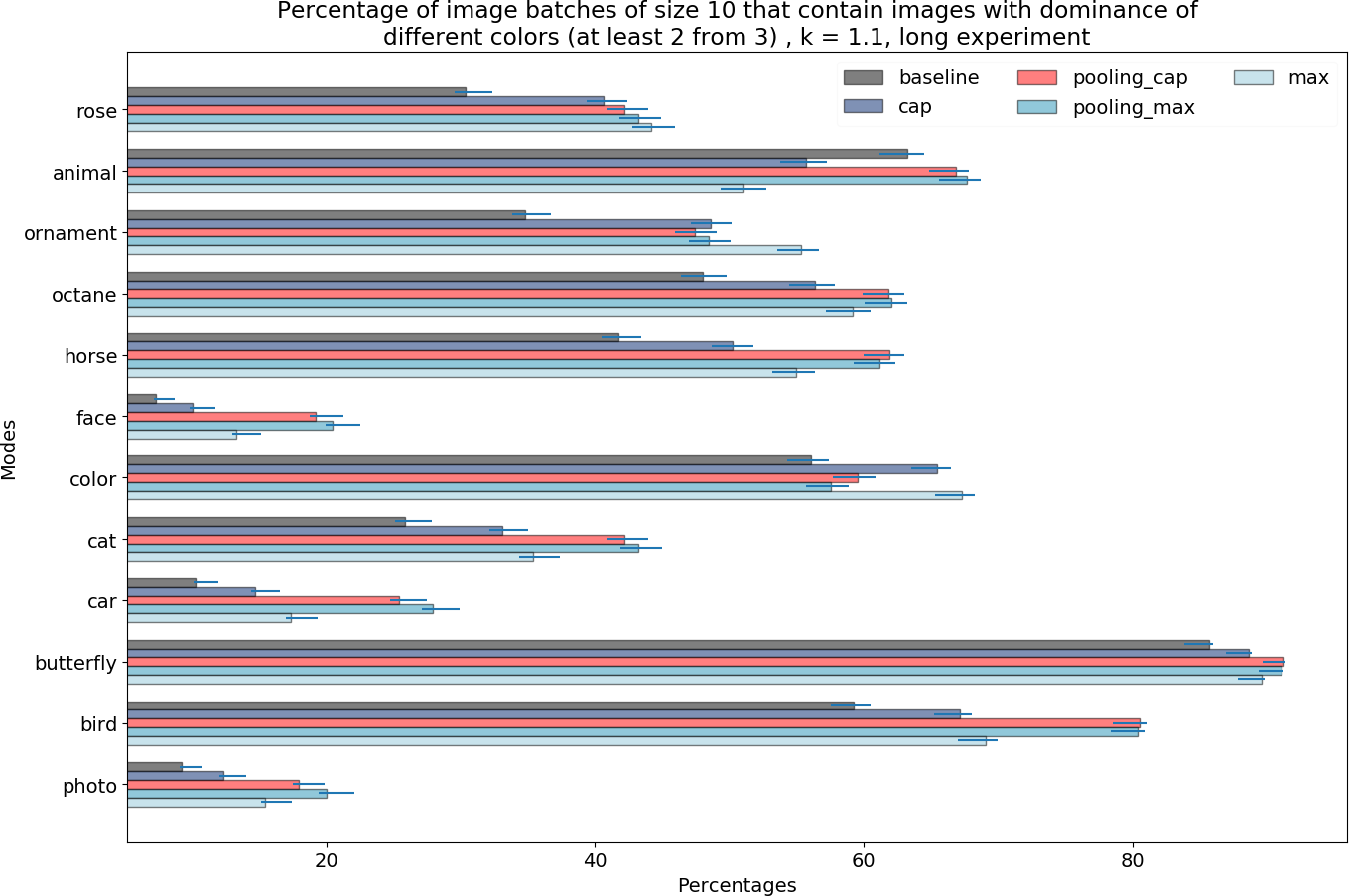}
\caption{Comparison of different modes for various prompts  in regard to the percentage of batches containing images with different dominant colors for the following parameters:   $K=1.1$, batch size =  10, long experiment.}
\label{k_11_10}
\end{figure*}

\begin{figure*}[h!]
\centering
\includegraphics[width=0.79\textwidth]{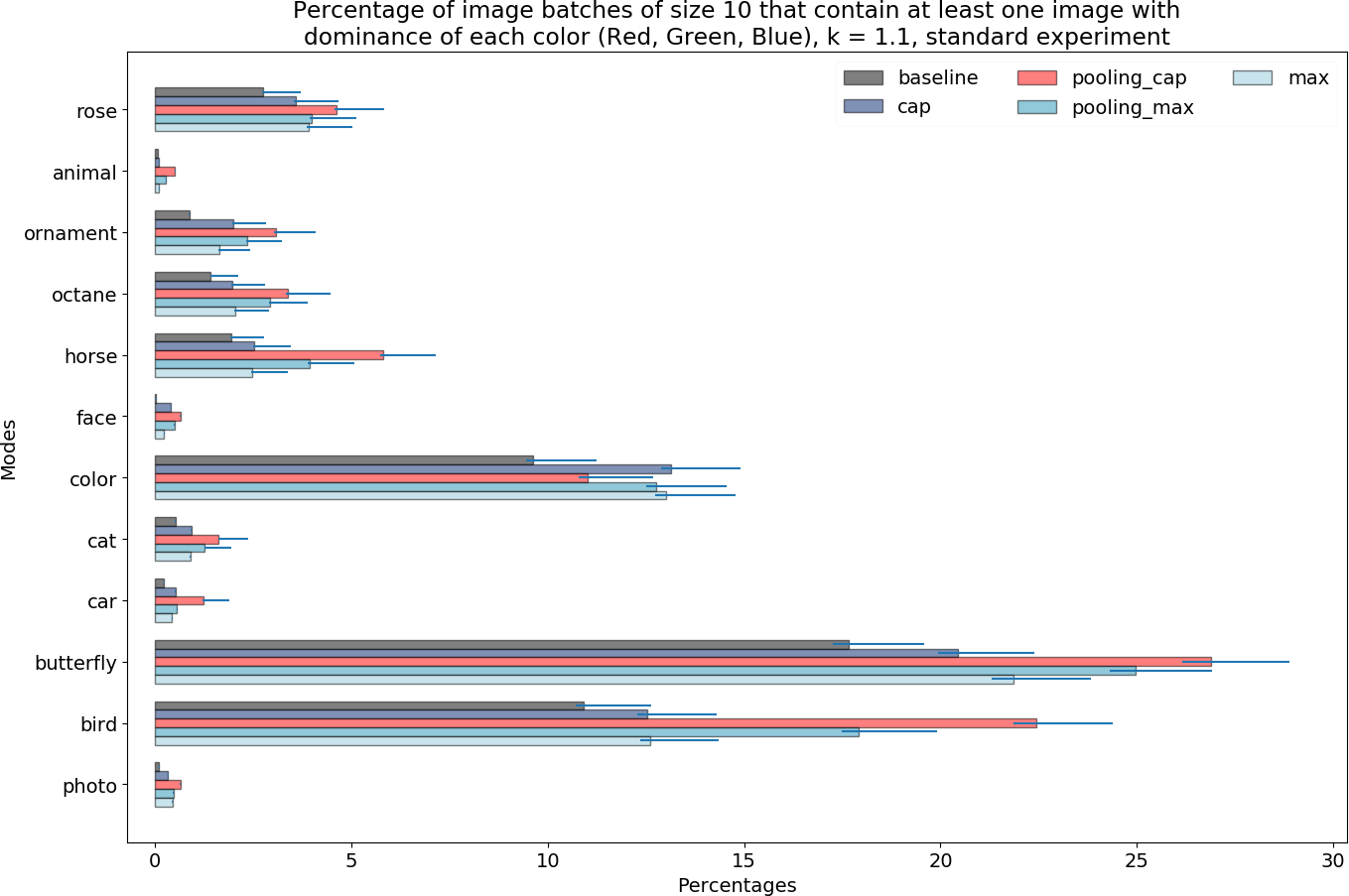}
\includegraphics[width=0.79\textwidth]{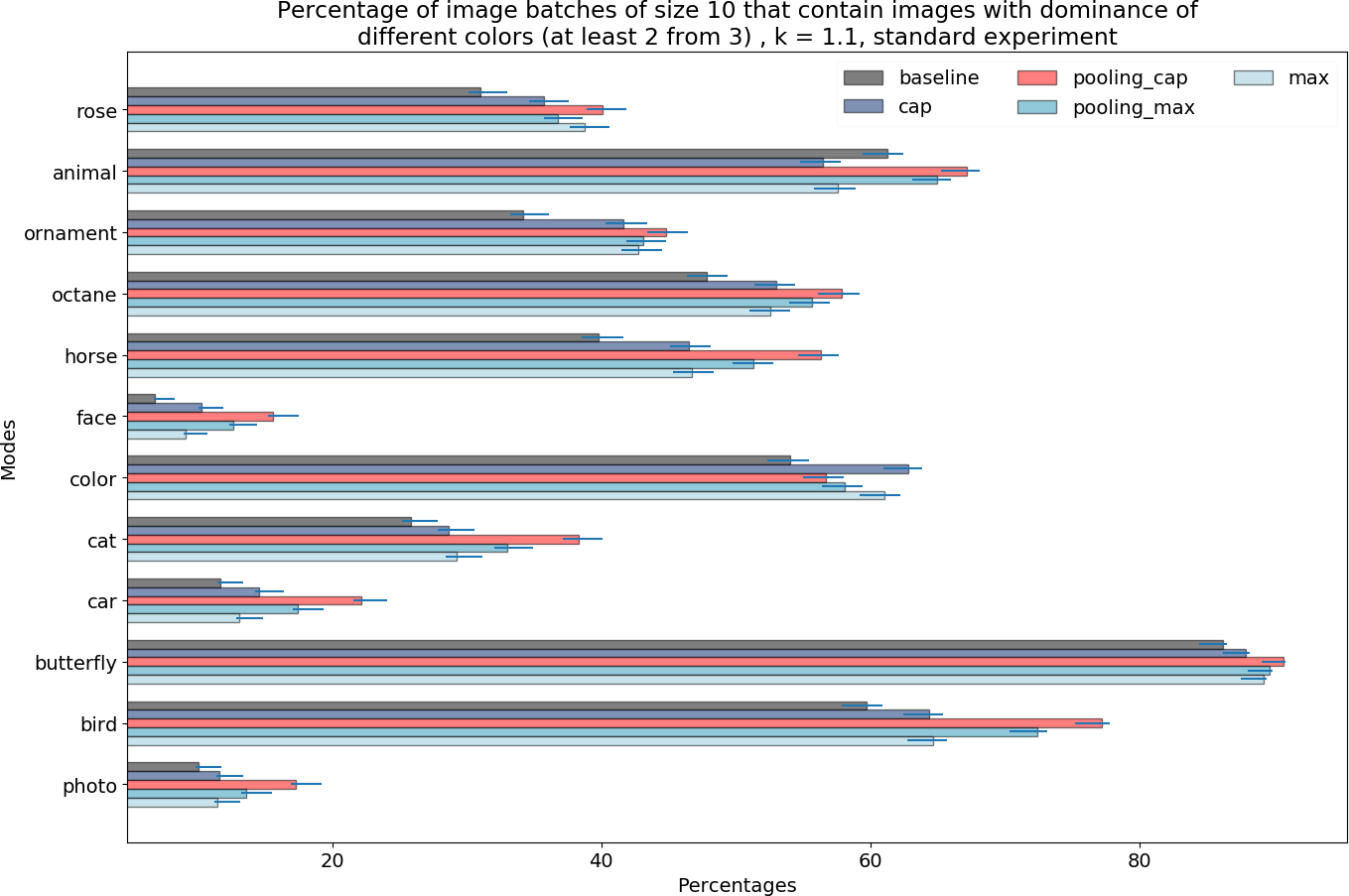}
\caption{Comparison of different modes for various prompts in regard to the percentage of batches containing images with different dominant colors for the following parameters:   $K=1.1$, batch size =  10, standard experiment.}
\label{k_11_10_standard}
\end{figure*}

Figures \ref{k_11_10} and \ref{k_11_10_standard} showcase the results for the $K=1.1$ coefficient and a batch size of 10 in both long and standard experiments.

\textbf{Batch size = 50}

\begin{figure*}[h!]
\centering
\includegraphics[width=0.79\textwidth]{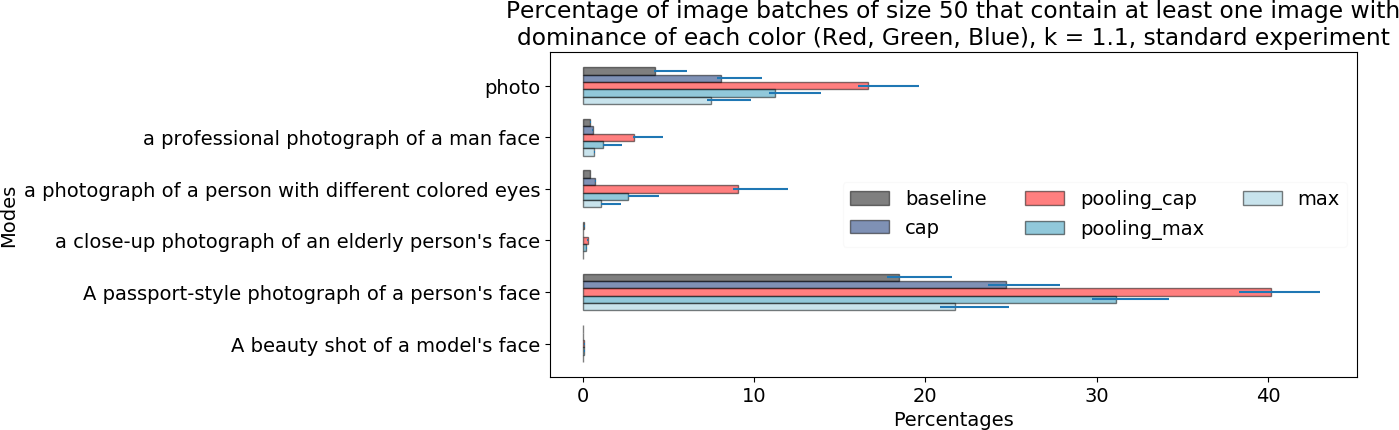}
\includegraphics[width=0.79\textwidth]{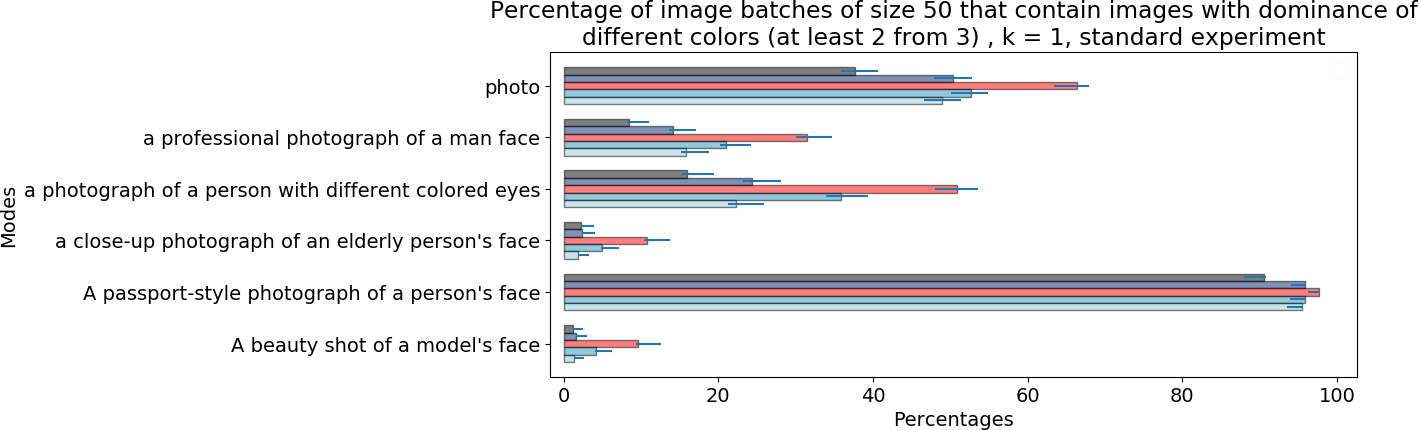}
\caption{Comparison of different modes for various prompts in regard to the percentage of batches containing images with different dominant colors for the following parameters:   $K=1.1$, batch size = 50, standard experiment.}
\label{k_11_50}
\end{figure*}

Figure \ref{k_11_50} illustrates the results for the $K=1$ coefficient and a batch size of 50 in the standard experiment.

\subsection{K=1.2}

\textbf{Batch size = 3}

\begin{figure*}[h!]
\centering
\includegraphics[width=0.79\textwidth]{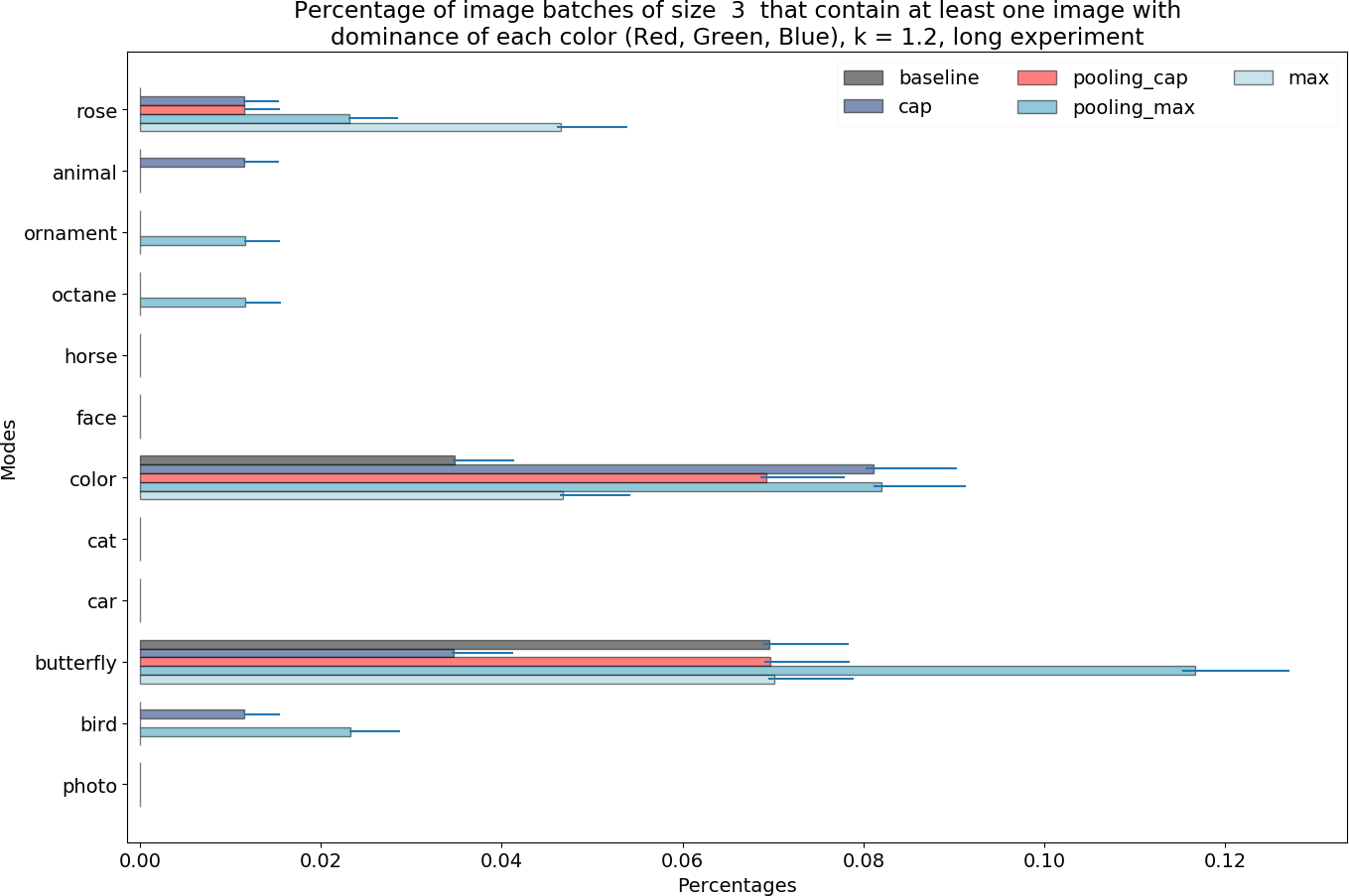}
\includegraphics[width=0.79\textwidth]{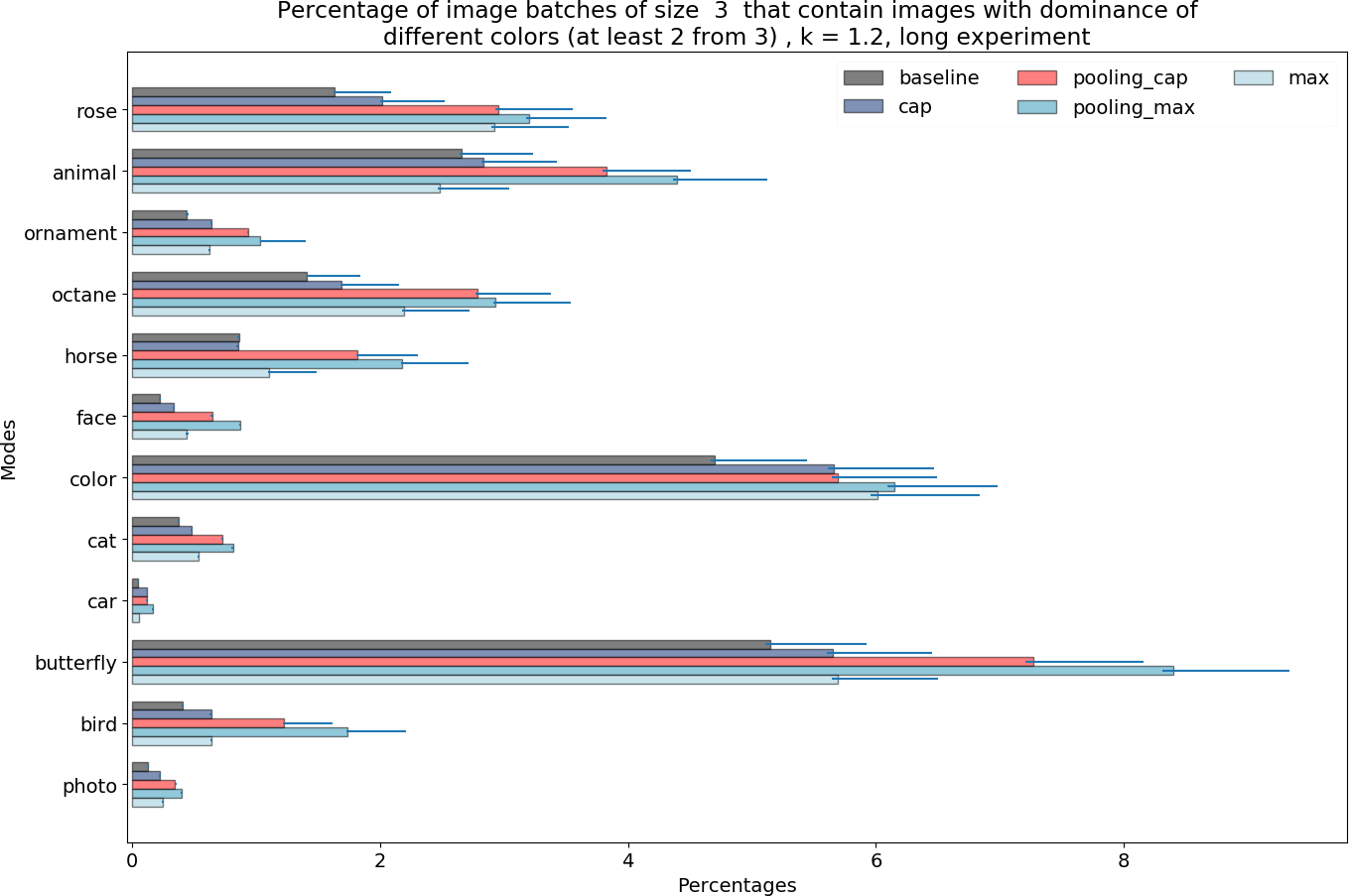}
\caption{Comparison of different modes for various prompts  in regard to the percentage of batches containing images with different dominant colors for the following parameters:   $K=1.2$, batch size =  3, long experiment.}
\label{k_12_3}
\end{figure*}

\begin{figure*}[h!]
\centering
\includegraphics[width=0.79\textwidth]{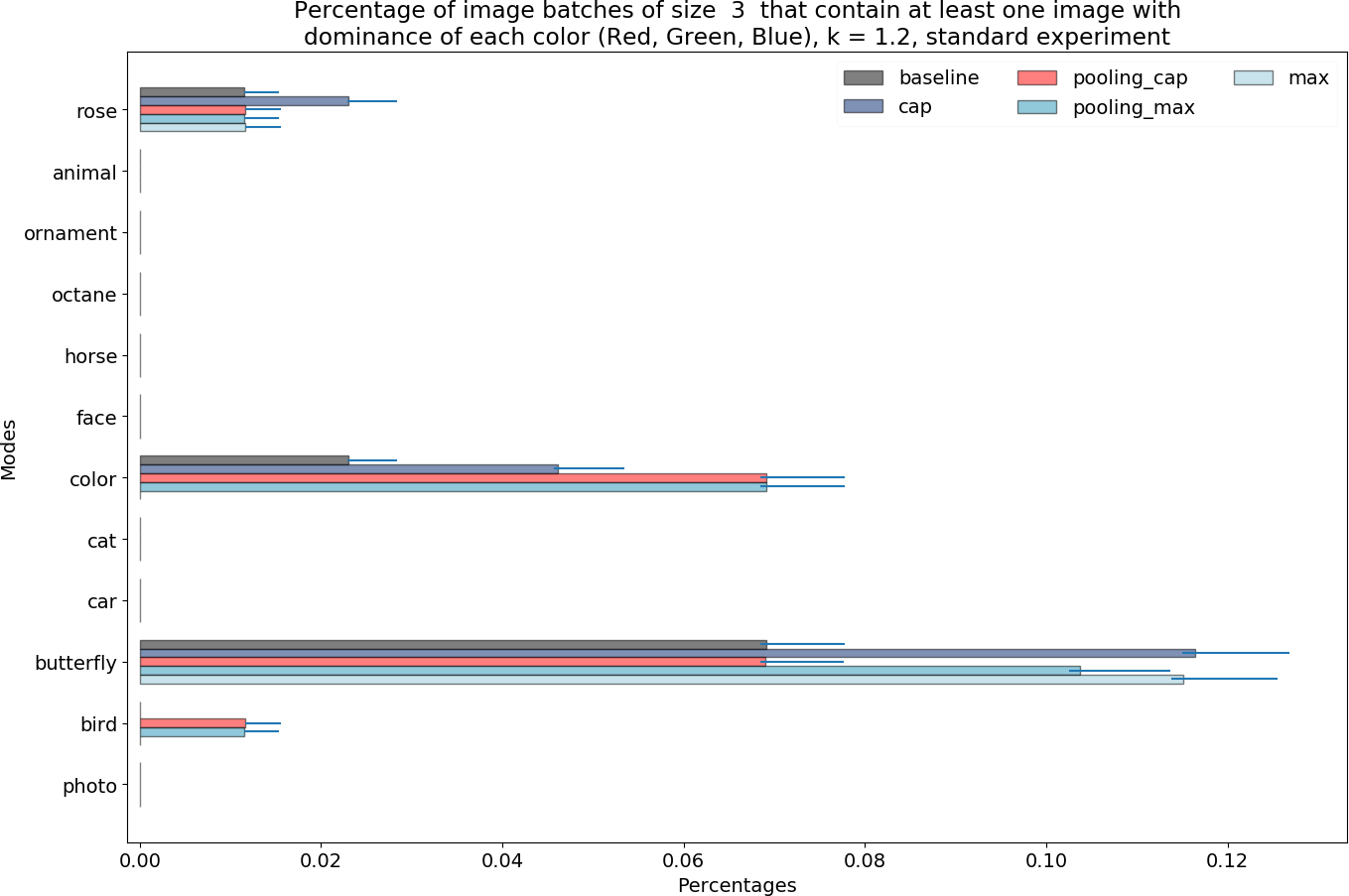}
\includegraphics[width=0.79\textwidth]{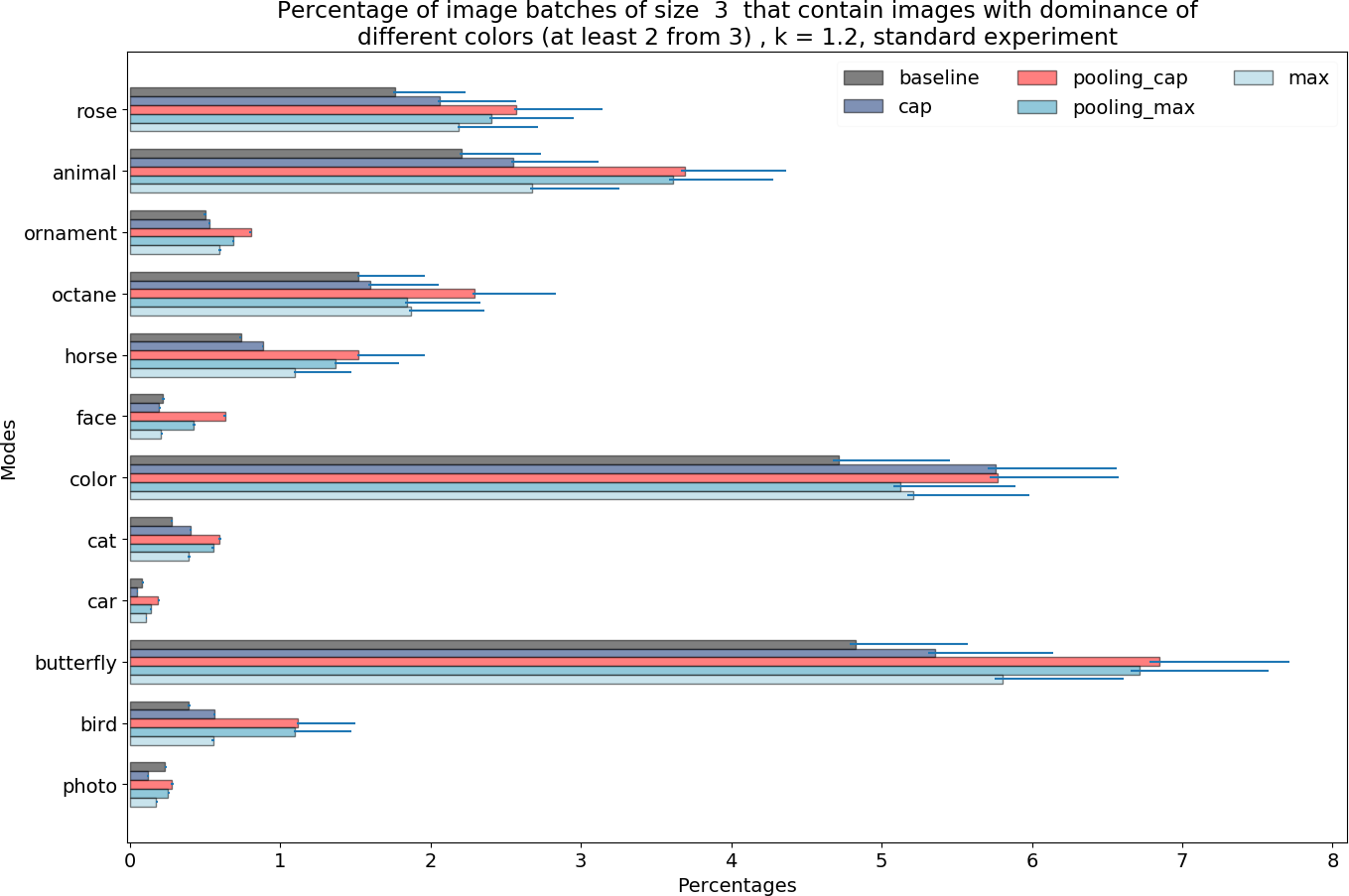}
\caption{Comparison of different modes for various prompts in regard to the percentage of batches containing images with different dominant colors for the following parameters:   $K=1.2$, batch size =  3, standard experiment.}
\label{k_12_3_standard}
\end{figure*}

Figures \ref{k_12_3} and \ref{k_12_3_standard} display the results for the $K=1.2$ coefficient and a batch size of 3 in both long and standard experiments.

\textbf{Batch size = 5}

\begin{figure*}[h!]
\centering
\includegraphics[width=0.79\textwidth]{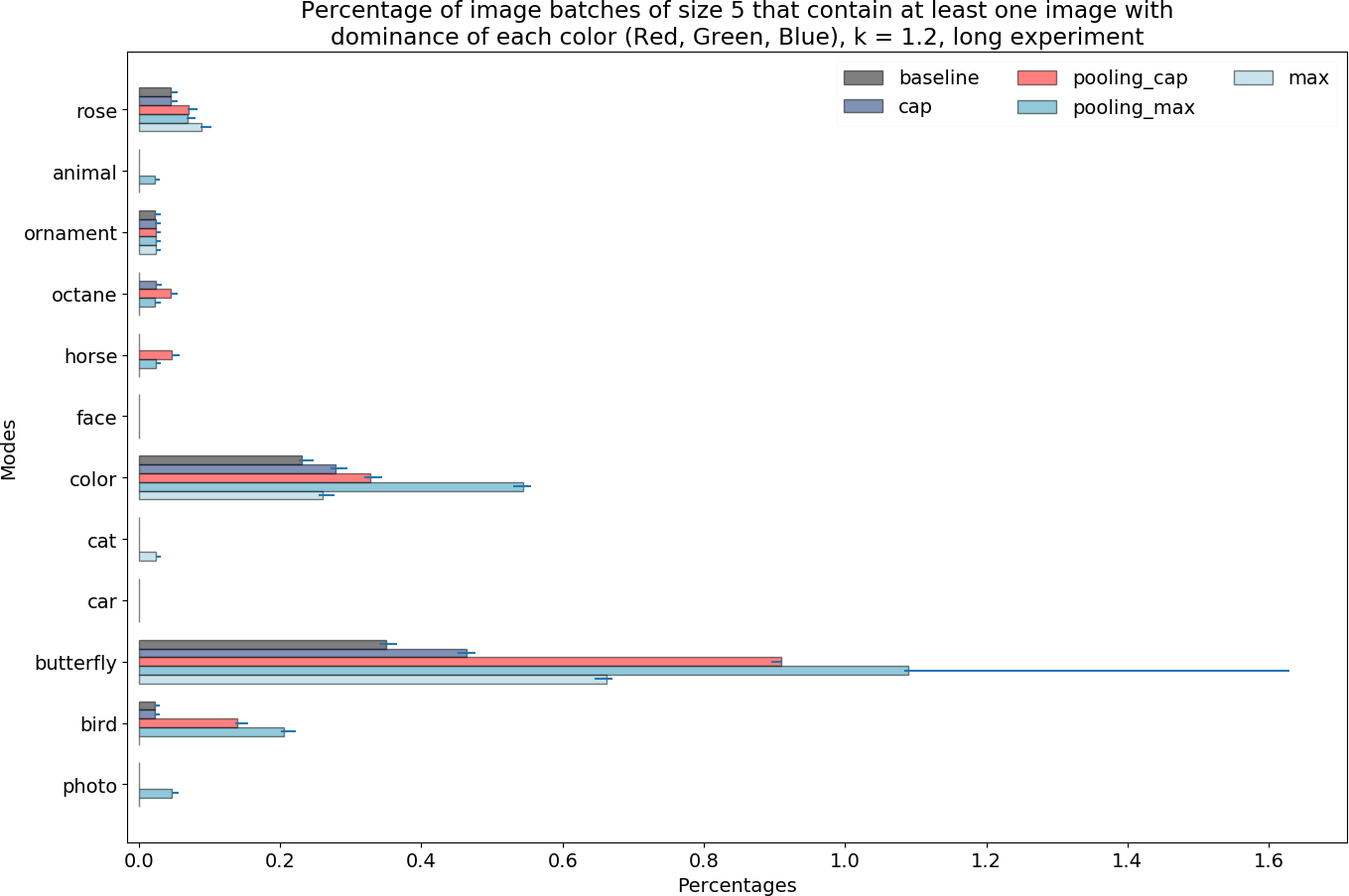}
\includegraphics[width=0.79\textwidth]{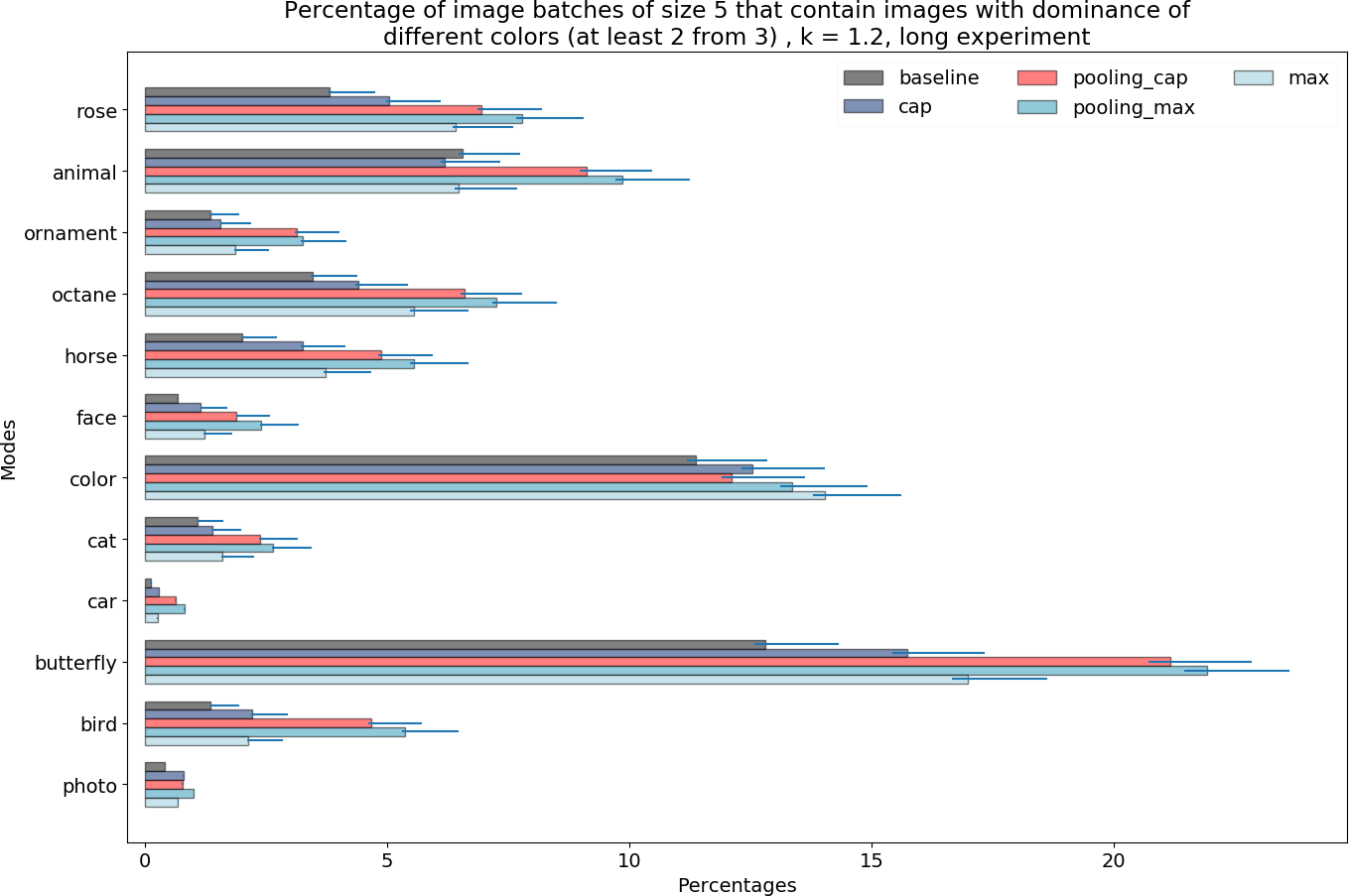}
\caption{Comparison of different modes for various prompts  in regard to the percentage of batches containing images with different dominant colors for the following parameters:   $K=1.2$, batch size =  5, long experiment.}
\label{k_12_5}
\end{figure*}

\begin{figure*}[h!]
\centering
\includegraphics[width=0.79\textwidth]{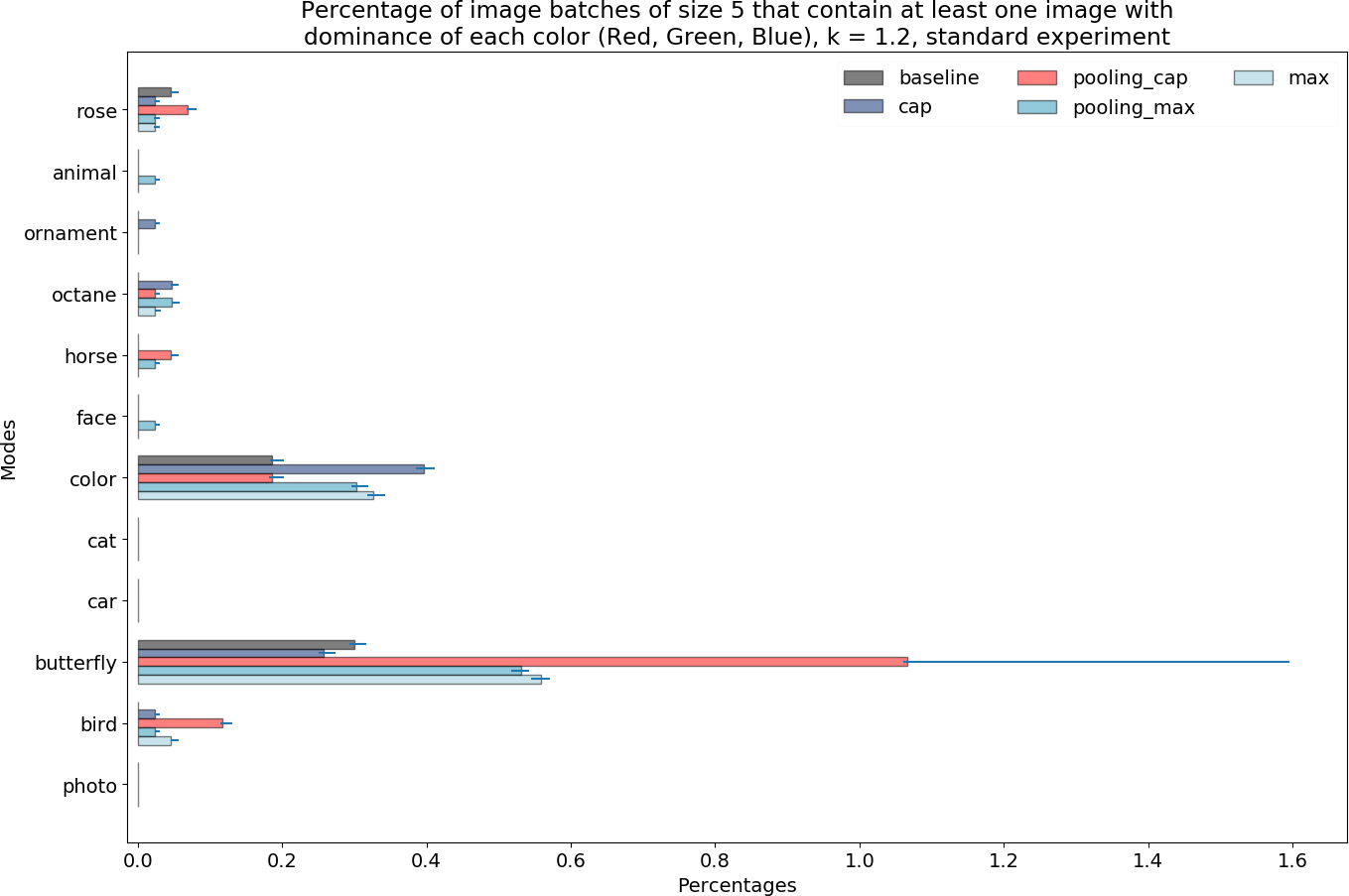}
\includegraphics[width=0.79\textwidth]{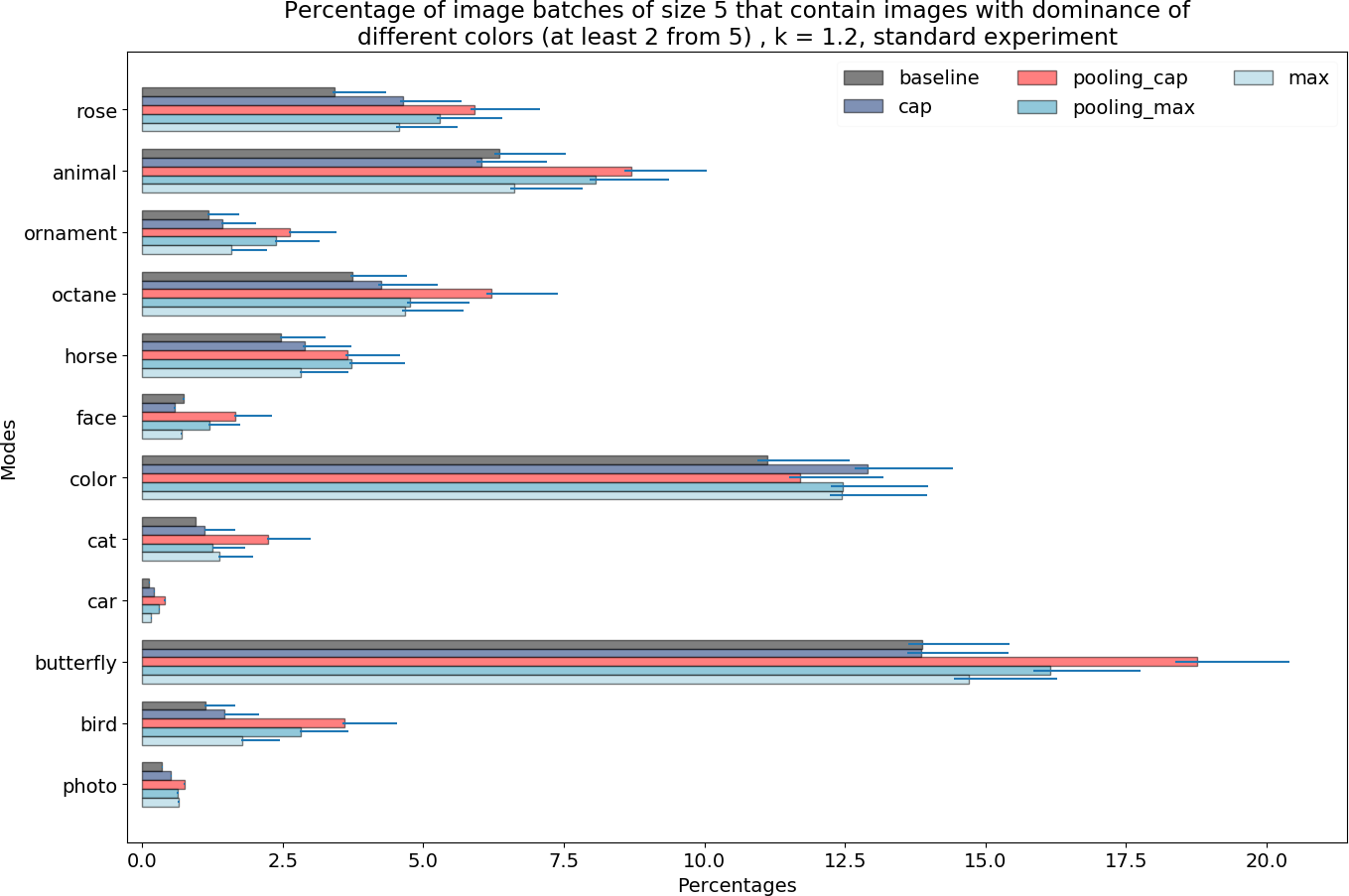}
\caption{Comparison of different modes for various prompts in regard to the percentage of batches containing images with different dominant colors for the following parameters:   $K=1.2$, batch size =  5, standard experiment.}
\label{k_12_5_standard}
\end{figure*}

Figures \ref{k_12_5} and \ref{k_12_5_standard} present the results for the $K=1.2$ coefficient and a batch size of 5 in both long and standard experiments.

\textbf{Batch size = 10}

\begin{figure*}[h!]
\centering
\includegraphics[width=0.79\textwidth]{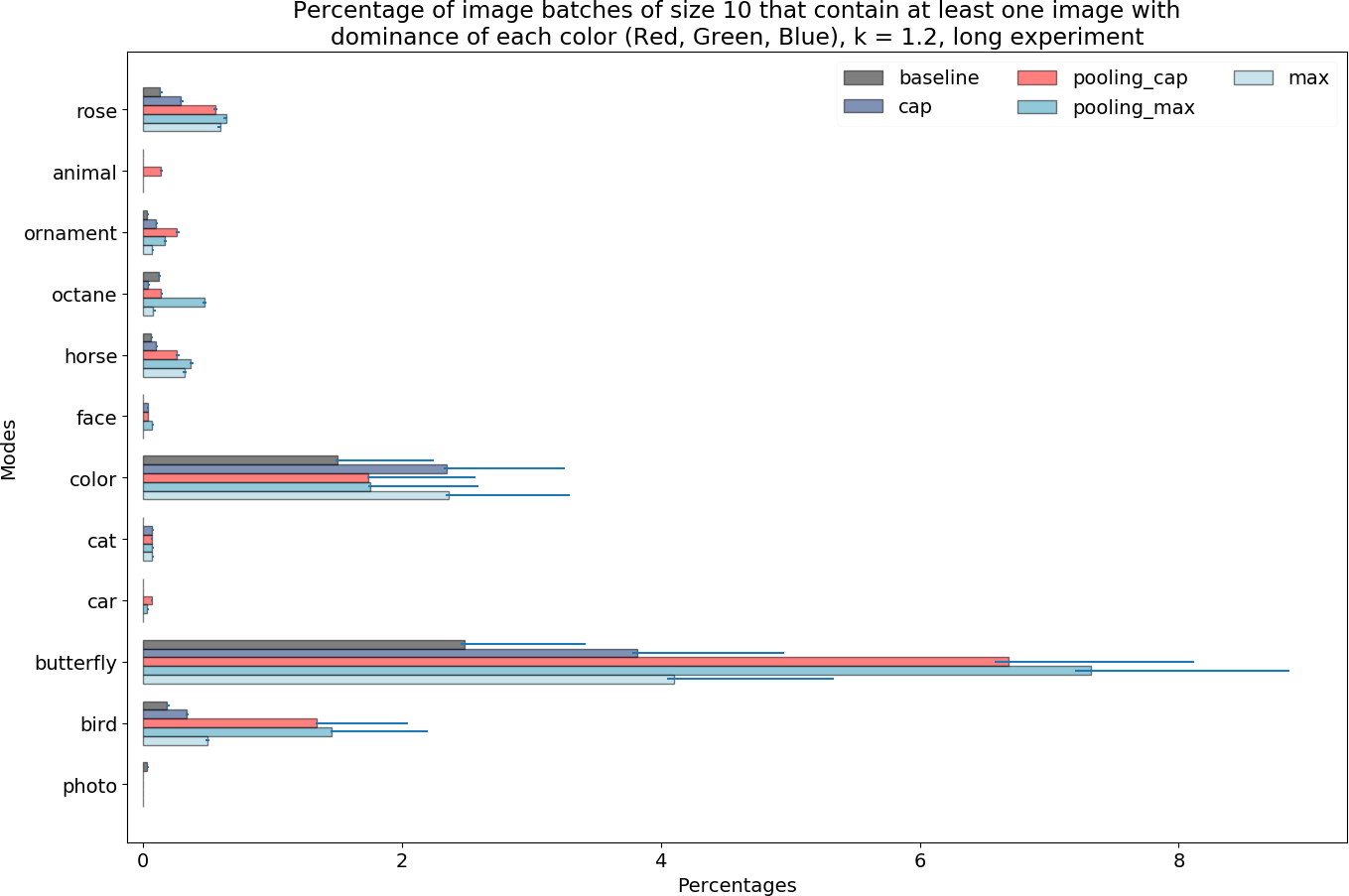}
\includegraphics[width=0.79\textwidth]{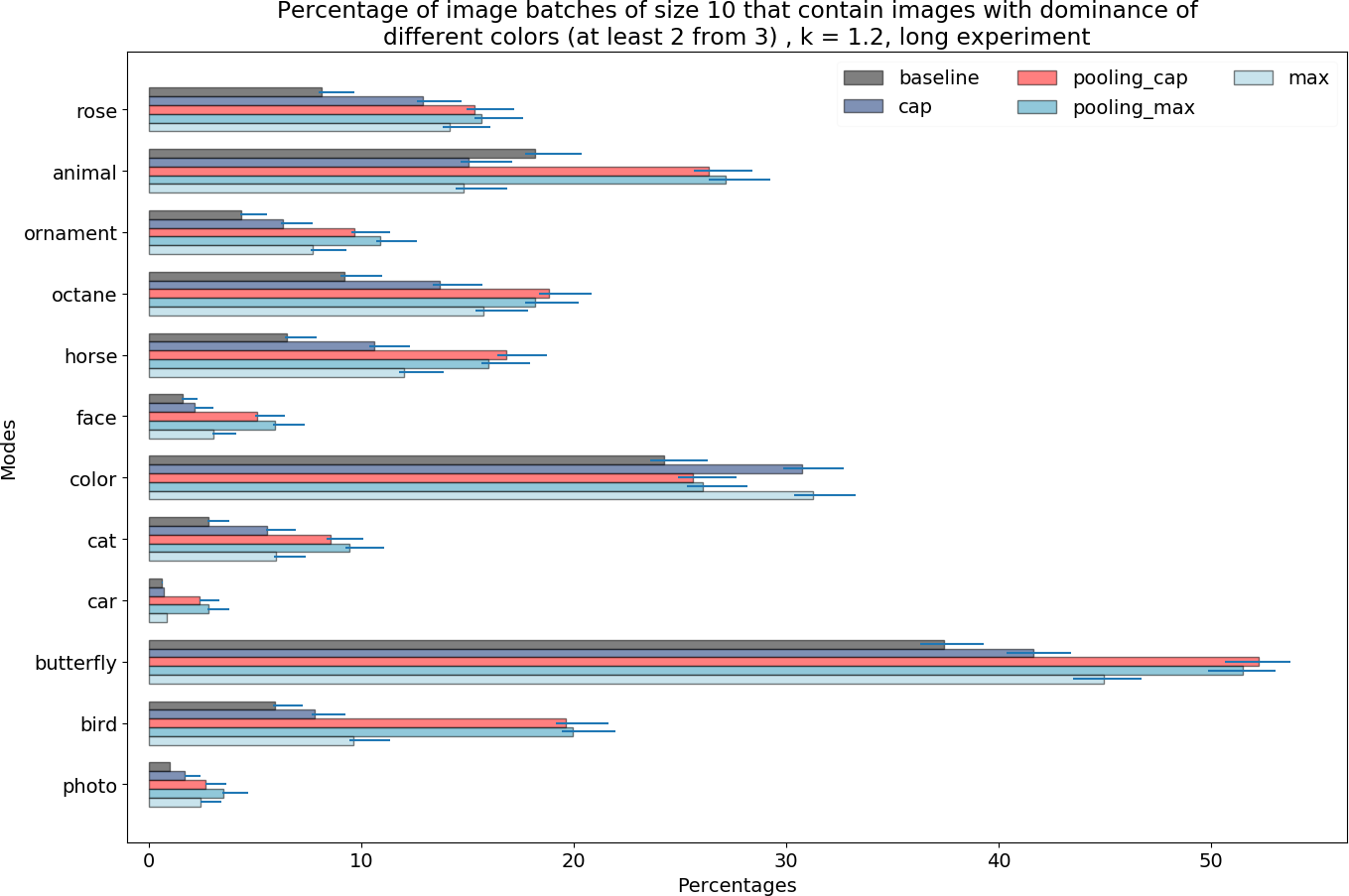}
\caption{Comparison of different modes for various prompts  in regard to the percentage of batches containing images with different dominant colors for the following parameters:   $K=1.2$, batch size =  10, long experiment.}
\label{k_12_10}
\end{figure*}

\begin{figure*}[h!]
\centering
\includegraphics[width=0.79\textwidth]{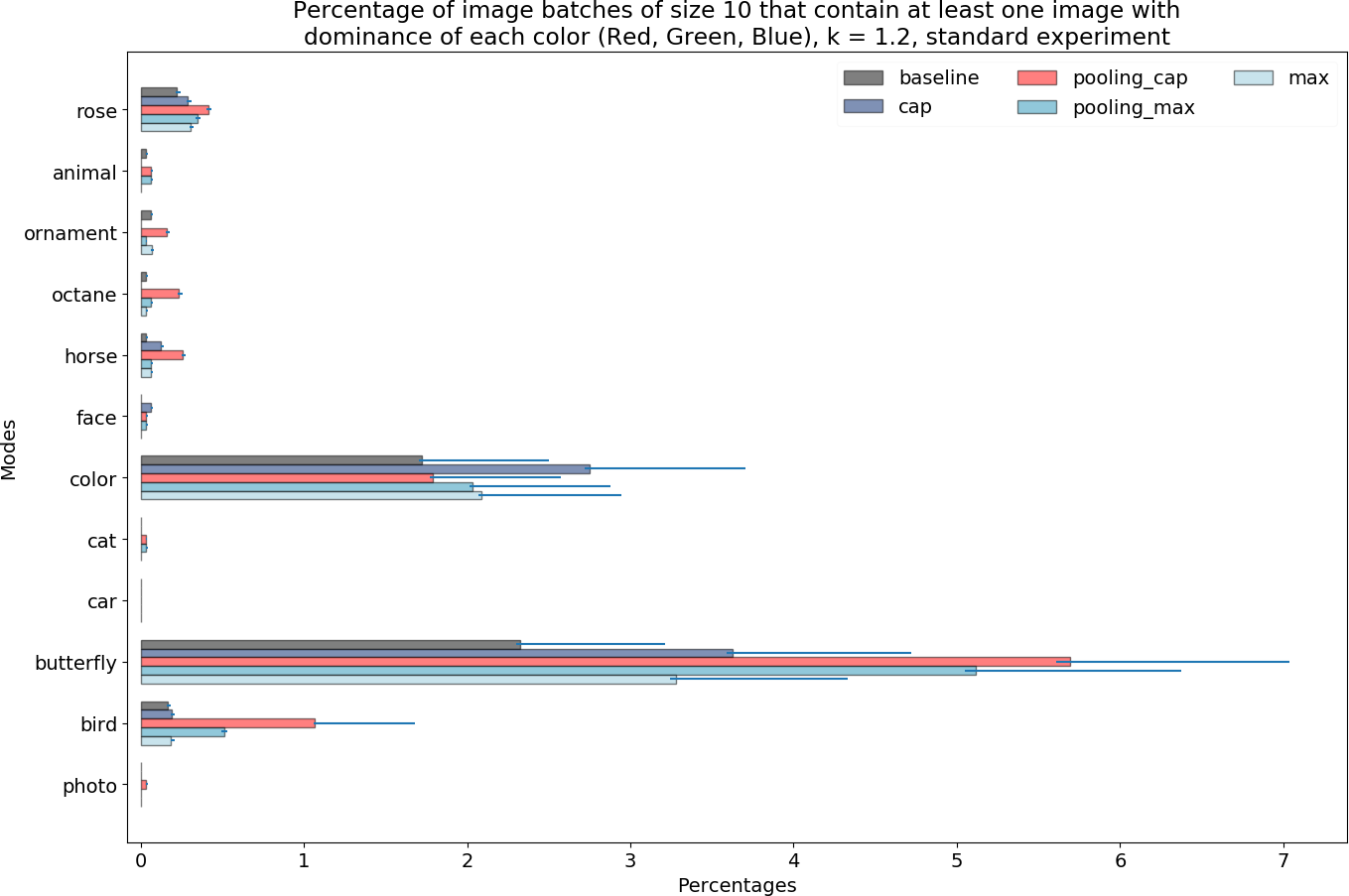}
\includegraphics[width=0.79\textwidth]{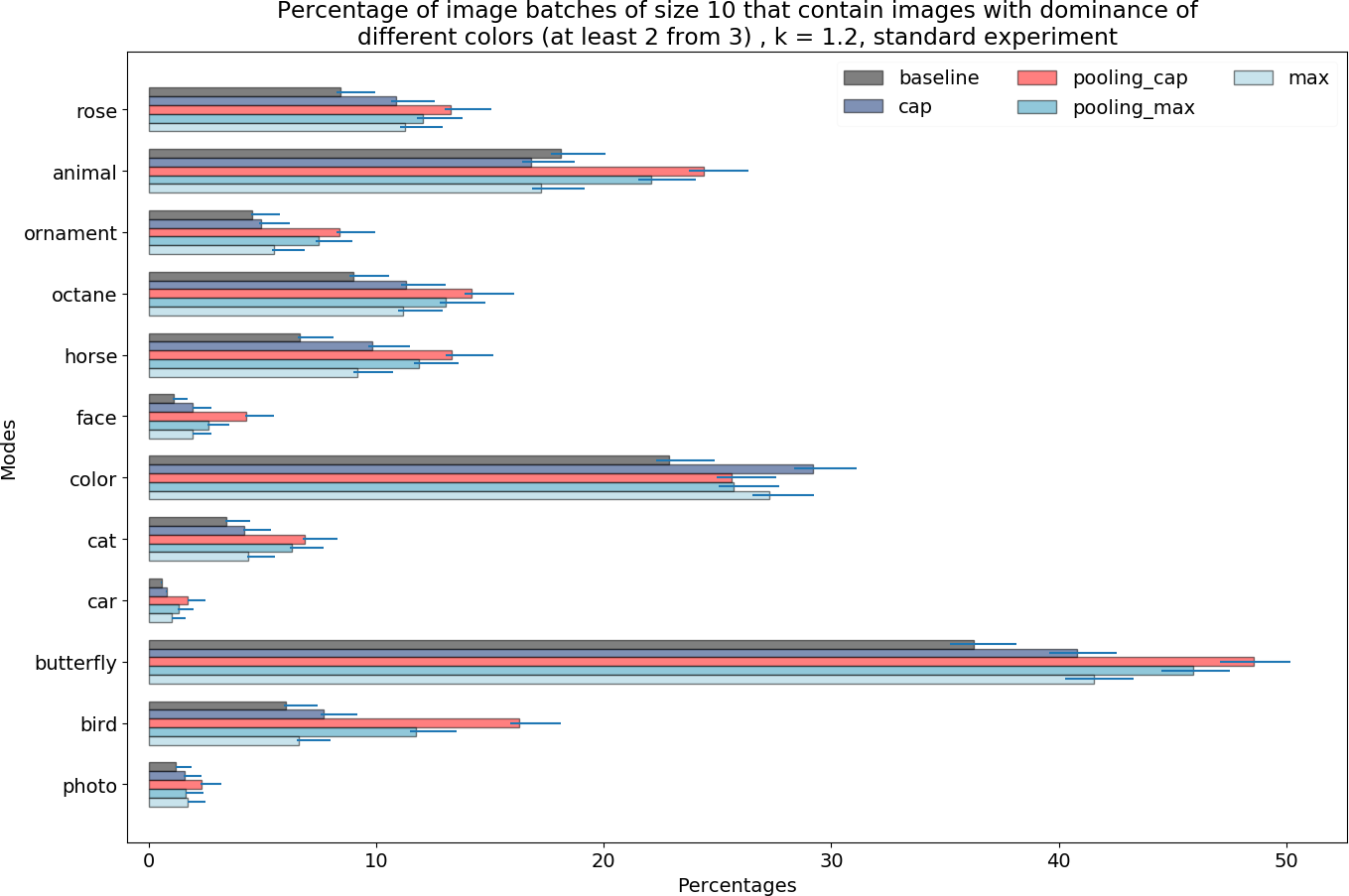}
\caption{Comparison of different modes for various prompts in regard to the percentage of batches containing images with different dominant colors for the following parameters:   $K=1.2$, batch size =  10, standard experiment.}
\label{k_12_10_standard}
\end{figure*}

Figures \ref{k_12_10} and \ref{k_12_10_standard} illustrate the results for the $K=1.2$ coefficient and a batch size of 10 in both long and standard experiments.

\textbf{Batch size = 50}

\begin{figure*}[h!]
\centering
\includegraphics[width=0.79\textwidth]{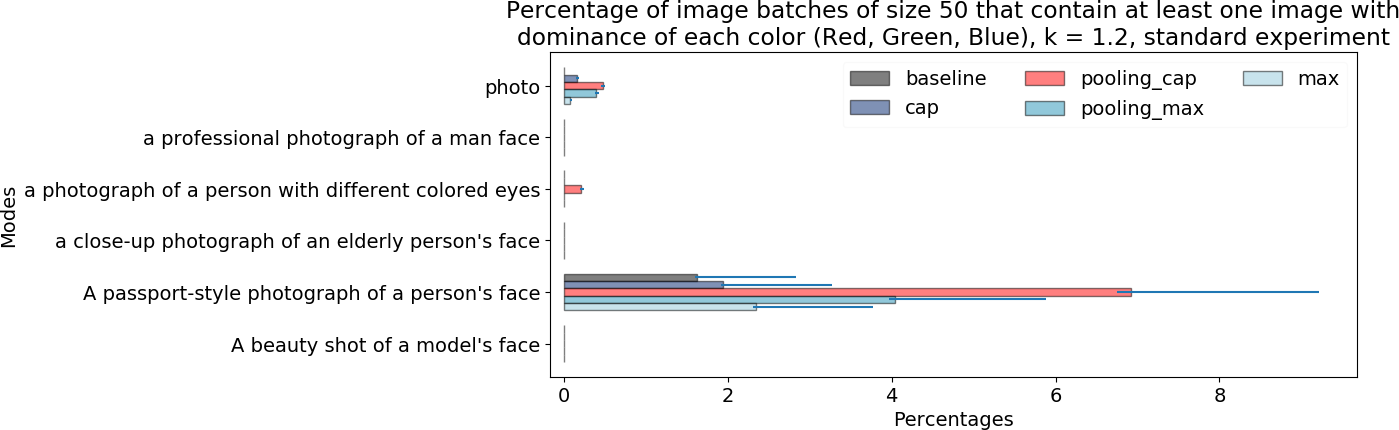}
\includegraphics[width=0.79\textwidth]{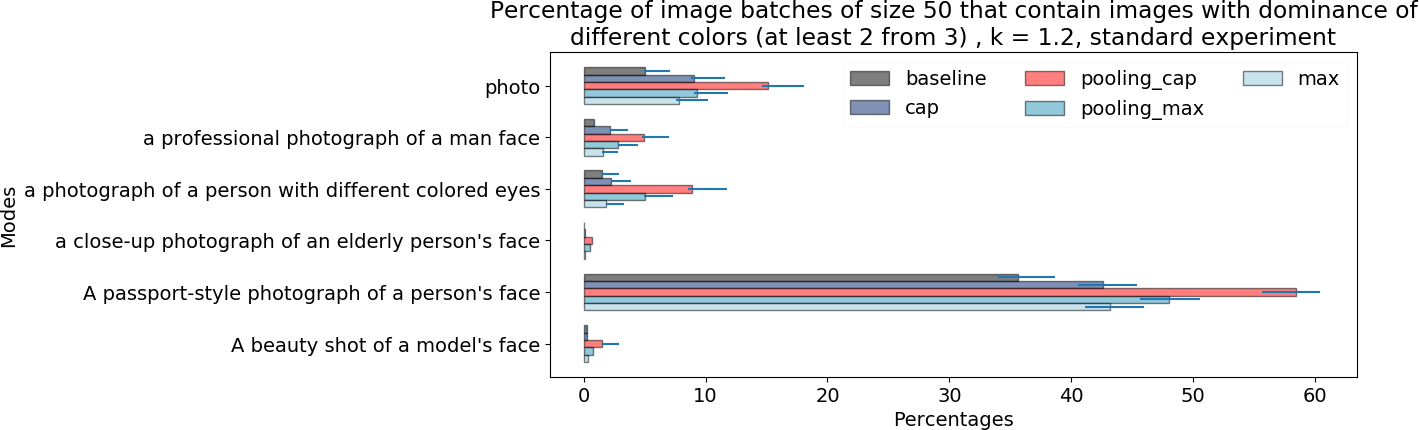}
\caption{Comparison of different modes for various prompts in regard to the percentage of batches containing images with different dominant colors for the following parameters:   $K=1.2$, batch size = 50, standard experiment.}
\label{k_12_50}
\end{figure*}

Figure \ref{k_12_50} illustrates the results for the $K=1$ coefficient and a batch size of 50 in the standard experiment.

\begin{figure*}[h!]
\centering
\includegraphics[width=0.79\textwidth]{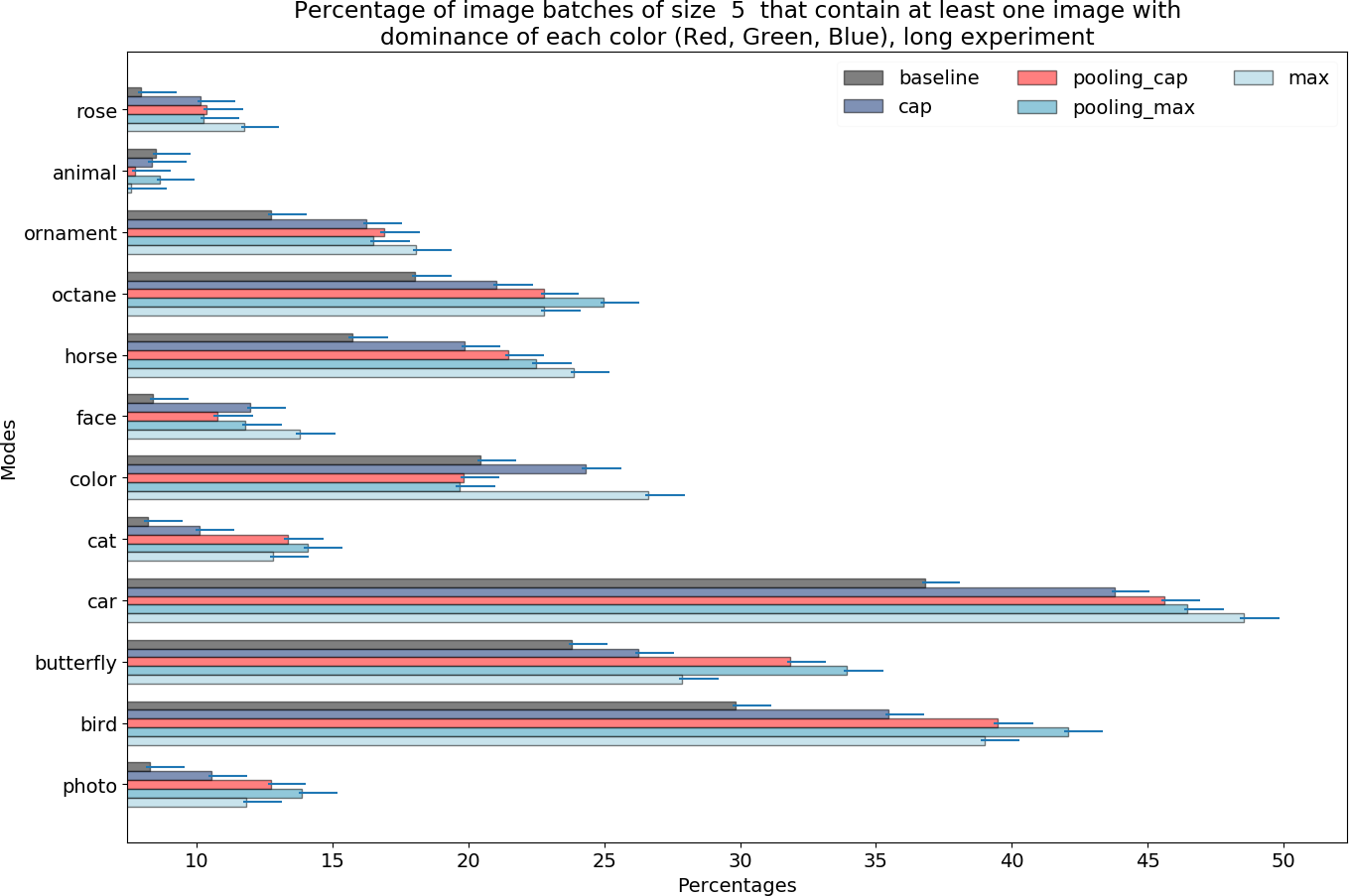}
\includegraphics[width=0.79\textwidth]{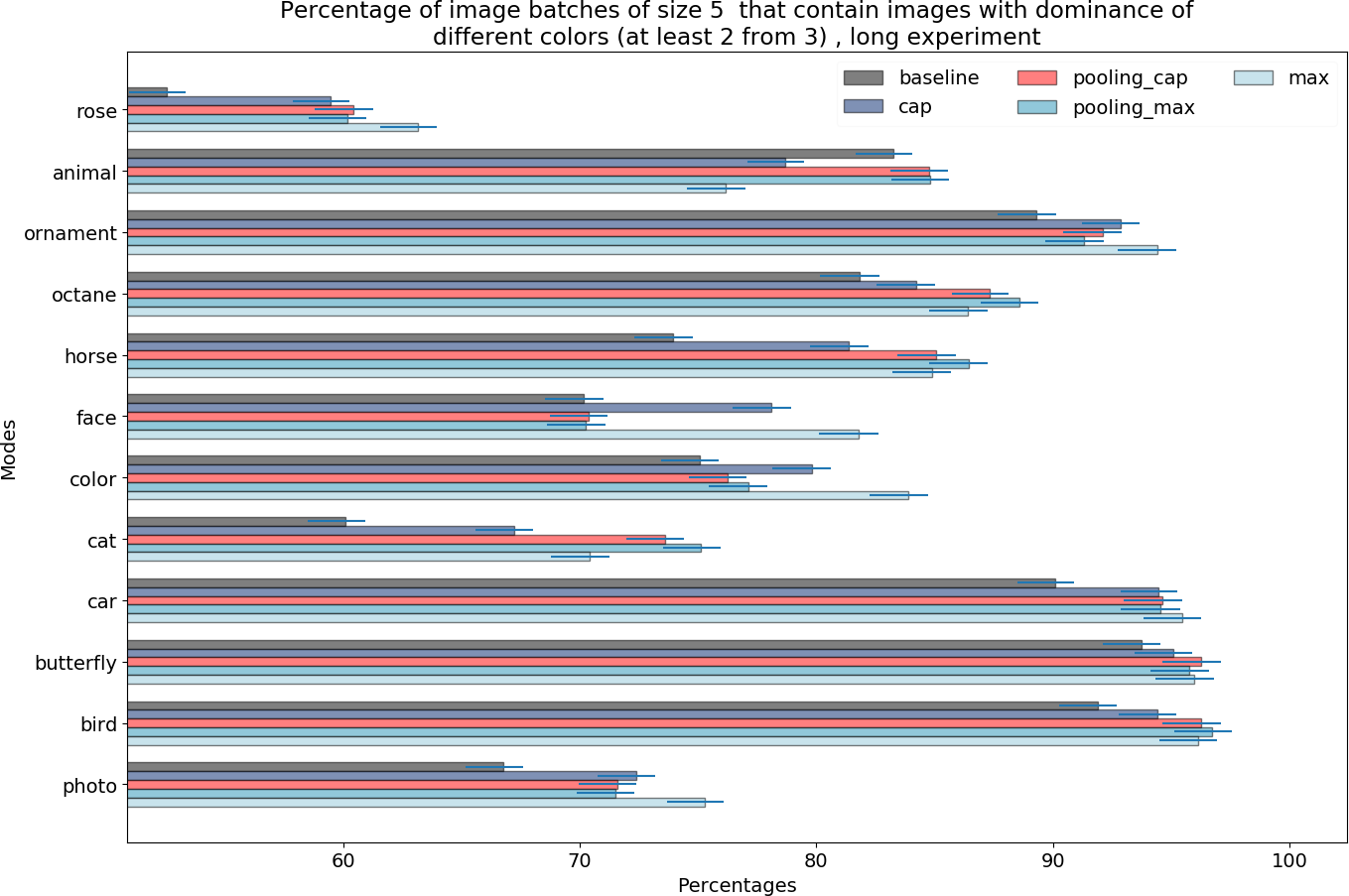}
\caption{Comparison of different modes for various prompts  in regard to the percentage of batches containing images with different dominant colors for the following parameters:   $K=1$, batch size =  5, long experiment}
\label{k_1_5}
\end{figure*}

These supplementary results provide an evaluation of the color diversity improvement achieved by our proposed approach across various coefficient values and batch sizes. The figures demonstrate the effectiveness of the ``pooling\_cap'' method in enhancing color diversity compared to the baseline Stable Diffusion method.

\section{Additional results for LPIPS evaluation}

In the main paper for the small batch sizes, we provide the LPIPS evaluation results only for the ``long experiment''. Here, we provide missing results for the ``standard experiment'' setting.

\begin{figure*}[h!] \begin{center}
\par
{\includegraphics[width=0.95\textwidth]{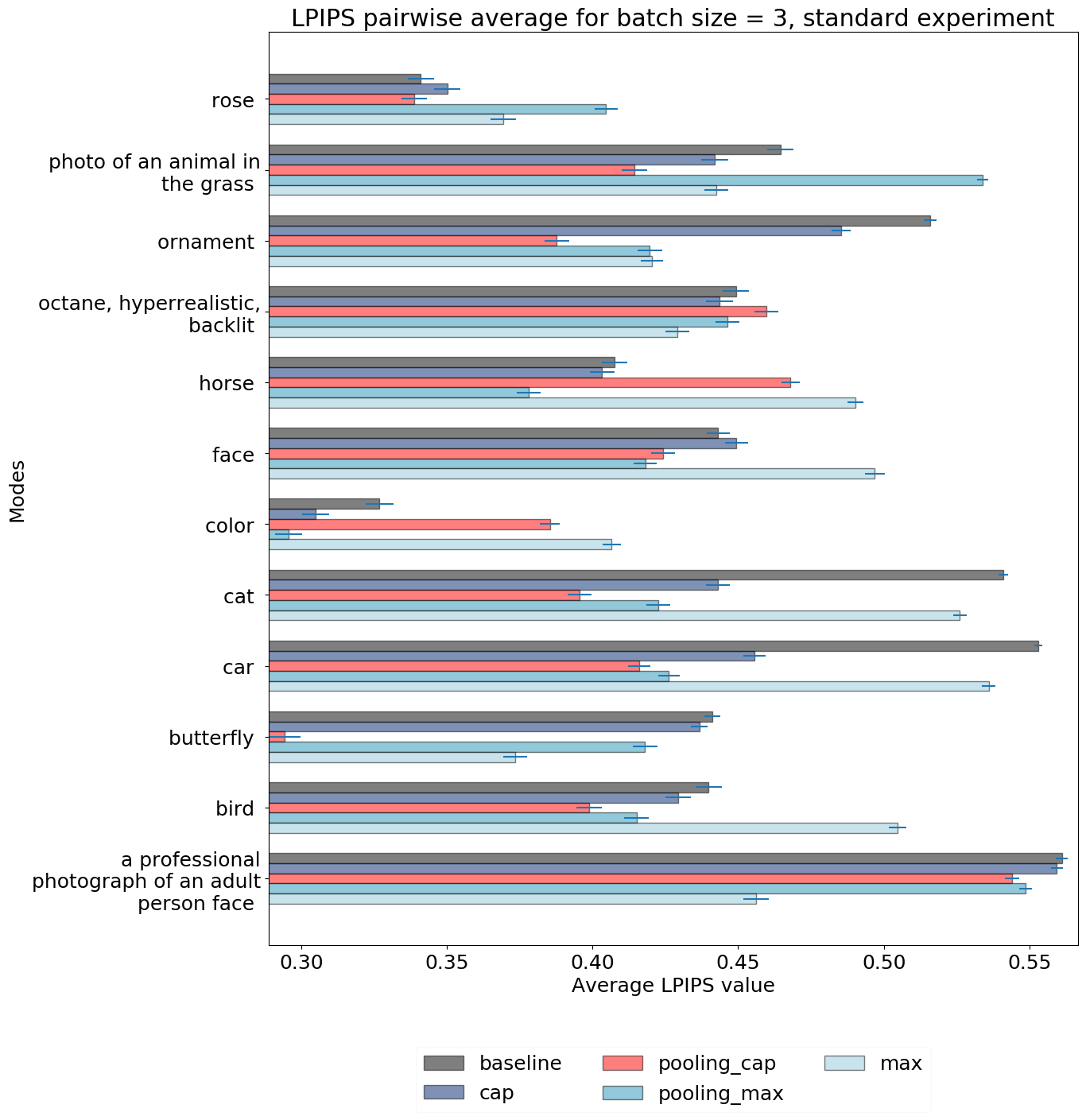} }%
\hfill
\caption{Average batch pairwise LPIPS, batch size=3,
 standard experiment \label{3_lpips_ST}: no clear conclusion overall,} {due to the small batch size.}
\end{center}
\end{figure*}

Figure \ref{3_lpips_ST} presents the average batch pairwise LPIPS distance for a batch size of 3 for the ``standard experiment'' setting. Here, same as in Figure \ref{3_lpips} it is hard to see what method is the best for diversity due to the small batch size

\begin{figure*}[h!] \begin{center}
\par
{\includegraphics[width=0.97\textwidth]{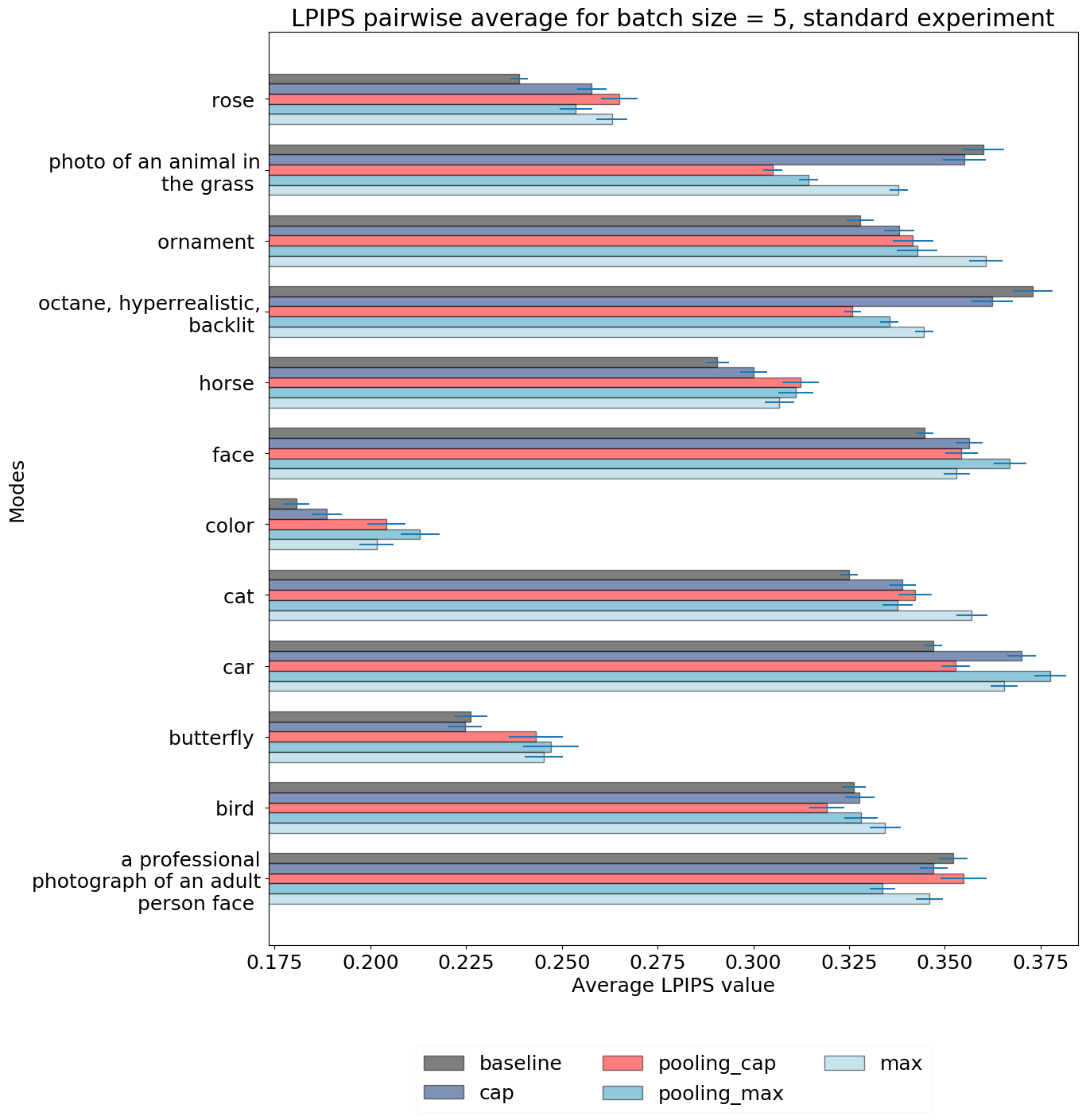} }%
\hfill
\caption{Average batch pairwise LPIPS, batch size=5,
standard experiment\label{5_lpips_ST}}
\end{center}
\end{figure*}

Figure \ref{5_lpips_ST} illustrates the average batch pairwise LPIPS distance for a batch size of 5 in the ``standard experiment'' setting. We can see that the results presented in Figure \ref{5_lpips_ST} are less conclusive than the results in Figure \ref{5_lpips}, which demonstrates that the ``long experiment'' setting is better suited for the small batch sizes.

\begin{figure*}[h!] \begin{center}
\par
{\includegraphics[width=0.97\textwidth]{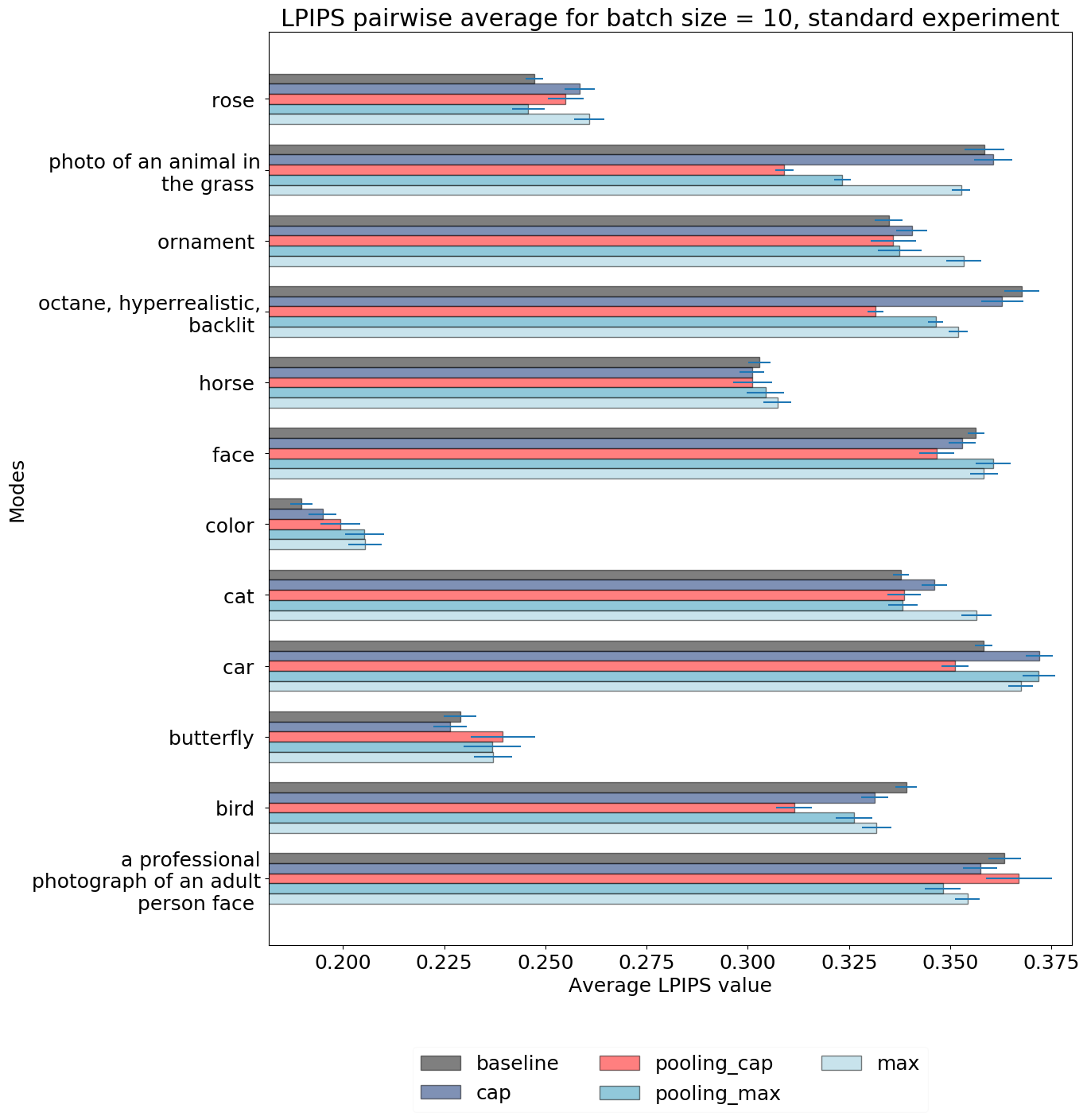} }%
\hfill
\caption{Average batch pairwise LPIPS, batch size=10\label{10_lpips_ST}}
\end{center}
\end{figure*}

Figure \ref{10_lpips_ST} illustrates the average batch pairwise LPIPS distance for a batch size of 10 in the ``standard experiment'' setting. We can see that similarly to the results for batch size 5, the results presented in Figure \ref{10_lpips_ST}  are less conclusive than the results in Figure \ref{10_lpips}, which again demonstrates that the ``long experiment'' setting is better suited for the small batch sizes.

\section{Reproducibility details}

In this section, we provide additional details on reproducibility.

\begin{itemize}
    \item In this paper, we do not provide any theoretical contributions, which is why the theoretical contribution section does not apply to the current work.
    \item {No external dataset is used, as the code generates all necessary data.}
    \item {The GPU/CPU requirements are specified  at the beginning of section ``Experiment setting''.}
    \item  {All results of all tested hyperparameters are included in the paper. } {There is no preliminary hyperparameter optimization conducted as part of the experimentation process.}
    \item  %{The anonymized code is available at https://anonymous.4open.science/r/DiverseDiffusion-1012.}
    {The code is and available online. It includes both the image generation attached and the evaluation methods that are discussed in this paper.} In addition, we have included supplementary examples of images generated by both our preferred method and the standard stable diffusion in the attachment. All these examples are structured similarly to Figure \ref{images_bs50} and \ref{SD_schema}.
\end{itemize}

\section{Ethics statement}

Our research is conducted with a commitment to ethical principles. The imagery used in this study comprises AI-generated faces, removing any potential ethical concerns related to real individuals.
The code for generating images and the evaluation methods are made openly available, enhancing transparency and facilitating reproducibility. To make our results statistically sound, we provide results with the  95\% confidence interval error bars.

Our core objective centers on enhancing image diversity within the AI framework, promoting inclusivity, and expanding the spectrum of artistic expression. With our research, we would like to contribute meaningfully to the advancement of AI-generated art while adhering to ethical standards.

\end{document}